\documentclass[10pt,twocolumn,letterpaper]{article}

\usepackage[pagenumbers]{cvpr} 

\usepackage{graphicx}
\usepackage{amsmath}
\usepackage{amssymb}
\usepackage{booktabs}
\usepackage{multirow}

\usepackage[pagebackref,breaklinks,colorlinks]{hyperref}

\usepackage[capitalize]{cleveref}
\crefname{section}{Sec.}{Secs.}
\Crefname{section}{Section}{Sections}
\Crefname{table}{Table}{Tables}
\crefname{table}{Tab.}{Tabs.}

\usepackage{color}

\def\cvprPaperID{2746} 
\def\confName{CVPR}
\def\confYear{2023}

\begin{document}

\title{High-fidelity 3D GAN Inversion by Pseudo-multi-view Optimization}

\author{Jiaxin Xie\thanks{Joint first authors} $^1$ \quad Hao Ouyang\footnotemark[1] $^1$ \quad  Jingtan Piao$^2$ \quad  Chenyang Lei$^3$  \quad  Qifeng Chen$^1$\\
$^1$HKUST  \qquad $^2$SenseTime Research \qquad $^3$ Centre for Artificial Intelligence and Robotics
}

\maketitle

\begin{figure*}[t]
    \centering
\begin{tabular}{@{}c@{\hspace{1mm}}c@{\hspace{1mm}}c@{\hspace{1mm}}c@{\hspace{1mm}}c@{\hspace{1mm}}c@{}}

\small{Input image}  & \small{Reconstruction}  & \small{Novel view 1} & \small{Novel view 2}  & \small{Novel view 3} \\

\includegraphics[width=0.200\linewidth]{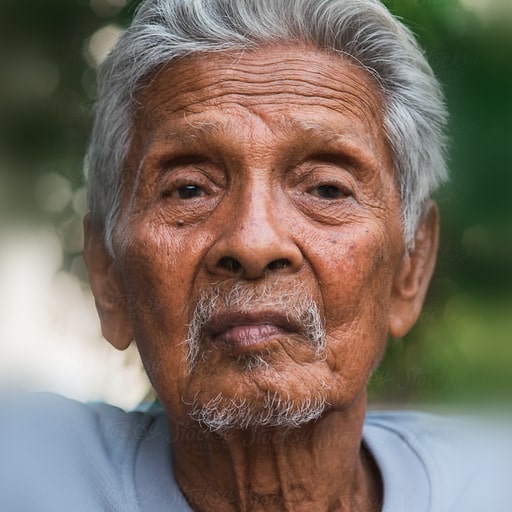}&
\includegraphics[width=0.200\linewidth]{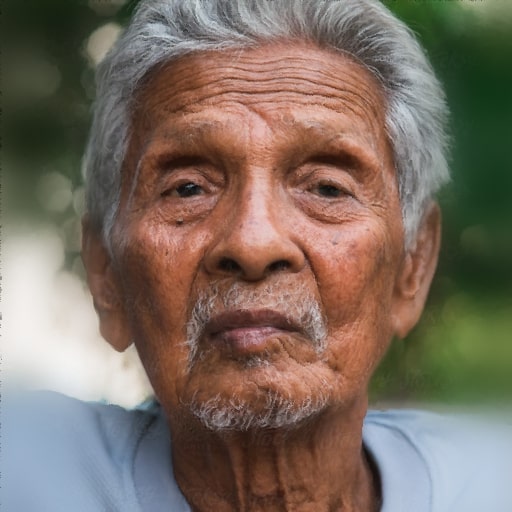}&
\includegraphics[width=0.200\linewidth]{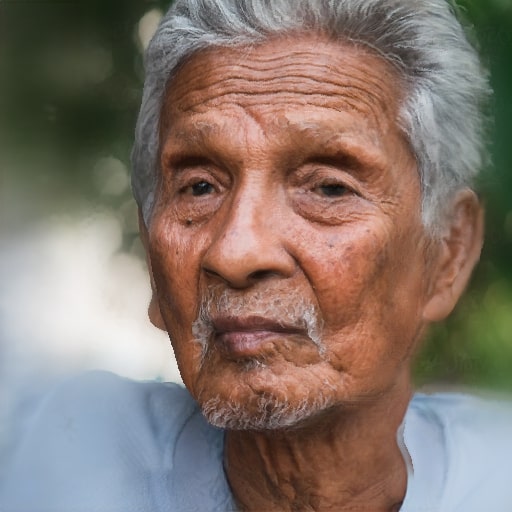}&
\includegraphics[width=0.200\linewidth]{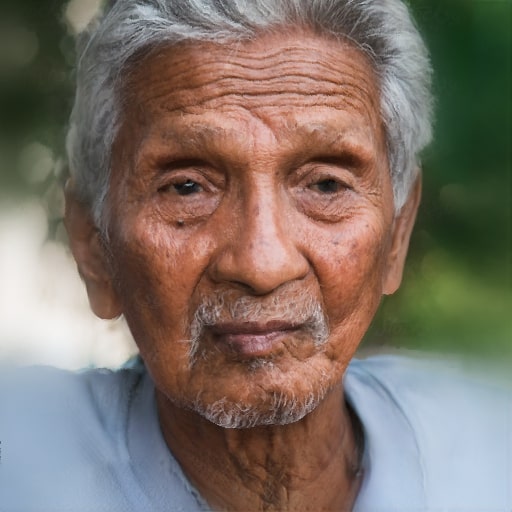}&
\includegraphics[width=0.200\linewidth]{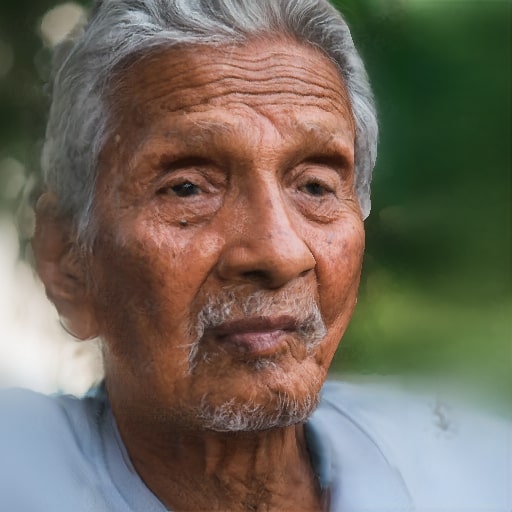}

\\
\includegraphics[width=0.200\linewidth]{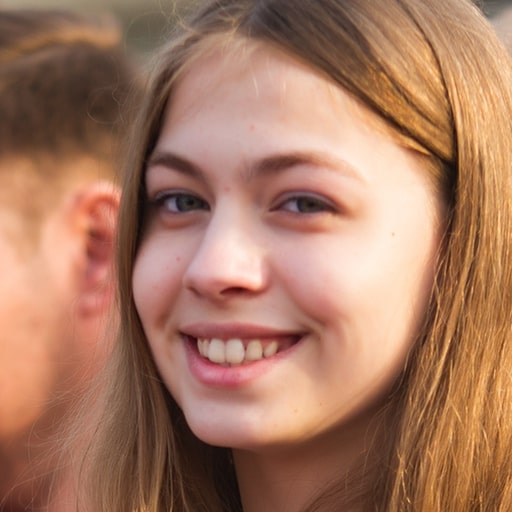}&
\includegraphics[width=0.200\linewidth]{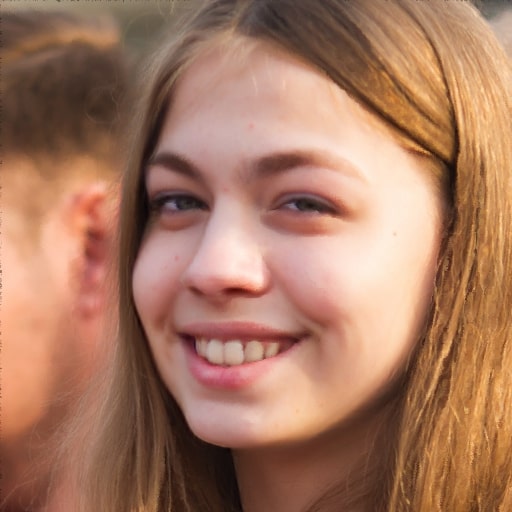}&
\includegraphics[width=0.200\linewidth]{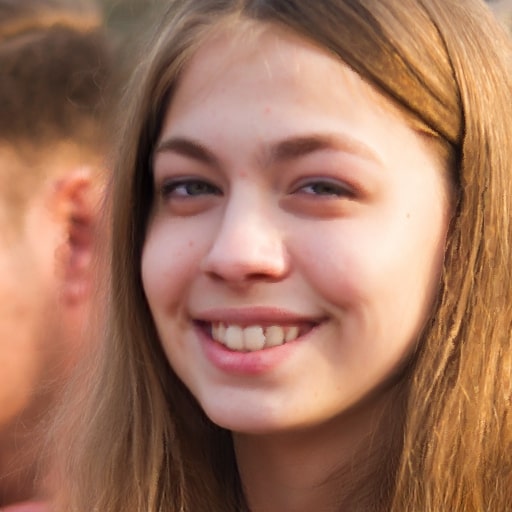}&
\includegraphics[width=0.200\linewidth]{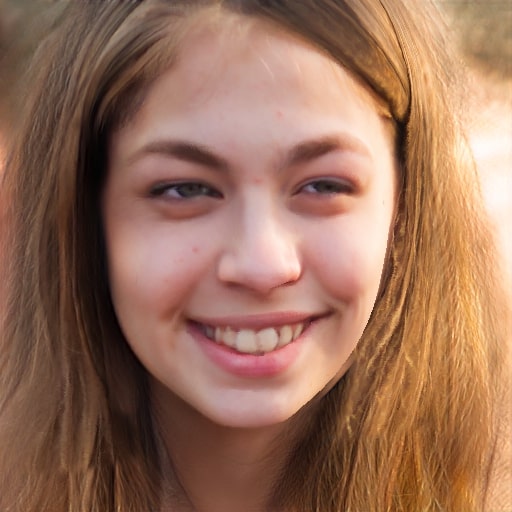}&
\includegraphics[width=0.200\linewidth]{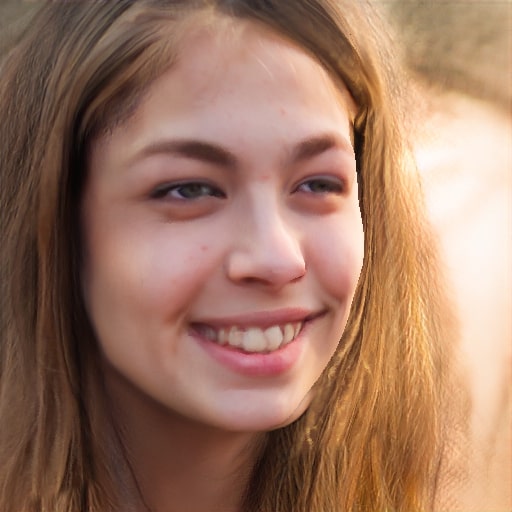}

\put(-340,-10){(a) High-fidelity 3D GAN inversion }
\\
\end{tabular}

\hspace*{-0.35cm}
\begin{tabular}{@{}c@{\hspace{0.5mm}}c@{\hspace{0.5mm}}c@{\hspace{0.5mm}}c@{\hspace{0.5mm}}c@{\hspace{0.5mm}}c@{\hspace{0.5mm}}c@{\hspace{0.5mm}}c@{\hspace{0.5mm}}c@{\hspace{0.5mm}}c@{\hspace{0.5mm}}c@{\hspace{0.5mm}}c@{}c@{}}

\rotatebox{90}{\small \hspace{3mm}Smile+}&
\includegraphics[width=0.100\linewidth]{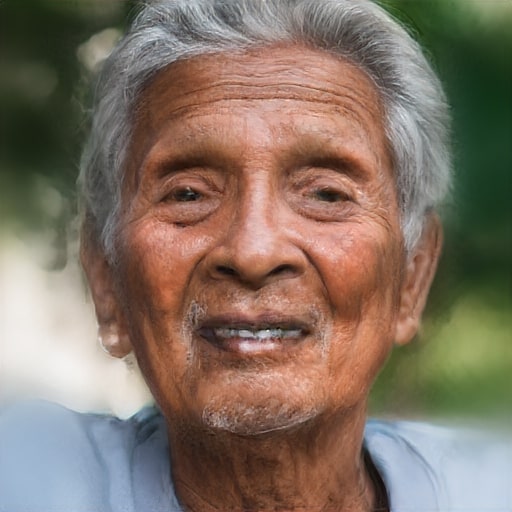}&
\includegraphics[width=0.100\linewidth]{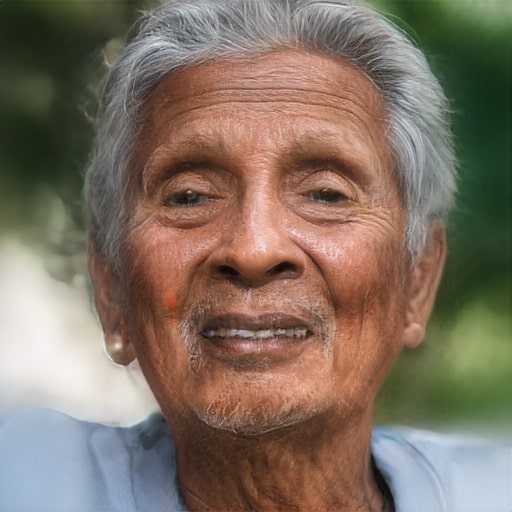}&
\includegraphics[width=0.100\linewidth]{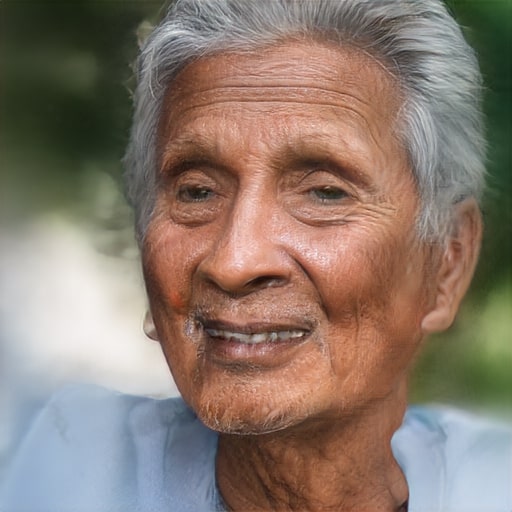}&
\includegraphics[width=0.100\linewidth]{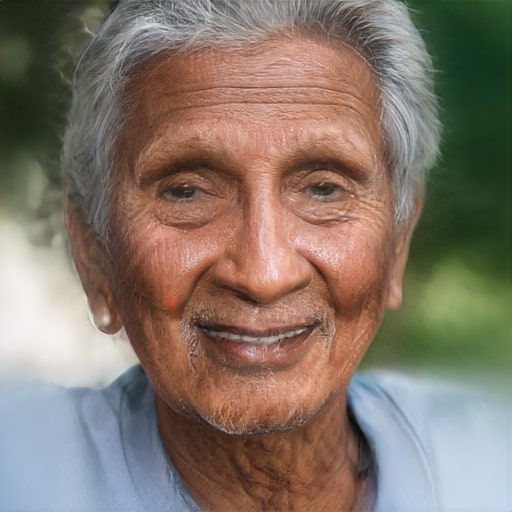}&
\includegraphics[width=0.100\linewidth]{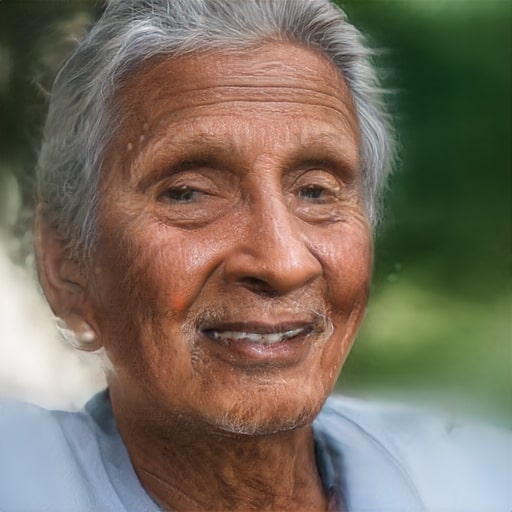}&
\includegraphics[width=0.100\linewidth]{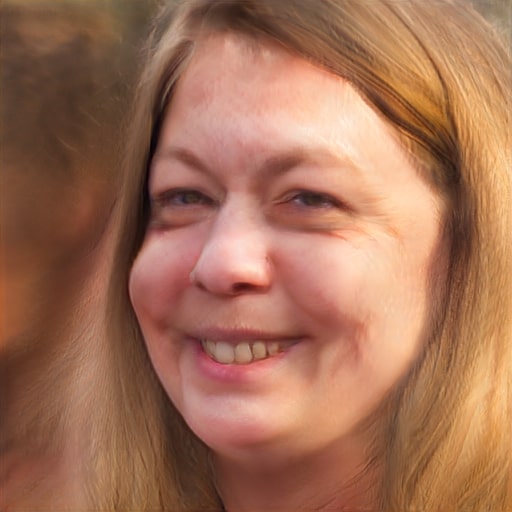}&
\includegraphics[width=0.100\linewidth]{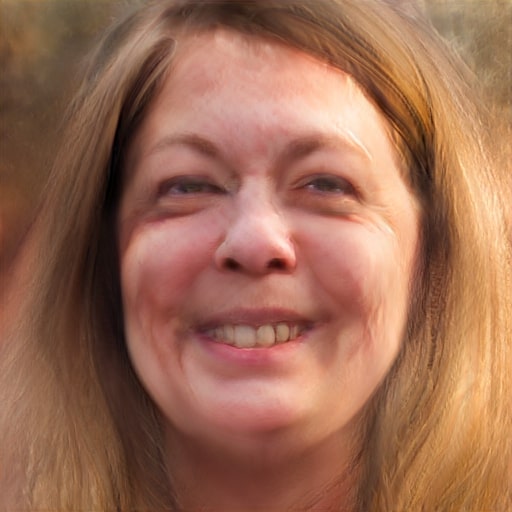}&
\includegraphics[width=0.100\linewidth]{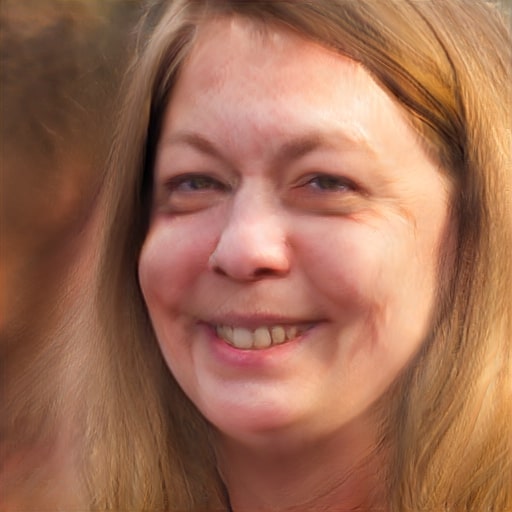}&
\includegraphics[width=0.100\linewidth]{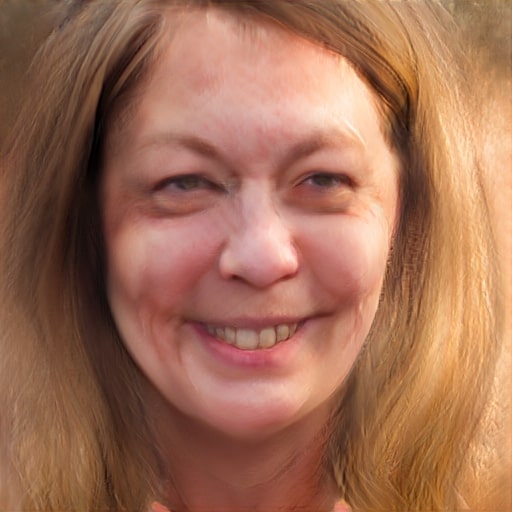}&
\includegraphics[width=0.100\linewidth]{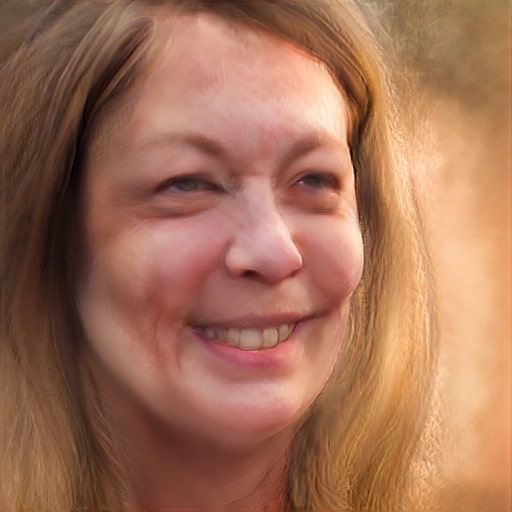}&
\rotatebox{90}{\small \hspace{3.5mm} Age+}
\put(-335,-10){(b) Latent Attributes Editing }
\put(-335,-15){ }
\\
&
\includegraphics[width=0.100\linewidth]{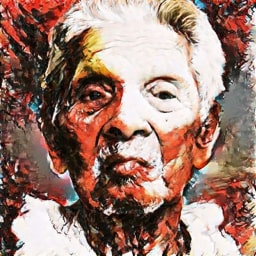}&
\includegraphics[trim={0 0 0.5cm 0.5cm},clip=true,width=0.100\linewidth]{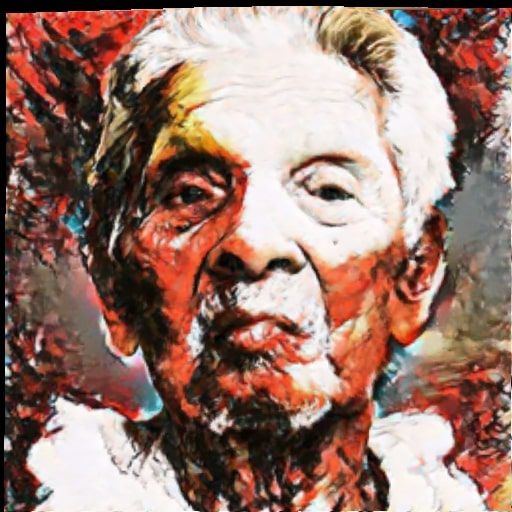}&
\includegraphics[trim={1.5cm 1.5cm 0 0},clip=true,width=0.100\linewidth]{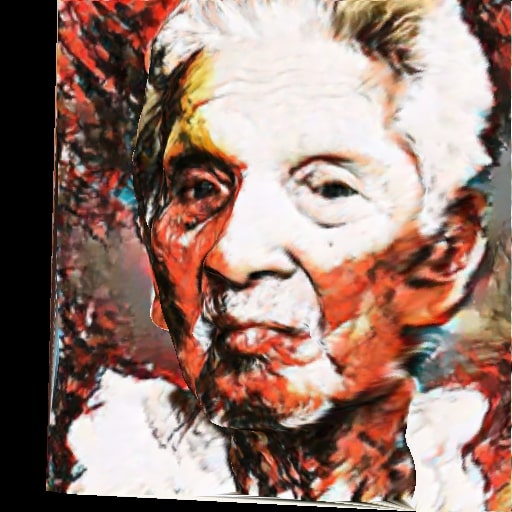}&
\includegraphics[trim={2cm 1.2cm 0 0.8cm},clip=true,width=0.100\linewidth]{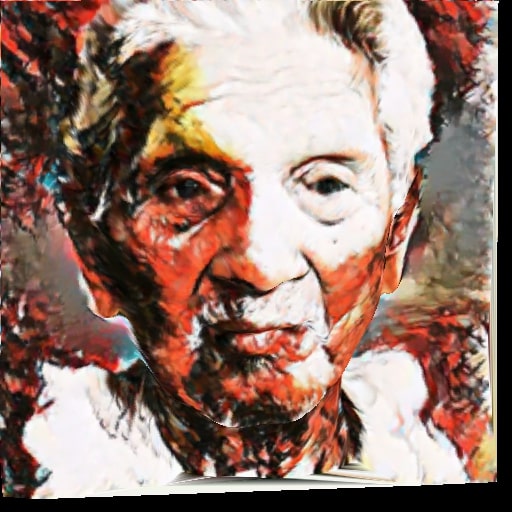}&
\includegraphics[trim={0cm 1.2cm 2.4cm 1.2cm},clip=true,width=0.100\linewidth]{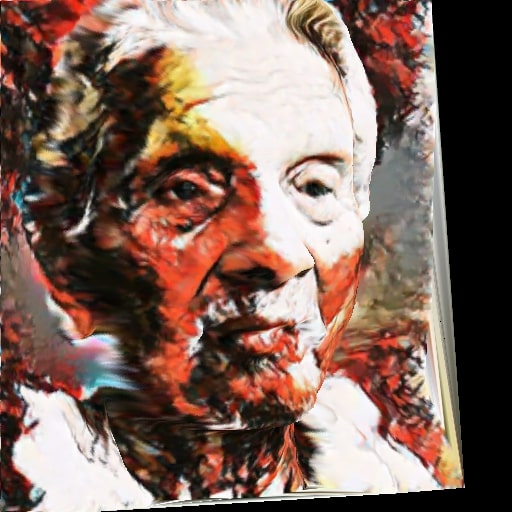}&
\includegraphics[width=0.100\linewidth]{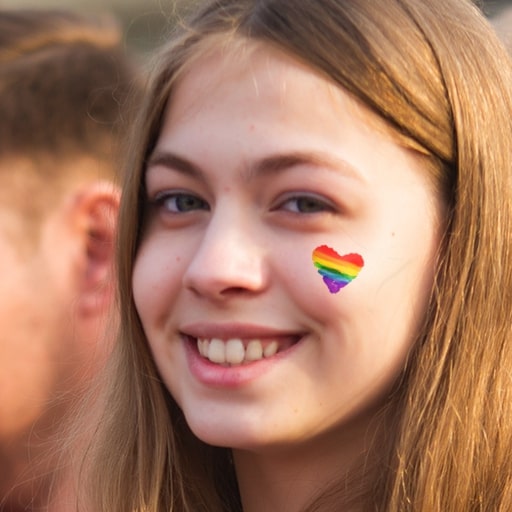}&
\includegraphics[width=0.100\linewidth]{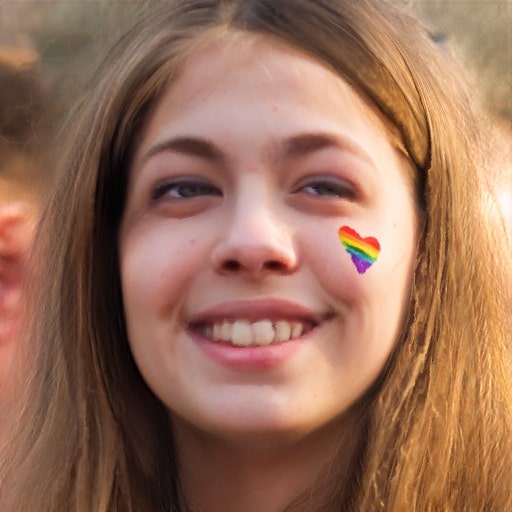}&
\includegraphics[width=0.100\linewidth]{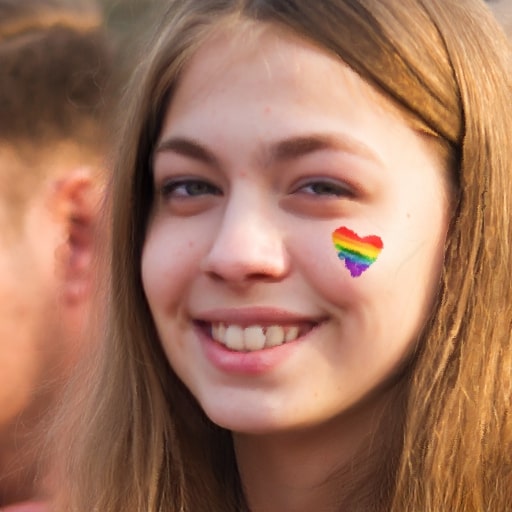}&
\includegraphics[width=0.100\linewidth]{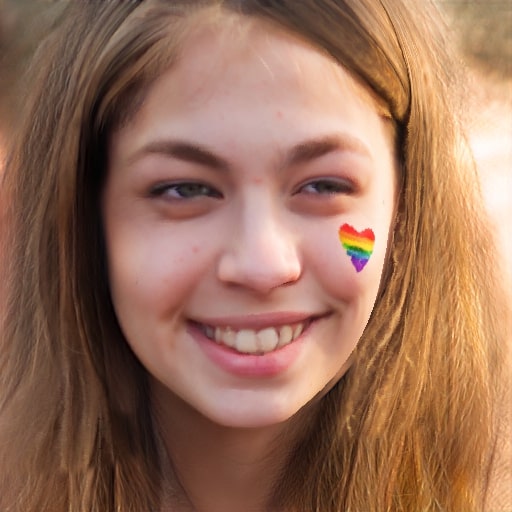}&
\includegraphics[width=0.100\linewidth]{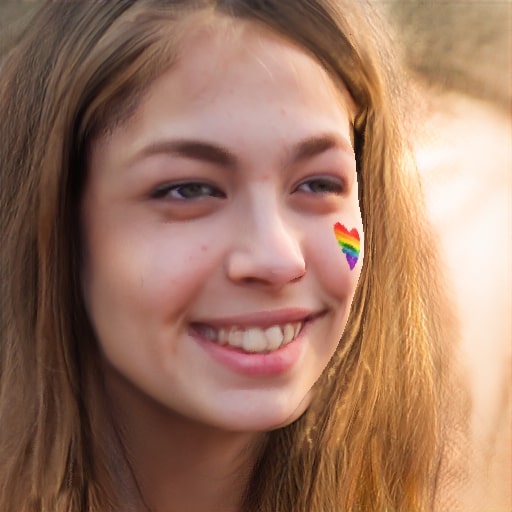} &
\put(-345,-10){(c) 3D-aware Textures Modification }
\\
\\


\end{tabular}
    \caption{High-fidelity 3D GAN inversion results on real-world images with two types of editing ability. Our method preserves compelling details and achieves high 3D consistency. 
   }
    \label{fig:teaser}
    \vspace{-1mm}
\end{figure*}

\begin{abstract}

   We present a high-fidelity 3D generative adversarial network (GAN) inversion framework that can synthesize photo-realistic novel views while preserving specific details of the input image. High-fidelity 3D GAN inversion is inherently  challenging due to the geometry-texture trade-off in 3D inversion, where overfitting to a single view input image often damages the estimated geometry during the latent optimization. To solve this challenge, we propose a novel pipeline that builds on the pseudo-multi-view estimation with visibility analysis. We keep the original textures for the visible parts and utilize generative priors for the occluded parts. Extensive experiments show that our approach achieves advantageous reconstruction and novel view synthesis quality over state-of-the-art methods, even for images with out-of-distribution textures. The proposed pipeline also enables image attribute editing with the inverted latent code and 3D-aware texture modification. Our approach enables high-fidelity 3D rendering from a single image, which is promising for various applications of AI-generated 3D content.  The source code is available \url{https://ken-ouyang.github.io/HFGI3D/index.html}.

\end{abstract}

\section{Introduction}
\label{sec:intro}

Real-world 3D-aware editing with \textit{a single 2D image} fascinates various essential applications in computer graphics, such as virtual reality (\textbf{VR}), augmented reality (\textbf{AR}), and immersive meetings. Recent advancement in 3D GANs~\cite{deng2022gram, chan2021pi, or2022stylesdf, chan2022efficient} has achieved photo-realistic 3D-consistent image generation. With the GAN inversion approaches~\cite{creswell2018inverting,perarnau2016invertible, xia2022gan}, which can map the images to the latent space of the pretrained 3D-aware model, high-fidelity 3D-aware editing becomes promising. 


High-fidelity 3D-aware inversion aims to generate novel views with high-quality reconstruction and 3D consistency, but existing methods can hardly meet these two goals simultaneously. Although current GAN inversion methods based on 2D GANs~\cite{abdal2019image2stylegan, roich2022pivotal, wang2022high, richardson2021encoding} can perform 3D-related attributes (e.g., head pose) editing, the generated view is inconsistent due to the lack of the underlying 3D representation. When applying the existing optimization-based inversion approaches on 3D-aware GANs~\cite{chan2022efficient, lin20223d}, we can retrieve high-fidelity reconstruction by overfitting to a single input image. However, different from 2D GAN inversion, the reconstruction quality of 3D GAN depends not only on the input view's faithfulness but also on the quality of the synthesized novel views. During the optimization process, the obvious artifacts in the synthesized novel views occur with the appearance of high-fidelity details in the input view as analyzed in Sec.~\ref{sec:method}. As only a single image is available in the optimization process, the reconstruction suffers from extreme ambiguity: infinite combinations of color and density can reconstruct the single input image, especially with out-of-distribution textures.



Based on the above observation, we propose our 3D-aware inversion pipeline by optimizing the reconstruction not only on the input image but also on a set of pseudo-multi-views. The pseudo views provide additional regularization, and thus the ambiguity is greatly reduced. Estimating the pseudo views is non-trivial as it should preserve the details of the textures with reasonable generation for the invisible parts from the input view. We first estimate an initial geometry and conduct a visibility analysis to solve these challenges. We directly utilize the textures from the input image for the visible parts to preserve the texture details. For the occlusion parts, we use a pretrained generator to synthesize the reasonable inpainted regions. With the additional supervision from the pseudo-multi-views, our approach achieves high-fidelity reconstruction results with the correct 3D geometry.  



Our approach enables two types of editing: latent attributes editing and 3D-aware texture modification. We follow the previous work~\cite{shen2020interpreting} and calculate the attribute direction in the latent code space. By modifying the inverted latent code in a specific direction, we can control the general attribute (e.g., smile, ages for portraits as in Figure~\ref{fig:teaser}(b)). Since the proposed pipeline enables inversion with out-of-distribution textures, we can achieve compelling 3D consistent editing by only modifying the textures of the input images (e.g., stylization or adding a  tattoo in Figure~\ref{fig:teaser}(c)). 


We perform extensive experiments, which indicate stable improvements over other GAN inversion methods. Our approach achieves 2.8 higher in PSNR and over 90 percent in user preference rate in the synthesized novel view videos. Our contribution can be summarized as follows:

\begin{itemize}
  \item We propose a high-fidelity 3D GAN inversion method by pseudo-multi-view optimization given an input image. Our approach can synthesize compelling 3D-consistent novel views that are visually and geometrically consistent with the input image.
  
  \item We demonstrate two types of editing ability: editing in the semantic attributes in latent space, such as age, gender, and expressions, and editing in image space, such as modifying textures and stylization. The camera pose of the edited image can be explicitly controlled.
  \item We conducted quantative and qualitive experiments, which demonstrate that our 3D GAN inversion approach outperforms other 2D/3D GAN inversion baselines in both photo-realism and faithfulness. 
\end{itemize}

\begin{figure*}[t]
\centering
\small
\hspace*{-0.1cm}
\begin{tabular}{@{}c@{\hspace{0.4mm}}c@{\hspace{0.4mm}}c@{\hspace{0.4mm}}c@{\hspace{0.4mm}}c@{\hspace{1.0mm} \unskip \vrule\hspace{1.0mm}}c@{}}
\multirow{2}[2]{0.16\linewidth}[13.2mm]
{
\includegraphics[width=\linewidth]{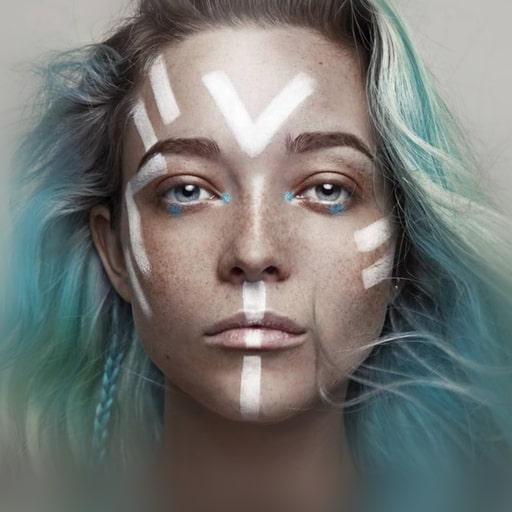}}&
\includegraphics[width=0.16\linewidth,height=0.16\linewidth]{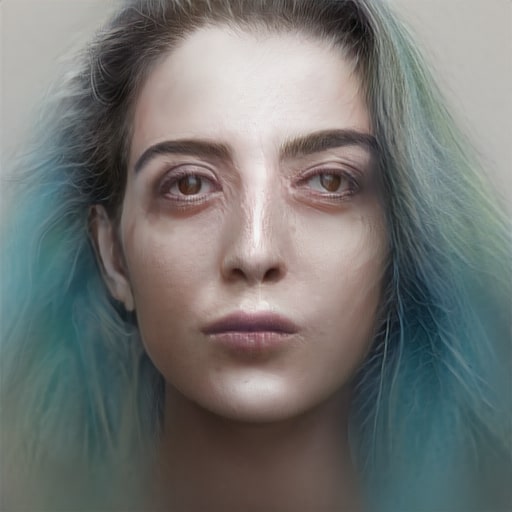}&
\includegraphics[width=0.16\linewidth,height=0.16\linewidth]{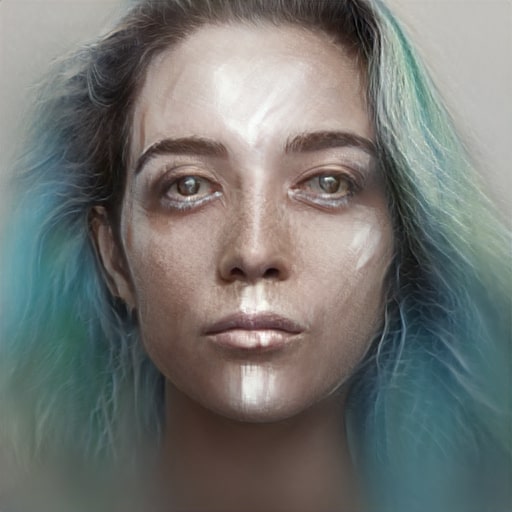}&
\includegraphics[width=0.16\linewidth,height=0.16\linewidth]{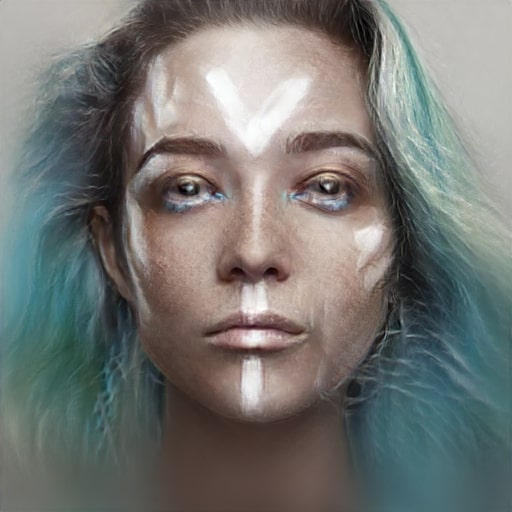}&
\includegraphics[width=0.16\linewidth,height=0.16\linewidth]{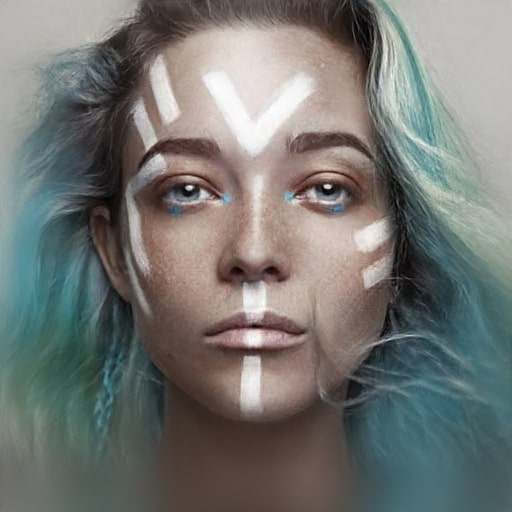}&
\includegraphics[width=0.16\linewidth,height=0.16\linewidth]{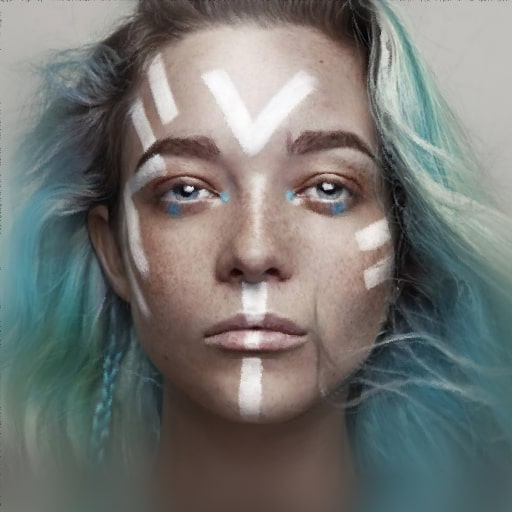}
\put(-402,85){$\rightarrow{}$----------------------------------- Reconstruction Quality -------------------------------- $\rightarrow{}$} 
\put(-470,36){Input Image}

\put(-50,85){Ours}
\\

&
\includegraphics[width=0.16\linewidth,height=0.16\linewidth]{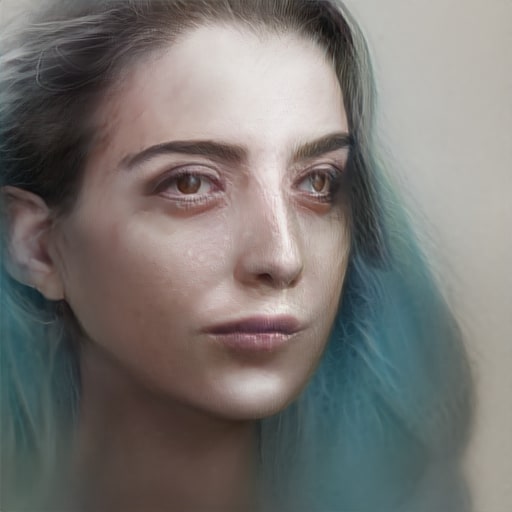}&
\includegraphics[width=0.16\linewidth,height=0.16\linewidth]{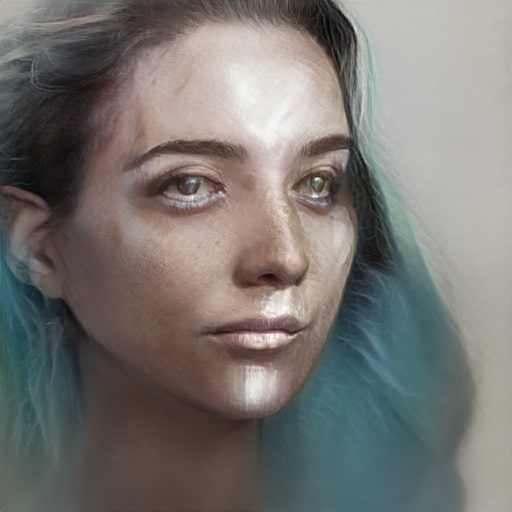}&
\includegraphics[width=0.16\linewidth,height=0.16\linewidth]{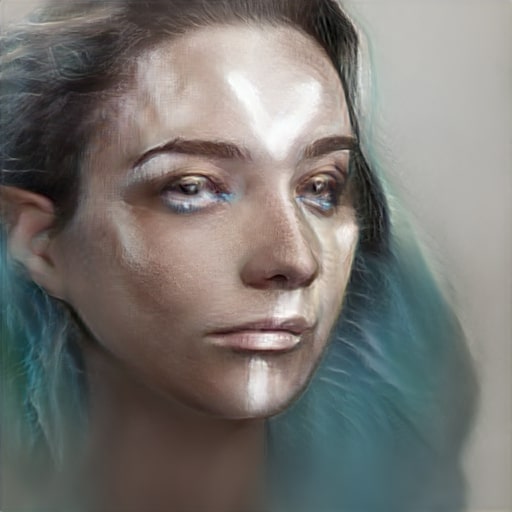}&
\includegraphics[width=0.16\linewidth,height=0.16\linewidth]{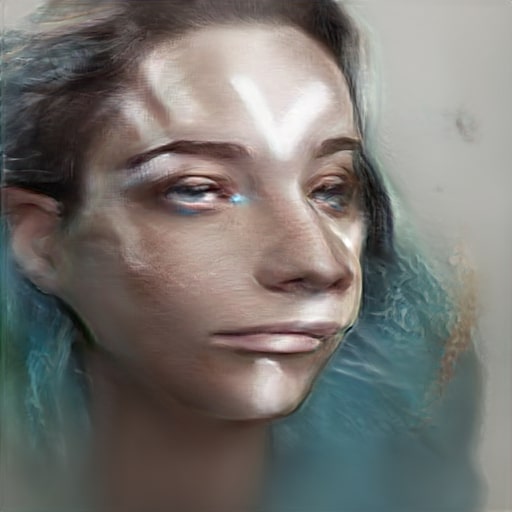}&
\includegraphics[width=0.16\linewidth,height=0.16\linewidth]{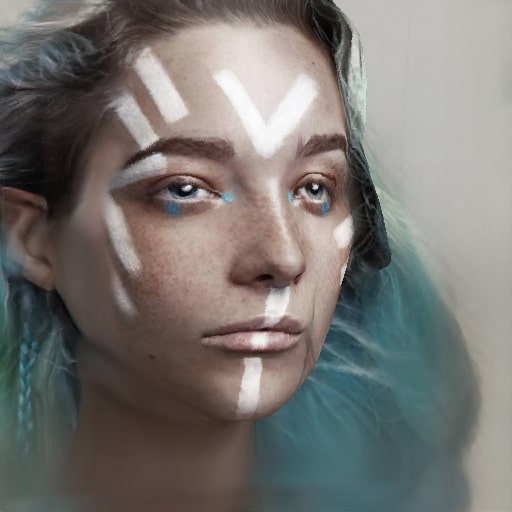} 

\put(-400,-10){$\rightarrow{}$----------------------------------- Novel View Quality ----------------------------------- $\rightarrow{}$} 
\put(-70,-10){Novel View(Ours)}
\\

\end{tabular}
    \caption{{Reconstruction quality vs. novel view quality during the optimization process. The reconstruction quality improves as the iteration increases, but the novel view quality decreases if we optimize the generator only using loss with the single input image. Our method achieves both highly preserved details with compelling novel view synthesis.}}
    \label{fig:analysis}
    \vspace{-1mm}
\end{figure*}

\section{Related Work}

\subsection{GAN Inversion}
GAN inversion~\cite{creswell2018inverting,perarnau2016invertible,zhu2016generative, xia2022gan} retrieves the latent code that can faithfully reconstruct the input image given a pretrained generative model. The recovered latent code facilitates various applications such as image editing~\cite{abdal2019image2stylegan,abdal2020image2stylegan++,shen2020interpreting, zhu2020domain}, interpolation~\cite{abdal2019image2stylegan,nitzan2020face}, and restoration~\cite{abdal2019image2stylegan,pan2021exploiting}. The existing inversion methods can be classified into three categories: optimization-based~\cite{abdal2019image2stylegan, abdal2020image2stylegan++, gu2020image, collins2020editing, huh2020transforming, wu2021stylespace, zhu2020improved}, encoder-based~\cite{richardson2021encoding, tov2021designing, alaluf2021restyle, wang2022high, dinh2022hyperinverter, bai2022high}, and hybrid~\cite{zhu2020domain,roich2022pivotal, chai2021ensembling, alaluf2022hyperstyle}. The optimization-based methods directly optimize a latent code by back-propagation to minimize the reconstruction objective, and the encoder-based approaches train an encoder to map from the image space to the latent space. A hybrid scheme combines the above methods by using the encoder for initialization and refining the latent code. Although recent 2D GAN inversion~\cite{wang2022high, alaluf2022third} has achieved faithful reconstruction with high editing capability, the editing related to 3D attributes (e.g., control camera angle and head pose) still suffers inevitable inconsistency and severe flickering as the pretrained generator is not 3D-aware. With the rapid development of 3D-aware GANs~\cite{chan2021pi, chan2022efficient}, 3D consistent editing becomes promising with GAN inversion techniques. Unlike the concept of fidelity of 2D GAN inversion, which considers only the pixel-wise difference of the input view, the 3D inversion additionally involves the quality of the synthesized novel views. With the optimization-based inversion scheme, even though the details in the input view are well-preserved, the overall fidelity can still be low as the synthesized view contains severe artifacts because of the issues such as entangled texture and geometry. This paper focuses on solving these issues and achieving high-fidelity 3D-aware reconstruction.      

\subsection{3D-aware GANs}
Geometrically consistent GANs~\cite{goodfellow2014generative} have recently become a trending research topic. Early works explore generating 3D consistent images with different 3D representations such as mesh~\cite{kanazawa2018learning,chen2019learning,goel2020shape,henderson2020leveraging,mustikovela2020self}, voxels~\cite{henzler2019platonicgan,tulsiani2017learning, nguyen2019hologan, nguyen2020blockgan}, multi-plane images~\cite{zhao2022generative, ouyang2022real} and point clouds~\cite{tatarchenko2016multi}. The generated images usually suffer from blurry details since the 3D resolution is low, considering the memory cost. Although a learned neural rendering module~\cite{nguyen2019hologan, nguyen2020blockgan} can increase the generation quality, it damages the view consistency and results in inconsistent novel views. The neural 3D representation~\cite{park2019deepsdf,michalkiewicz2019implicit,atzmon2019sal,gropp2020implicit,sitzmann2019srns,peng2020convolutional, mescheder2019occupancy,chen2019learning,mildenhall2020nerf, wang2021neus} (Neural Radiance Fields~\cite{mildenhall2020nerf} especially) achieves stunning photorealism for novel view synthesis and can serve as the underlying representation for the 3D-aware generation. The recently proposed 3D-aware GANs~\cite{deng2022gram, chan2021pi, or2022stylesdf, zhou2021cips, chan2022efficient, niemeyer2021giraffe} rely on implicit 3D representations and render high-resolution outputs with impressive details and excellent 3D consistency. Our works adopt EG3D~\cite{chan2022efficient} as the pretrained architecture as it generates photorealistic 3D consistent images on par with StyleGan~\cite{karras2019style,Karras2020stylegan2} with high computational efficiency. Note that the proposed inversion pipeline can be easily adapted to other 3D-aware GANs by replacing the underlying models from triplanes with other 3D representations. 

Researchers have started early exploration in 3D GAN inversion~\cite{lin20223d, sun2022ide, wang2022narrate, ko20223d}. LIN \textit{et al.}~\cite{lin20223d} explores the latent space of EG3D and enables animating a portrait with a single image. IDE-3D~\cite{sun2022ide} re-trains EG3D in a semantic-aware way and achieves semantic-conditioned editing with a hybrid inversion scheme. Note that NARRATE~\cite{wang2022narrate} utilizes inversion to retrieve the 3D shapes, but their goals are to estimate the normal for novel view portrait editing. Different from these works, which mainly focus on the downstream applications of 3D GAN inversion, our work aims to improve the faithfulness of the 3D inversion while preserving the editing ability. 

\section{Method}
\label{sec:method} 

\begin{figure*}[t]
    \centering
    \includegraphics[width=1.0\linewidth]{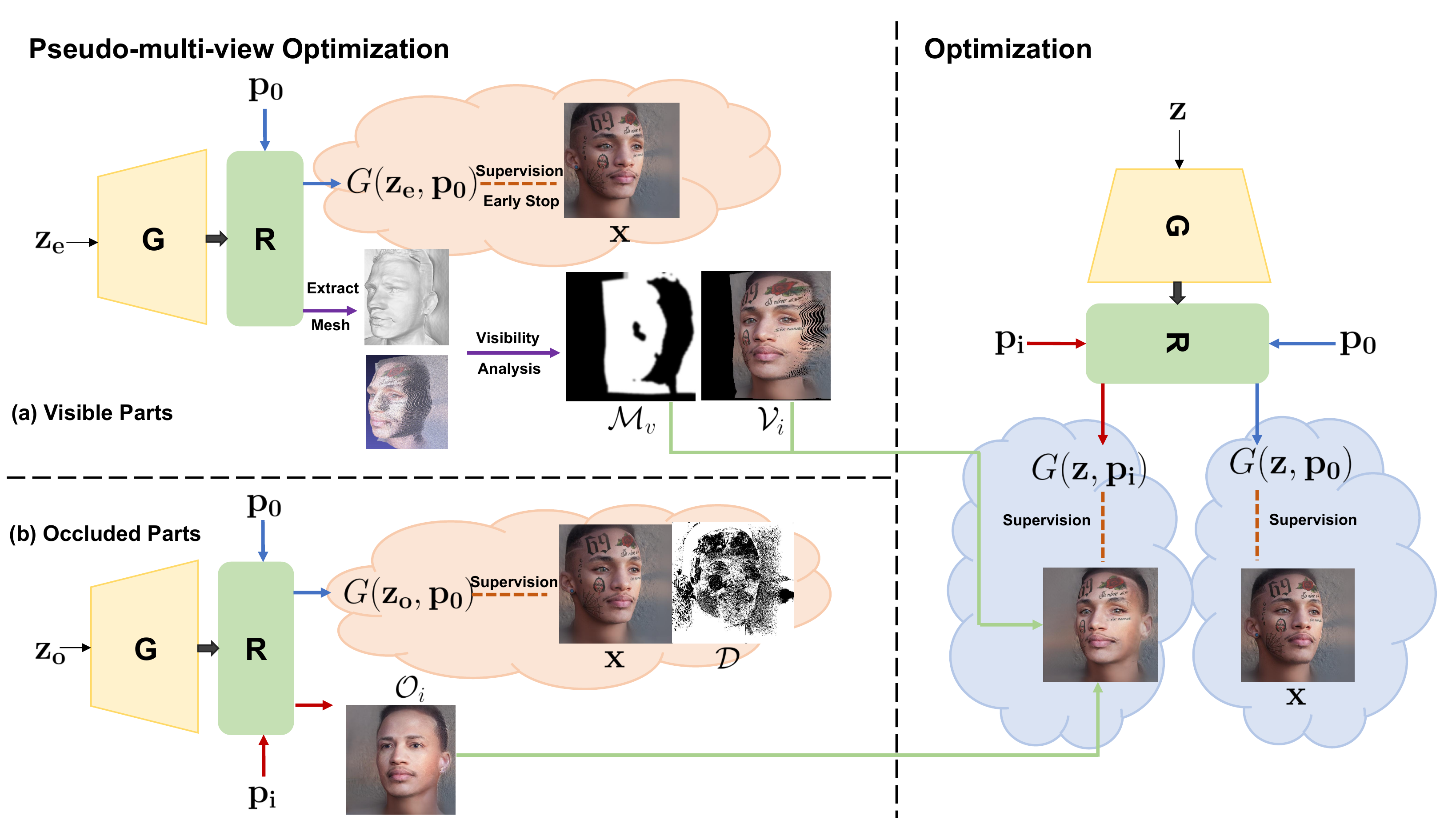}
    \caption{The pipeline of the proposed 3D-aware inversion framework. $G$ and $R$ are the pretrained 3D-aware generator and the corresponding renderer, respectively. To improve the inversion quality, we synthesize the pseudo-view with camera pose $p_i$ to regularize the optimization. The pseudo-multiview estimation consists of two steps: the visible image $\mathcal{V}_i$ warping (upper-left) and the occlusion image $\mathcal{O}_i$  generation (lower-left), combined with the estimated visibility mask $\mathcal{M}_v$. With the additional supervision of the pseudo-views, the pipeline achieves high-fidelity 3D-consistent inversion.}
    \label{fig:method}
    \vspace{-1mm}
\end{figure*} 

\subsection{Overview}
\noindent\textbf{3D GAN inversion.}
Given the generator $G$ of a pretrained GAN that maps from latent space $\mathcal{Z}$ to image space $\mathcal{X}$, GAN inversion aims at mapping from $\mathcal{X}$ back to $\mathcal{Z}$, where the latent $\mathbf{z}$ can faithfully reconstruct the input image $\mathbf{x}$. As the 3D-aware GANs involve the physics-based rendering stage, the additional camera pose $\mathbf{p}$ is involved in synthesizing 3D consistent images. Formally, we formulate the 3D GAN inversion problem as follows:
\begin{equation}
\mathbf{z}^* = \mathop{\arg\min}\limits_{\mathbf{z}} \mathcal{L}(G(\mathbf{z}, \mathbf{p}), \mathbf{x}),
\end{equation}
where $\mathcal{L}(\cdot)$ is the optimization loss that represents image distance in image or feature spaces.

\vspace{1mm}
\noindent \textbf{Analysis.} Researchers have conducted some attempts in 3D GAN inversion by directly applying optimization or hybrid methods for portrait reconstruction. Although the general facial features of the input images can be reconstructed with these approaches, the image-specific details, such as the freckles, wrinkles, and tattoos, are lost, and the fidelity is thus degraded. We observe that the optimization procedure in these methods is ``early stopped" at a certain iteration before the input image $\mathbf{x}$ is ``overfitted."

However, if we trivially increase the max optimization iteration, as shown in Figure~\ref{fig:analysis},  while the inversion preserves more image-specific details in the input image, obvious artifacts appear in the synthesized novel views. We further find that as the optimization iterations increase, although the reconstruction quality of the input image increases, the 3D consistency, on the contrary, decreases(more quantitative results are shown in the \textit{supplement}).  Our experiments demonstrate that ``overfitting'' to the input image that considers only the input camera pose damages the geometry of 3D GAN inversion. As the latent code is optimized to generate the image-specific textures, the corresponding geometry goes out of the distribution and leads to visually-unpleasant novel view results. To achieve high-fidelity inversion with high-quality novel view synthesis, we need a more delicate optimization scheme to handle the texture-geometry trade-off.

\vspace{1mm}
\noindent \textbf{Design.} 
As analyzed above, even though the optimized latent $\mathbf{z}$ can render an image similar to the input image $\mathbf{x}$ given the input camera pose $\mathbf{p_0}$,  $\mathbf{z}$ does not guarantee the reasonable rendering under other camera poses.  In this work, we improve the inversion performance by regularizing the outputs of additional camera poses $\{\mathbf{p_1}, \mathbf{p_2} ... \mathbf{p_n}\}$ with the corresponding pseudo-multi-views $\{\mathbf{v_1}, \mathbf{v_2} ... \mathbf{v_n}\}$. To this end, we can reformulate the objective of the 3D GAN inversion as follows:
\begin{equation}
\mathbf{z}^* = \mathop{\arg\min}\limits_{\mathbf{z}} {\mathcal{L}(G(\mathbf{z}, \mathbf{p_0}), \mathbf{x}) + \alpha \sum_{i}^{n} \mathcal{L}(G(\mathbf{z}, \mathbf{p_{i}}), \mathbf{v_i})},
\end{equation}
where $\alpha$ controls the strength of the regularization from the pseudo-views. For a pseudo-view $\mathbf{x_i}$ with camera pose $\mathbf{p_i}$, to increase the quality of the inversion, we should have the following properties: (1) if the texture from $\mathbf{x}$ is visible under the camera pose $\mathbf{p_i}$, the texture should be preserved; (2) the occluded parts should reasonably inpainted.  Next, we will describe in detail how to synthesize the reasonable pseudo-views that satisfy the above properties.

Note that we have explored other alternative regularization strategies, such as regularizing the density or keeping the coarse geometry unchanged during the optimization while the reconstruction loss is only calculated on a single input, and the details are provided in the \textit{supplement}.   


\begin{figure*}[t]
\centering
\small
\begin{tabular}{@{}c@{\hspace{1mm}}c@{\hspace{1mm}}c@{\hspace{1mm}}c@{\hspace{1mm}}c@{\hspace{1mm}}c@{}}

\multirow{2}[2]{0.16\linewidth}[10.2mm]
{\includegraphics[width=\linewidth]{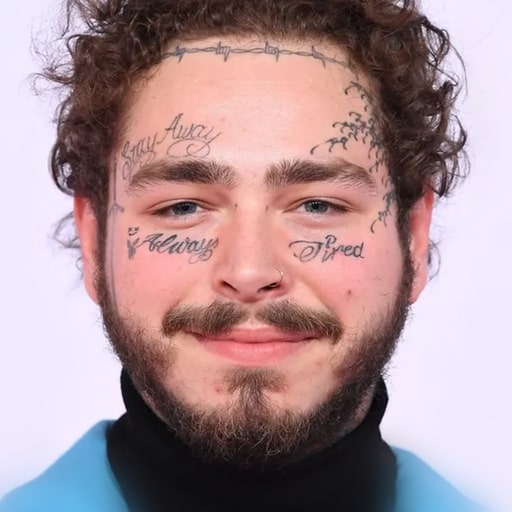}}&
\includegraphics[width=0.16\linewidth,height=0.16\linewidth]{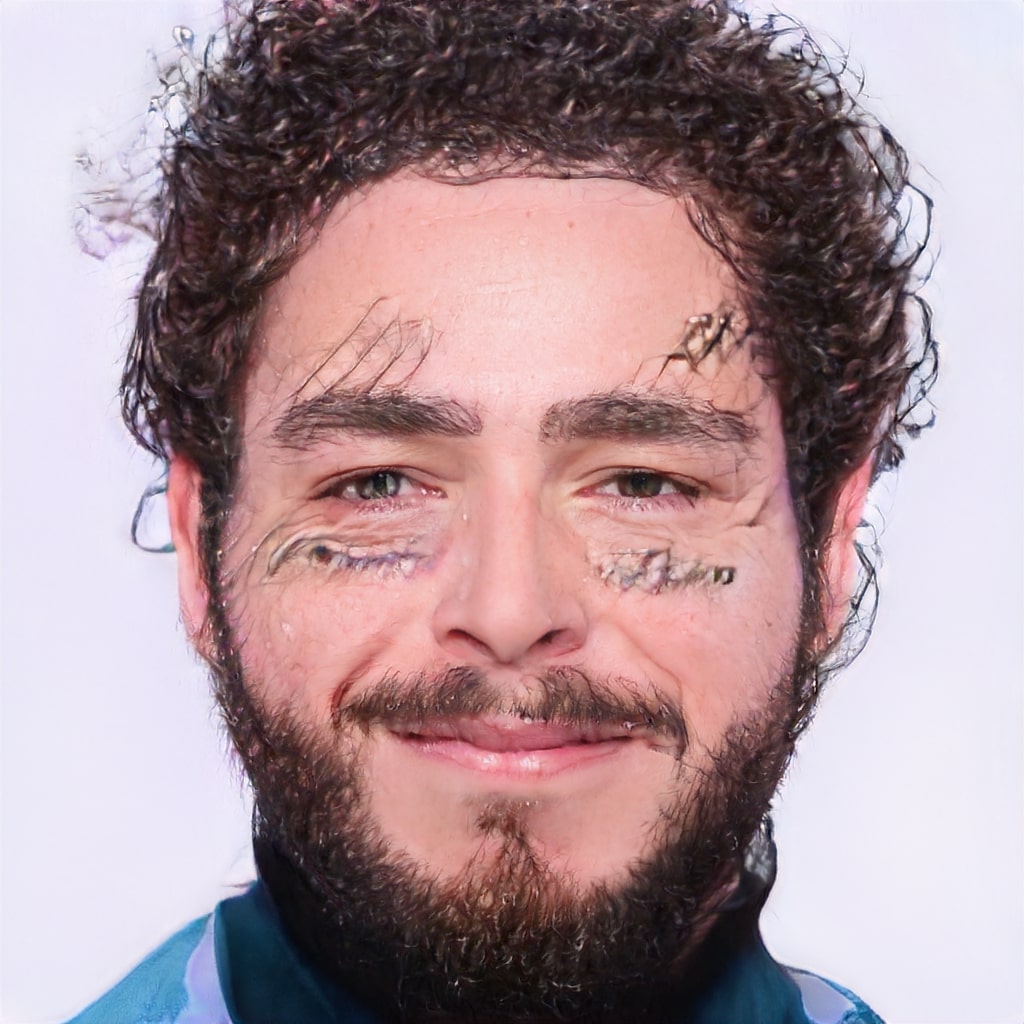}&
\includegraphics[width=0.16\linewidth,height=0.16\linewidth]{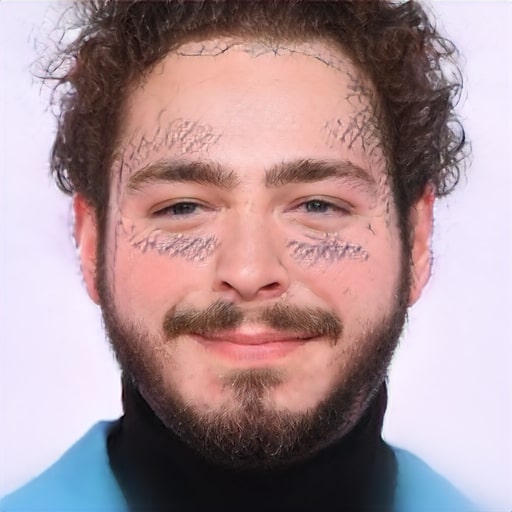}&
\includegraphics[width=0.16\linewidth,height=0.16\linewidth]{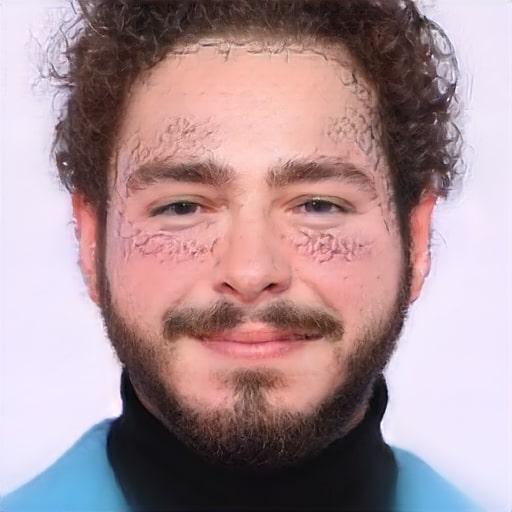}&
\includegraphics[width=0.16\linewidth,height=0.16\linewidth]{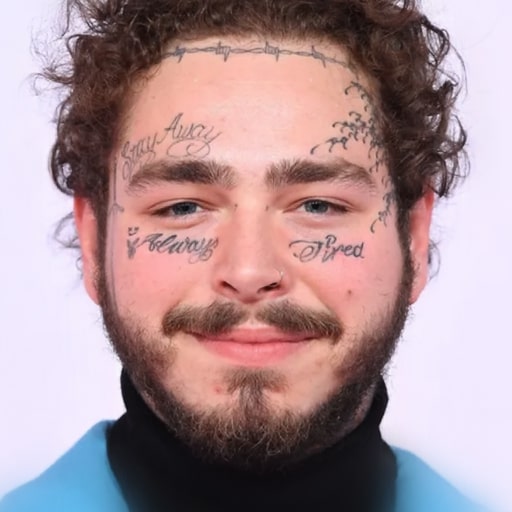}&
\rotatebox{90}{\small \hspace{3.5mm} Reconstruction}\\

&
\includegraphics[width=0.16\linewidth,height=0.16\linewidth]{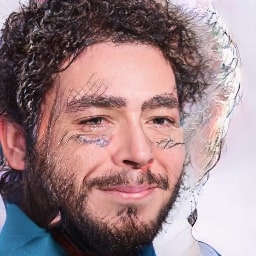}&
\includegraphics[width=0.16\linewidth,height=0.16\linewidth]{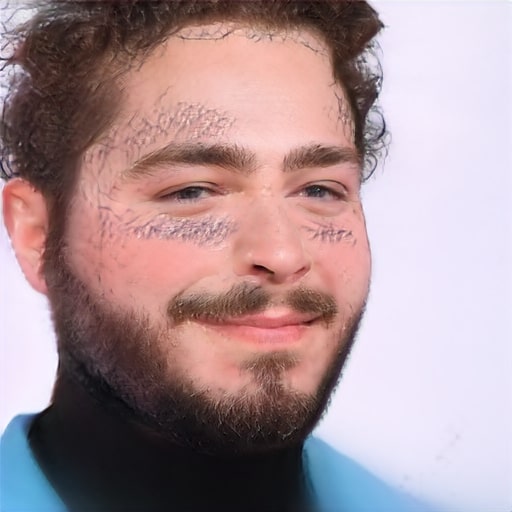}&
\includegraphics[width=0.16\linewidth,height=0.16\linewidth]{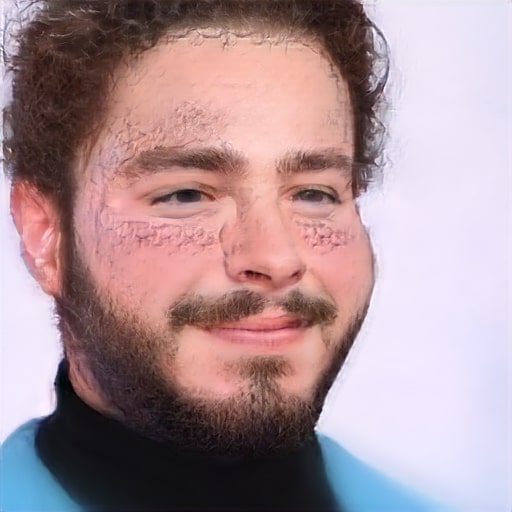}&
\includegraphics[width=0.16\linewidth,height=0.16\linewidth]{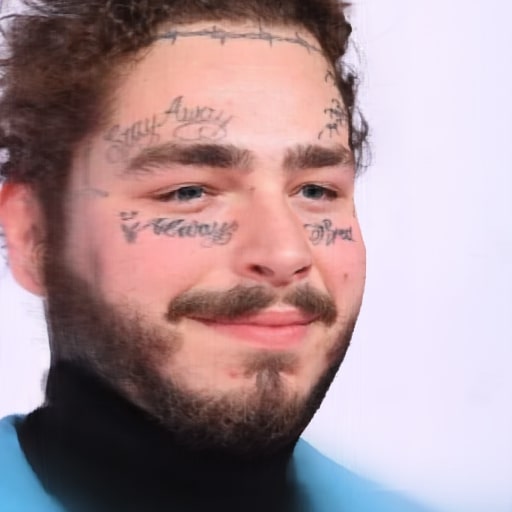}&
\rotatebox{90}{\small \hspace{5mm} Novel View}
\\

\multirow{2}[2]{0.16\linewidth}[10.2mm]
{\includegraphics[width=\linewidth]{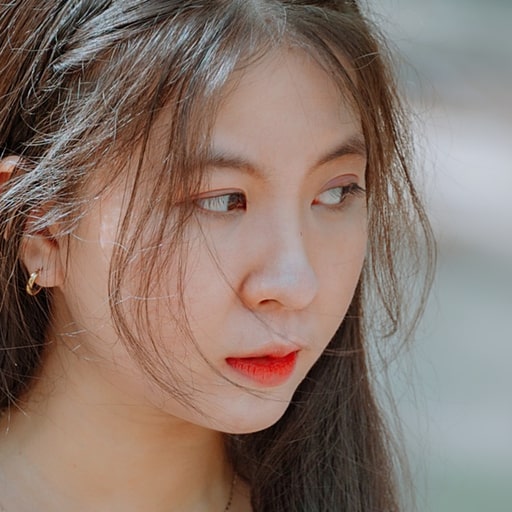}}&
\includegraphics[width=0.16\linewidth,height=0.16\linewidth]{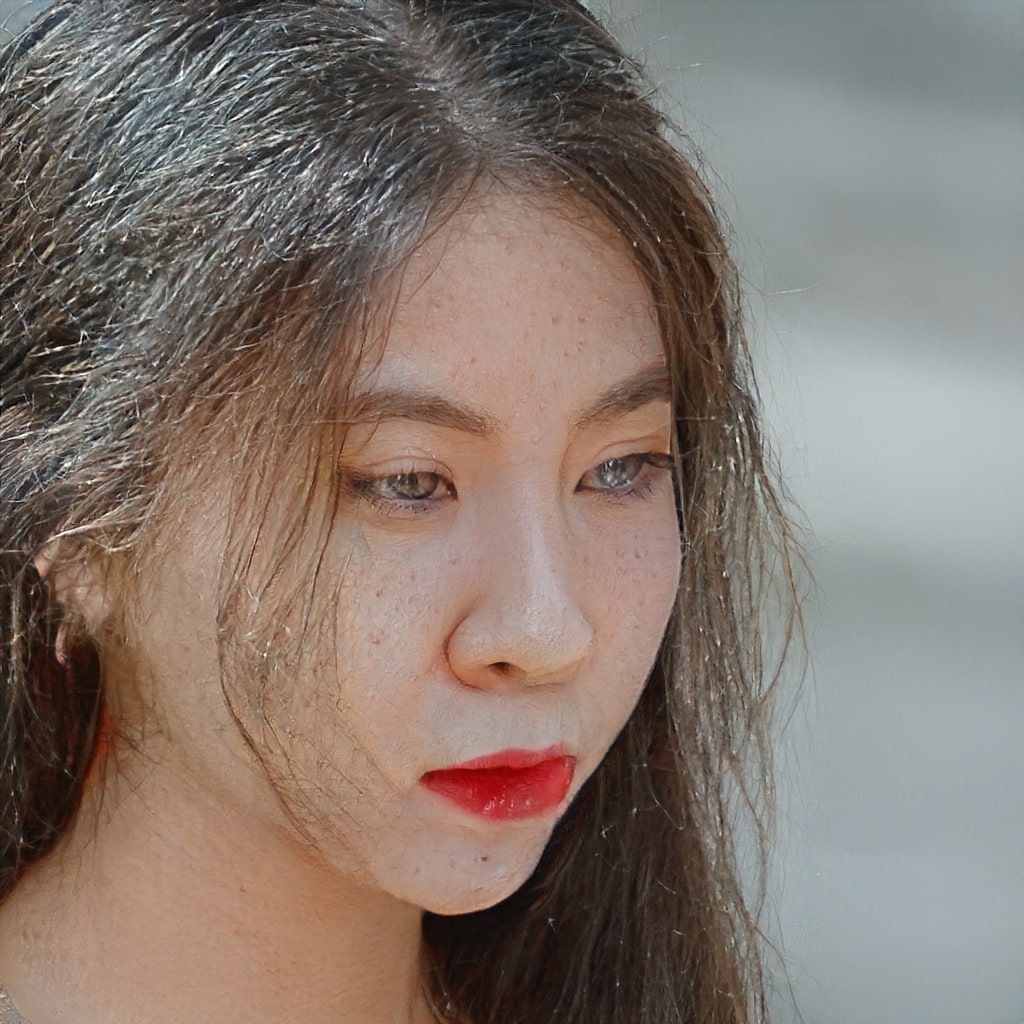}&
\includegraphics[width=0.16\linewidth,height=0.16\linewidth]{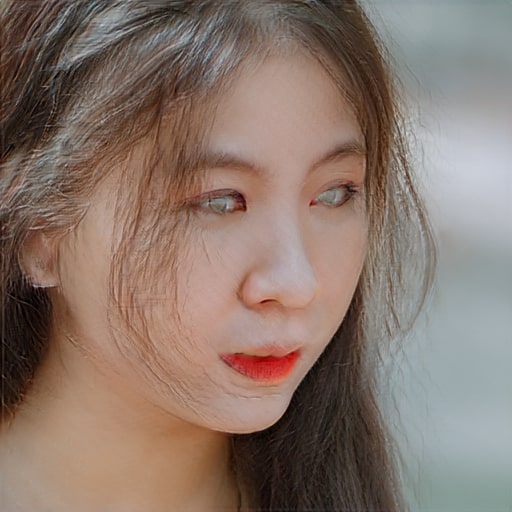}&
\includegraphics[width=0.16\linewidth,height=0.16\linewidth]{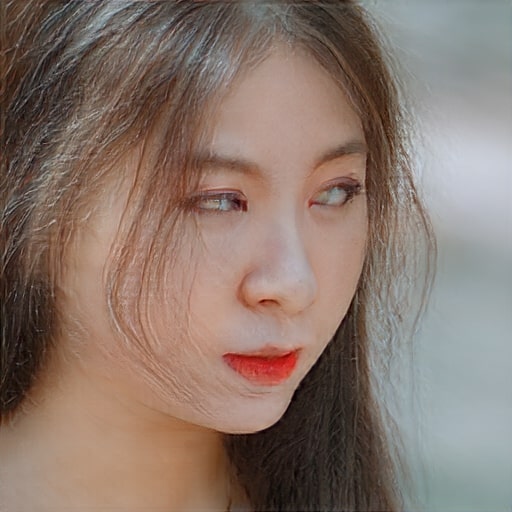}&
\includegraphics[width=0.16\linewidth,height=0.16\linewidth]{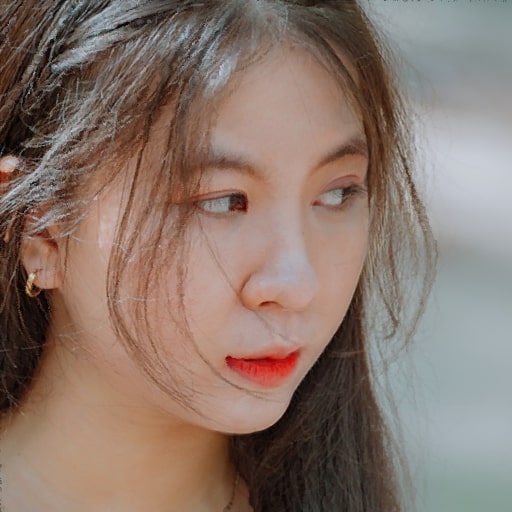}&
\rotatebox{90}{\small \hspace{3.5mm} Reconstruction}\\

&
\includegraphics[width=0.16\linewidth,height=0.16\linewidth]{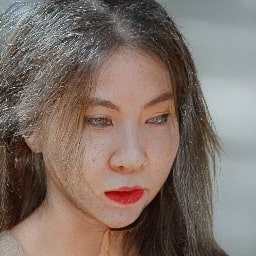}&
\includegraphics[width=0.16\linewidth,height=0.16\linewidth]{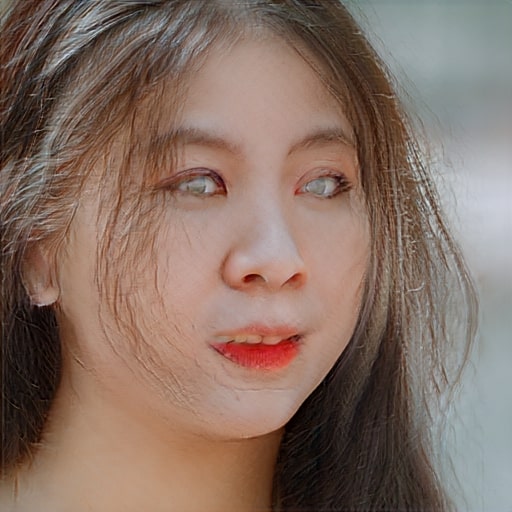}&
\includegraphics[width=0.16\linewidth,height=0.16\linewidth]{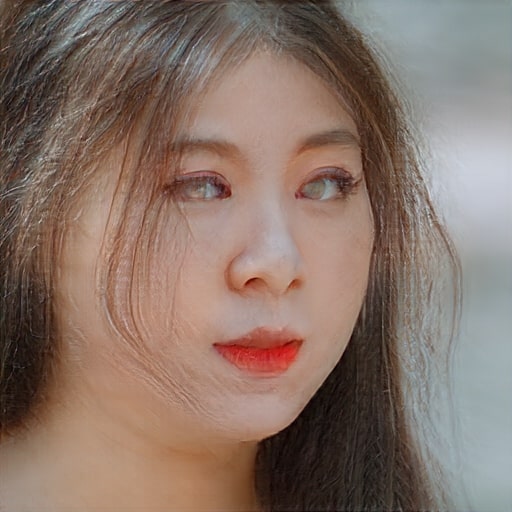}&
\includegraphics[width=0.16\linewidth,height=0.16\linewidth]{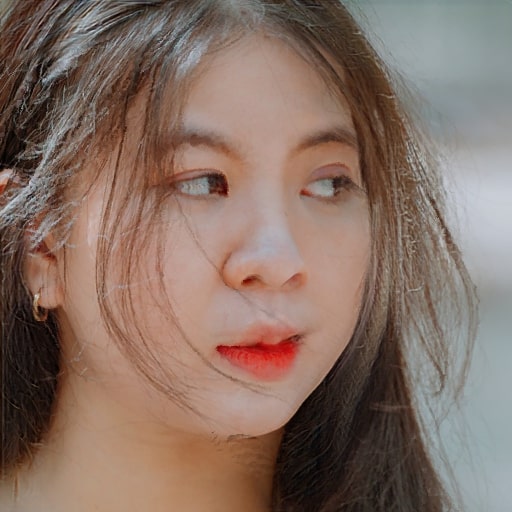}&
\rotatebox{90}{\small \hspace{5mm} Novel View}\\

\multirow{2}[2]{0.16\linewidth}[10.2mm]
{\includegraphics[width=\linewidth]{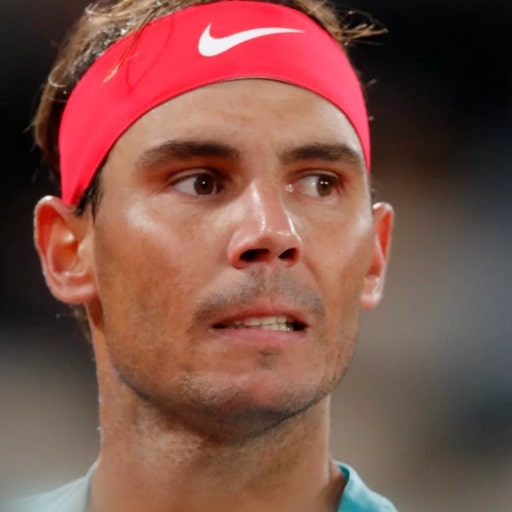}}&
\includegraphics[width=0.16\linewidth,height=0.16\linewidth]{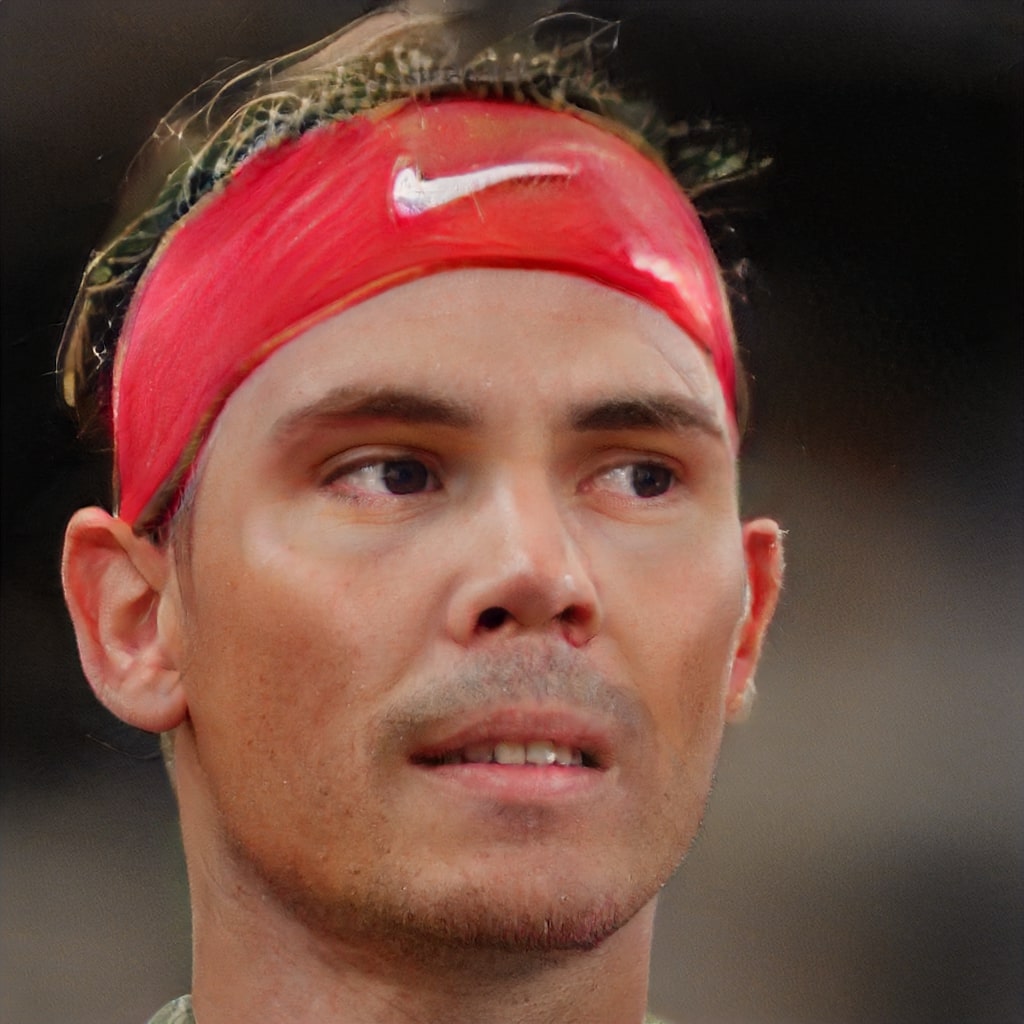}&
\includegraphics[width=0.16\linewidth,height=0.16\linewidth]{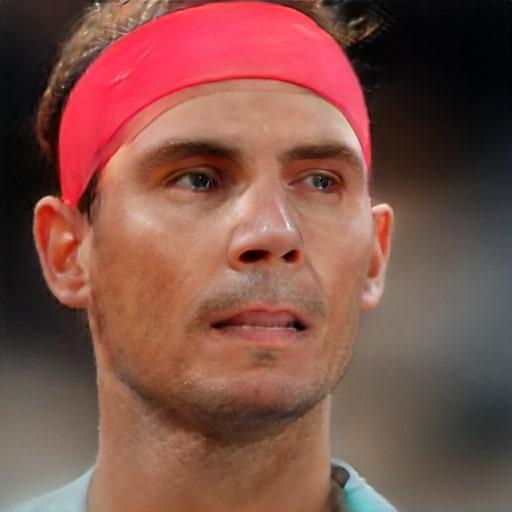}&
\includegraphics[width=0.16\linewidth,height=0.16\linewidth]{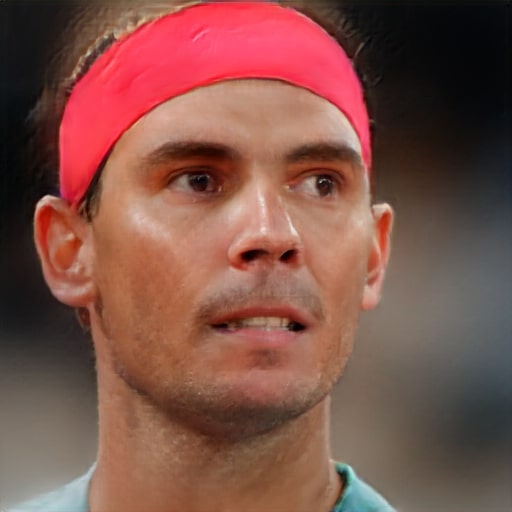}&
\includegraphics[width=0.16\linewidth,height=0.16\linewidth]{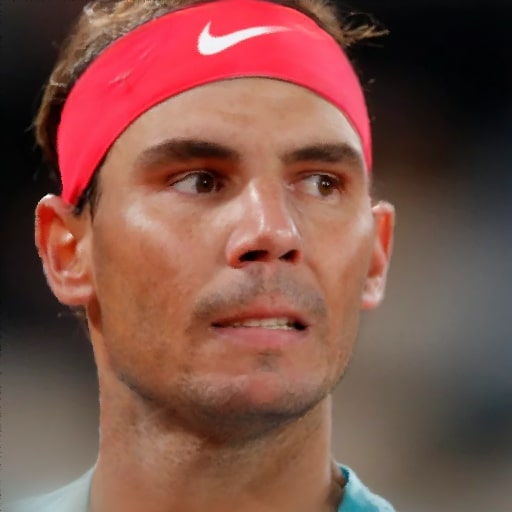}&
\rotatebox{90}{\small \hspace{3.5mm} Reconstruction}\\

&
\includegraphics[width=0.16\linewidth,height=0.16\linewidth]{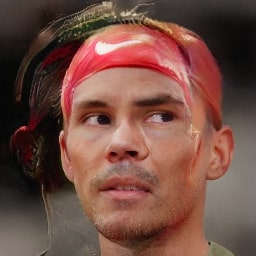}&
\includegraphics[width=0.16\linewidth,height=0.16\linewidth]{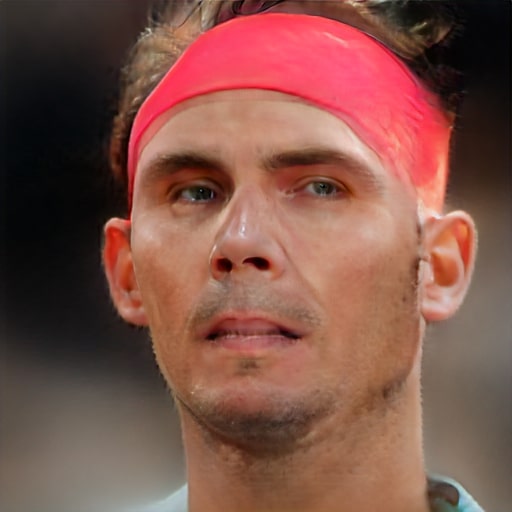}&
\includegraphics[width=0.16\linewidth,height=0.16\linewidth]{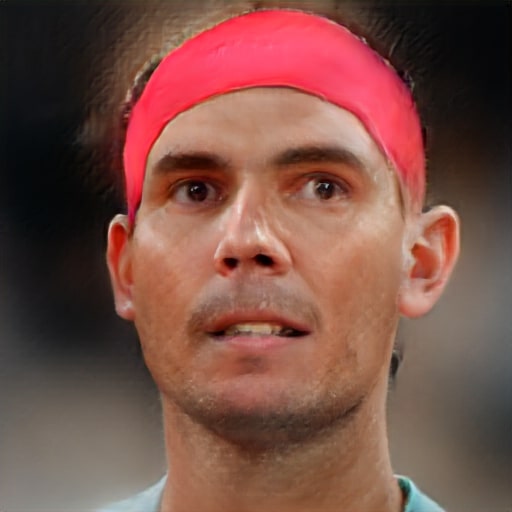}&
\includegraphics[width=0.16\linewidth,height=0.16\linewidth]{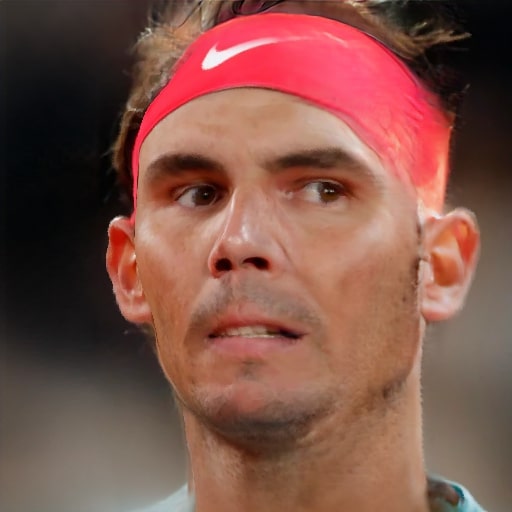}&
\rotatebox{90}{\small \hspace{5mm} Novel View}\\

 Input image & HFGI~\cite{wang2022high} & PTI~\cite{roich2022pivotal} &IDE-3D~\cite{sun2022ide} &Ours\\ 
\end{tabular}
    \caption{{\textbf{Qualitative comparison with baselines.} More results are attached in the \textit{Supplement}.}
    }
    \label{fig:qualitative_comparison}

\end{figure*}

\subsection{Pseudo-multi-view Optimization}
We divide the pseudo-multi-view estimation into two steps. Given a novel view, the synthesized image is expected to contain some parts visible from the input image. As only a single input view is provided, certain parts are occluded in the novel view. For the visible parts, we can directly warp the original textures from the input image. For the occluded parts,  the pretrained generator $G$ is able to synthesize various photo-realistic 3D consistent images, which can be utilized to inpaint the occluded areas.
\subsubsection{Visible Part Reconstruction} 

From the analysis in Figure~\ref{fig:analysis}, although the early stage of the optimization fails to reconstruct faithful textures, the coarse geometry generally matches the input image. Thus we can utilize the initially estimated geometry with the early stop to conduct the visibility analysis.  We denote the optimized latent code as $\mathbf{z_e}$, and then we can reproject the input color image onto a 3D mesh derived from $\mathbf{z_e}$, as shown in Figure~\ref{fig:method}.  For a new camera pose $\mathbf{p_i}$, we project the mesh to the image plane and denote the image regions with projected mesh as visible parts $\mathcal{M}_v$ (other regions as occluded parts $\mathcal{M}_o$). There are abrupt changes in the geometry along the boundary in $\mathcal{M}_v$, so we erode $\mathcal{M}_v$ with blur kernels around the boundary regions. The detailed description of each step is included in the \textit{supplement}.


With the estimated geometry, we can warp the texture from the input image to the novel view. For the novel pose $\mathbf{p_i}$, we obtain the pseudo textures $\mathcal{V}_{i}$ by projecting the color mesh. As shown in Figure~\ref{fig:method}, $\mathcal{V}_{i}$ preserves the details from the input image but contains large missing parts that are  occluded. Next, we will focus on inpaint these occluded parts. 

\subsubsection{Occluded Part Reconstruction}

The inpainted textures for the occluded parts should generate reasonable shapes, be consistent with $\mathcal{V}_{i}$, and be 3D consistent for different views. One possible choice is to directly inpaint the missing regions with the pre-trained image or video inpainting pipelines~\cite{xu2019deep, wang2021image, yu2019free, zhao2021large}. While inpainting for a single view can be reasonable in 2D, the 3D consistency suffers due to the lack of underlying 3D representations. Note that the pretrained 3D-aware generator $G$ can synthesize photo-realistic 3D consistent images, and thus we propose to inpaint the occluded parts with $G$. However, as in Figure~\ref{fig:analysis},  the novel view quality degrades because of the out-of-distribution textures. To increase the robustness of the inpainting, we exclude the out-of-distribution textures for the inversion and then synthesize reasonable novel views to inpaint the occluded parts. 

As shown in Figure~\ref{fig:method}, given a camera pose $\mathbf{p_i}$, the GAN inversion produces a reconstructed image $O_i$, which keeps the general appearance of the input image with inpainted textures and shapes. To exclude the out-of-distribution textures, we calculate a difference map $\mathcal{D}$ as$||\mathbf{x}-G(\mathbf{z_e}, \mathbf{p_0})||_2$. Then we binarize $\mathcal{D}$ by setting the pixel value of  $\mathcal{D}$ to 0 if the difference is greater than a threshold $\theta$; otherwise, its value is set to 1.

Finally, we can optimize a latent code $\mathbf{z_o}$ for occluded parts based on the binary difference map $\mathcal{D}$ with the camera pose $\mathbf{p_i}$: 
 \begin{align}
\mathbf{z}_o =& \mathop{\arg\min}\limits_{\mathbf{z}} \mathcal{L}(G(\mathbf{z}, \mathbf{p_i})\mathcal{D}, x\mathcal{D}), \\
\mathcal{O}_i =& G(\mathbf{z_o}, \mathbf{p_i}).
\end{align}
With $\mathbf{z_o}$, we can generate any pseudo view $\mathcal{O}_i$ for the camera pose $\mathbf{p_i}$. 


\subsubsection{Optimization}
With the estimated pseudo-multi-views as additional supervision, we perform optimization to retrieve the latent code. We unfreeze the pretrained generator $G$ and perform optimization in the $\mathcal{W+}$ space following~ Roich et al.~\cite{roich2022pivotal}. For each iteration, we randomly select an auxiliary camera pose $p_i$ for optimization. Considering the visible parts and the occlusion parts, we can represent the loss in each gradient descent step as
\begin{align}
 \mathcal{L}_{rec} (G(\mathbf{z},\mathbf{p_0}), x) +  \mathcal{L}_{rec} (\mathcal{M}_o G(\mathbf{z},\mathbf{p_i}),  \mathcal{M}_o \mathcal{O}_i ) \nonumber\\
 + \mathcal{L}_{rec} (\mathcal{M}_v G(\mathbf{z},\mathbf{p_i}),  \mathcal{M}_v \mathcal{V}_i  ),
\end{align}
where the reconstruction loss $\mathcal{L}_{rec}$ is the weighted sum of the $\mathcal{L}_2$ loss and the perceptual loss LPIPS with the features extracted from the VGG network~\cite{simonyan2014very,zhang2018unreasonable,Chen2017}.
With the perceptual loss (LPIPS), we achieve higher quality for the perceptual details such as the hairs. After the fitting, the optimized latent code can be used for synthesizing novel views or various attribute editing. 



\begin{figure*}[!t]
    \centering 
    \small
    \vspace{-1em}
    \begin{tabular}{@{}c@{\hspace{.4 mm}}c@{\hspace{0.mm}}c@{\hspace{.4mm}}c@{\hspace{0.mm}}c@{\hspace{.4mm}}c@{\hspace{0.mm}}c@{}}
     \includegraphics[width=0.14\linewidth]{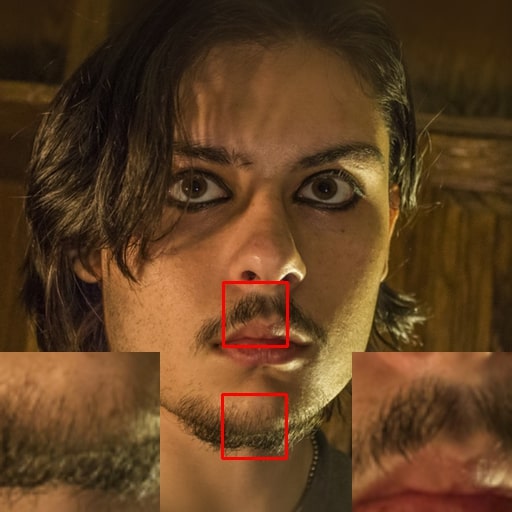}       &   
     \includegraphics[width=0.14\linewidth]{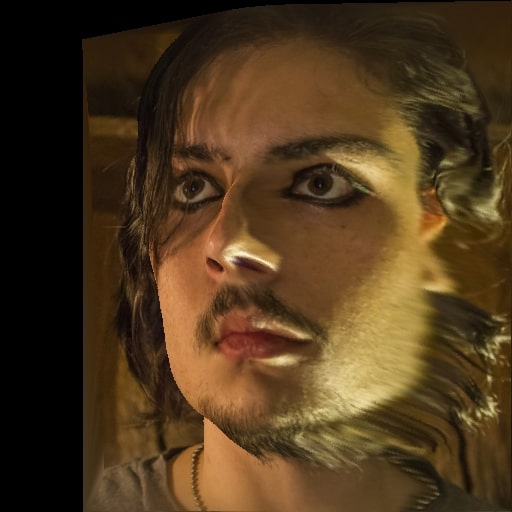}&        
     \includegraphics[width=0.14\linewidth]{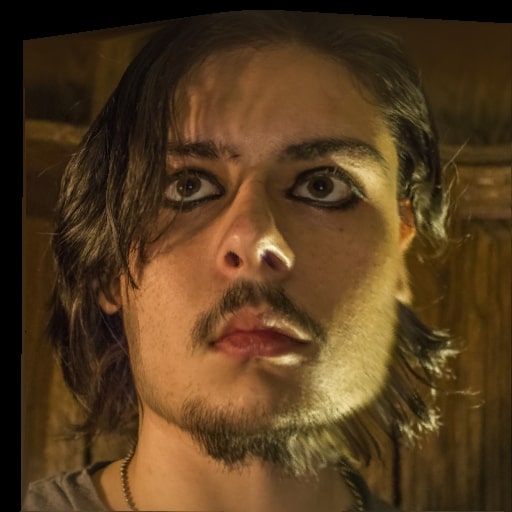} &   
     \includegraphics[width=0.14\linewidth]{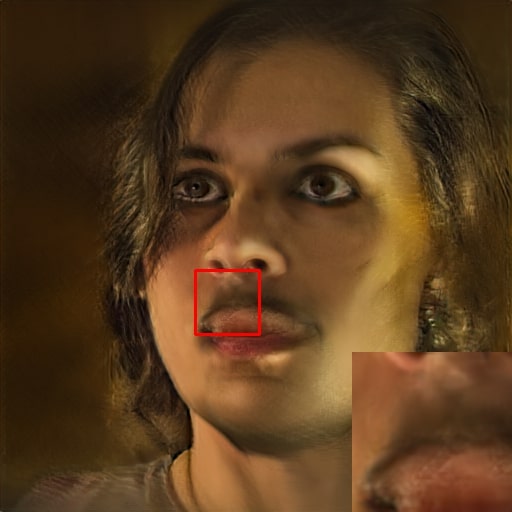} &   
     \includegraphics[width=0.14\linewidth]{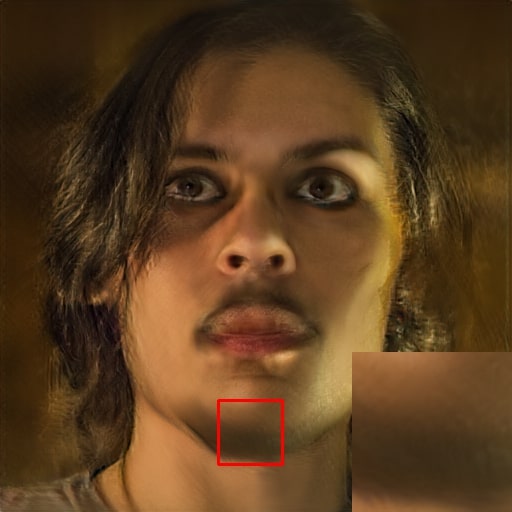} &
     \includegraphics[width=0.14\linewidth]{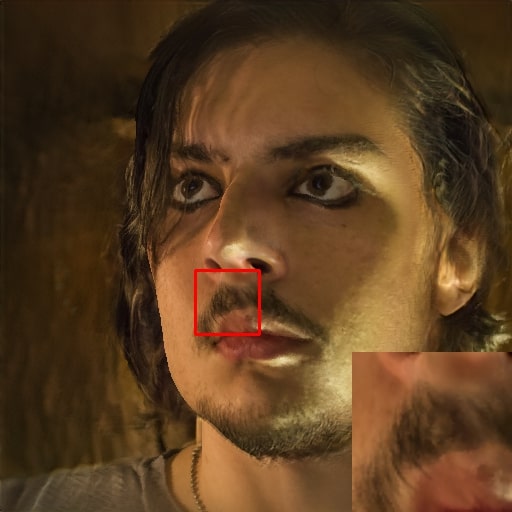} &
     \includegraphics[width=0.14\linewidth]{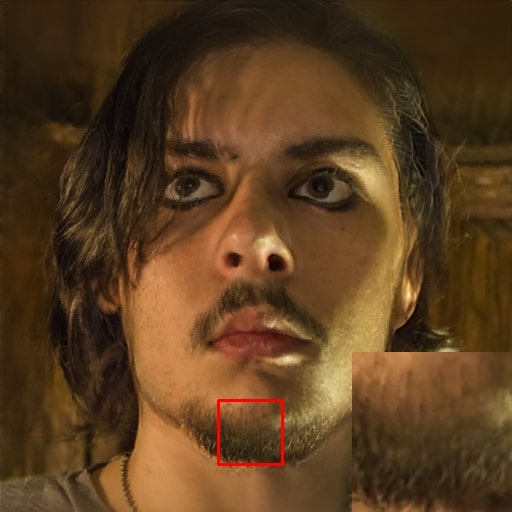} \\ 
     \includegraphics[width=0.14\linewidth]{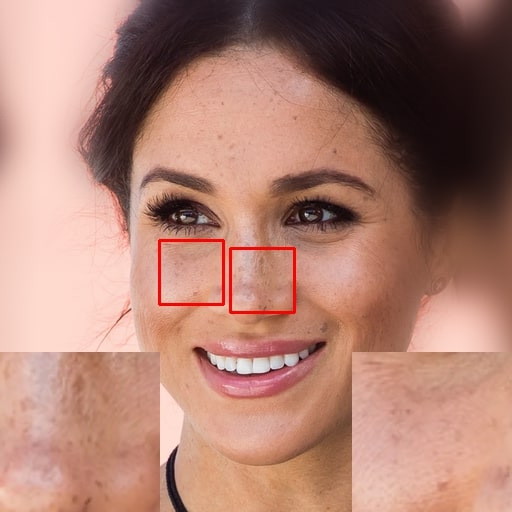}       &   
     \includegraphics[width=0.14\linewidth]{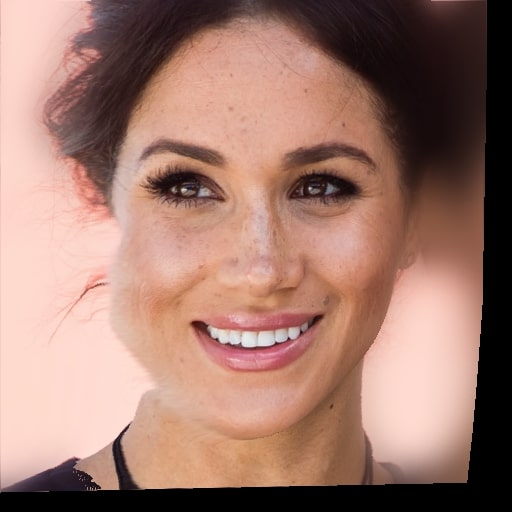}&        
     \includegraphics[width=0.14\linewidth]{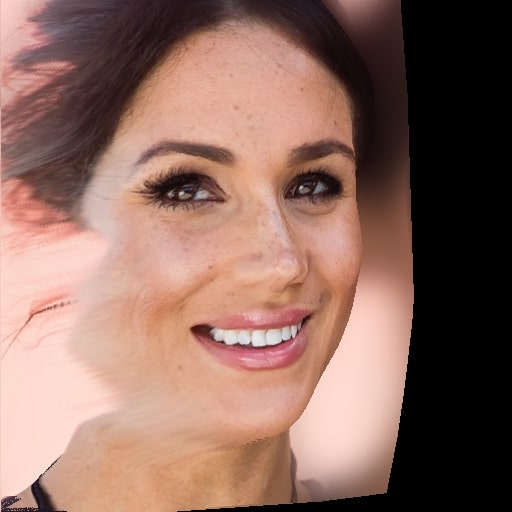}&   
     \includegraphics[width=0.14\linewidth]{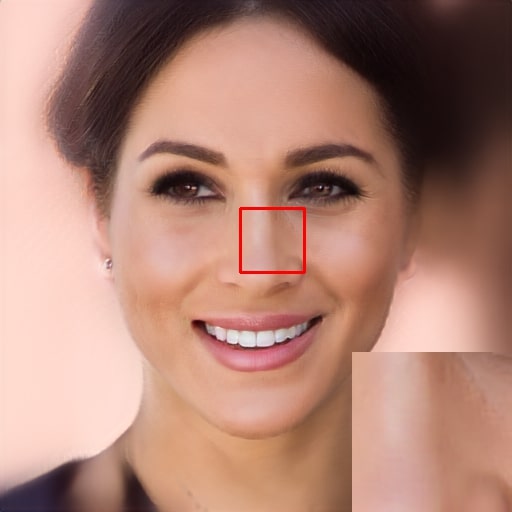} &   
     \includegraphics[width=0.14\linewidth]{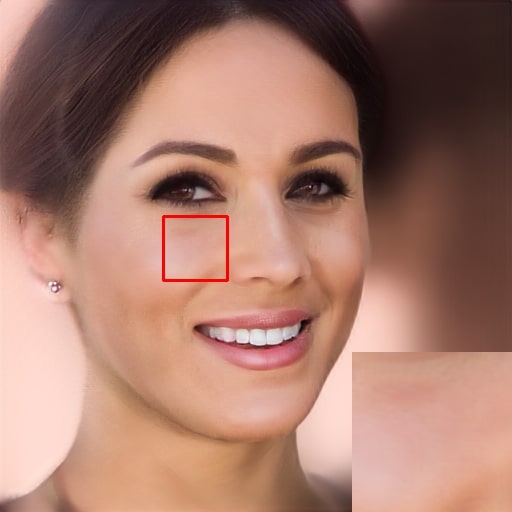} &
     \includegraphics[width=0.14\linewidth]{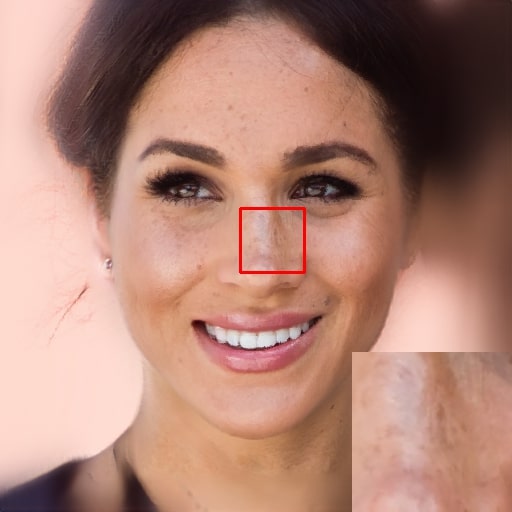} &
     \includegraphics[width=0.14\linewidth]{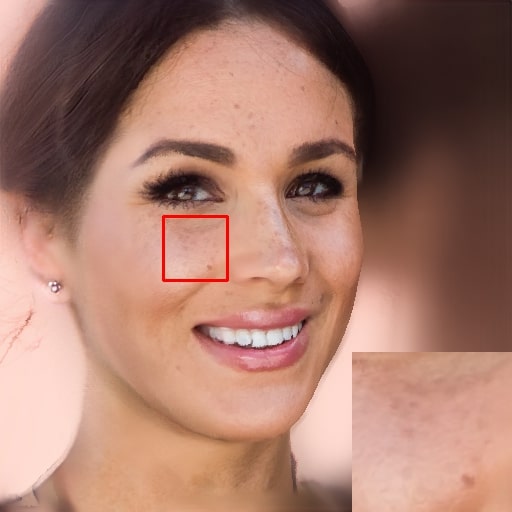} \\
    Input & \multicolumn{2}{c}{Without occluded part}  &  \multicolumn{2}{c}{Without original textures}   &  \multicolumn{2}{c}{Full model}   \\
    \end{tabular}
   \caption{\textbf{Ablation study analysis.} Without the occluded part reconstruction, the synthesized face shape is incorrect. Without the visible textures, the image-specific details, such as the beard and skin textures, are not preserved. Zoom in for details.}
    \label{fig:ablation}
\end{figure*}

\section{Experiments}
\subsection{Experimental Setup}
For the initial visibility estimation stage, we set the learning rate at 5e-3 and the training iteration at 1000. We set the learning rate for the optimization stage at 3e-4 and the training iteration at 3000.  We choose EG3D~\cite{chan2022efficient} as the 3D-aware generator $G$ as it synthesize photo-realistic images with high 3D consistency.  We utilize a pretrained EG3D model trained on the FFHQ~\cite{karras2019style} dataset for optimization and then evaluate the performance on CelebA-HQ~\cite{lee2020maskgan} dataset. All experiments are done on a single NVIDIA 2080 GPU. We attach more detailed settings in the \textit{supplement}. 

\subsection{Evaluation}

We perform both qualitative and quantitative evaluations for the proposed approach in terms of faithfulness and 3D consistency. We compare our method with three state-of-the-art inversion methods, HFGI~\cite{wang2022high}, PTI~\cite{roich2022pivotal} and IDE-3D~\cite{sun2022ide}. HFGI~\cite{wang2022high} is the state-of-the-art encoder-based 2D GAN inversion method that achieves high-fidelity image reconstruction. Although PTI~\cite{roich2022pivotal} was originally proposed for inversing 2D GAN, the method has been proven to achieve reasonable performance on 3D GAN inversion~\cite{chan2022efficient, lin20223d}. IDE-3D~\cite{sun2022ide} proposes a hybrid inversion approach on 3D GAN and trains an encoder that maps from the input image to the latent for initialization. 

\subsubsection{Qualitative analysis}
 
We demonstrate the visual comparisons in Fig.~\ref{fig:qualitative_comparison}. Note that HFGI is for 2D GAN inversion, and thus it has a slightly different viewpoint and is inconsistent in 3D-aware editing. The proposed approach is robust to the out-of-distribution textures such as the tattoos and can faithfully reconstruct the image-specific details, while PTI and IDE-3D fail to keep the identity from the source for the novel views. 



\begin{table} [t]
\footnotesize
\centering 
\begin{tabular}{l@{\hspace{1mm}}c@{\hspace{6mm}}c@{\hspace{6mm}}c@{\hspace{6mm}}c} 
\toprule 
Method &  PSNR$\uparrow$ &SSIM$\uparrow$  &Lpips$\downarrow$ & 3D Consistency$\uparrow$    \\ 
\midrule 
PTI~\cite{roich2022pivotal}  &   26.64  &  0.879 & 0.271 & 21.20\\
IDE-3D~\cite{sun2022ide}   &  26.45  &  0.878 & 0.273 & 20.69\\
HFGI~\cite{wang2021image}   &  22.51  &  0.772 & 0.268 &  \textit{N/A}\\
Ours &   \textbf{29.43}  &   \textbf{0.918} & \textbf{0.172} & \textbf{21.69}\\

\bottomrule
\end{tabular}
\caption{Quantitative evaluation of different GAN inversion methods.} 
\label{tab:quantitative} 
\end{table}

\begin{table} [t]
\footnotesize
\centering 
\begin{tabular}{@{\hspace{1mm}}l @{\hspace{6mm}} c @{\hspace{6mm}} c @{\hspace{1mm}}} 

\toprule 
 &  Ours $>$ PTI~\cite{roich2022pivotal} &  Ours $>$ IDE-3D~\cite{sun2022ide}     \\ 
\midrule 
Preference rate &   90.7  &   92.5 \\
\bottomrule
\end{tabular}
\caption{The result of the user study.} 
\label{tab:user_study} 
\end{table}

\subsubsection{Quantitative analysis}


Table~\ref{tab:quantitative} presents the quantitative comparison between our method and baselines. For test data, We randomly select 1500 images from CelebA-HQ dataset. For reconstruction fidelity, we adopt PSNR, MS-SSIM, and LPIPS~\cite{zhang2018unreasonable} on input image as the evaluation metrics. Our approach obtains the best scores on all the evaluated metrics compared with baselines, which indicates that our method reconstructs high-fidelity details of input images.

In terms of the 3D consistency, we adopt the evaluation setting from ~\cite{gu2021stylenerf}. Specifically, we use 5 synthesized novel views near the input camera pose to predict the input image using the IBRNet~\cite{wang2021ibrnet} and calculate the difference with PSNR. The reported metrics in the novel view synthesis prove that more stable 3D consistency is achieved by our approach. Note that the quantitative evaluation of the 3D consistency is still an open question, and we report more metrics in the \textit{supplement}.  We highly recommend readers watch the supplementary video for a comprehensive evaluation of 3D consistency.

\begin{figure*}[t]
    \centering

\small
\begin{tabular}{@{}c@{\hspace{0.4mm}}c@{\hspace{0.4mm}}c@{\hspace{0.4mm}}c@{\hspace{0.4mm}}c@{\hspace{0.4mm}}c@{\hspace{0.4mm}}c@{\hspace{0.4mm}}c@{\hspace{0.4mm}}c@{\hspace{0.4mm}}c@{}}
\multirow{2}[2]{0.20\linewidth}[15.0mm]
{\includegraphics[width=\linewidth]{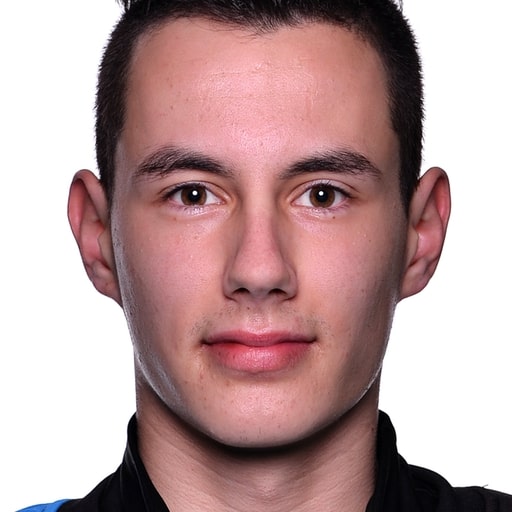}}&
\includegraphics[width=0.10\linewidth,height=0.10\linewidth]{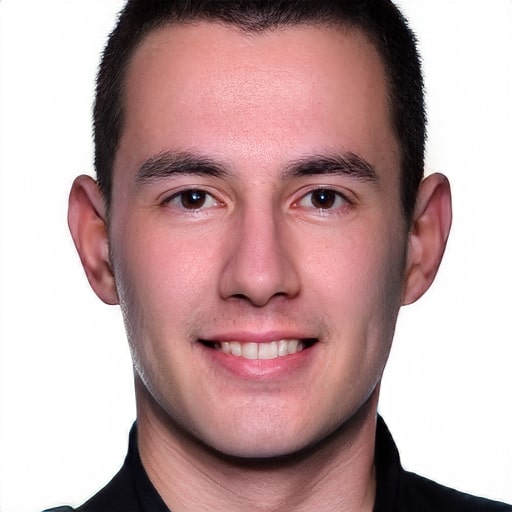}&
\includegraphics[width=0.10\linewidth,height=0.10\linewidth]{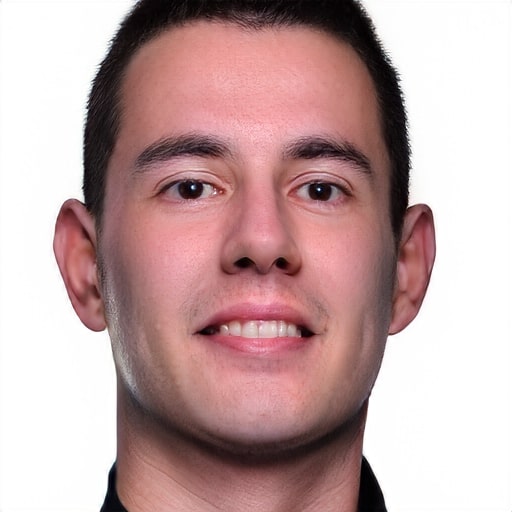}&
\includegraphics[width=0.10\linewidth,height=0.10\linewidth]{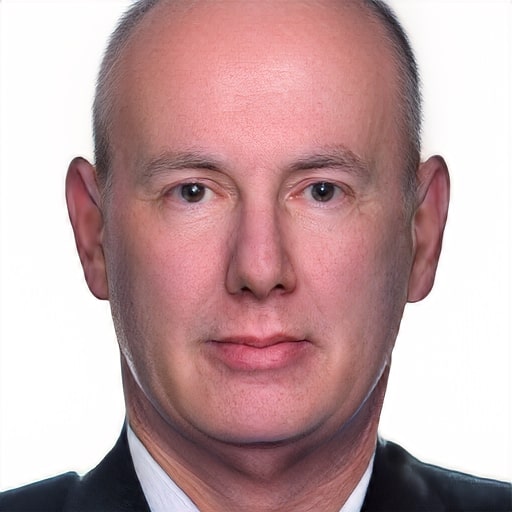}&
\includegraphics[width=0.10\linewidth,height=0.10\linewidth]{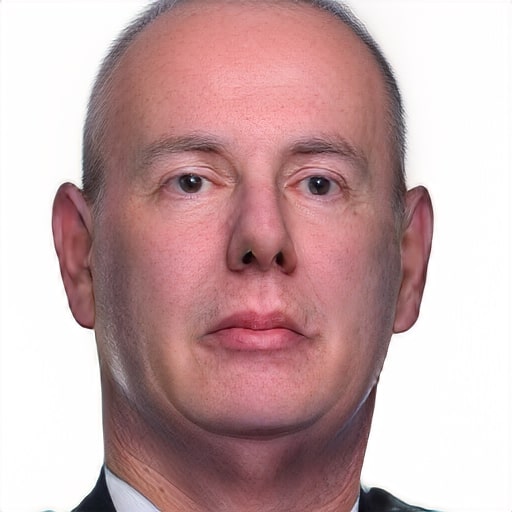}&
\includegraphics[width=0.10\linewidth,height=0.10\linewidth]{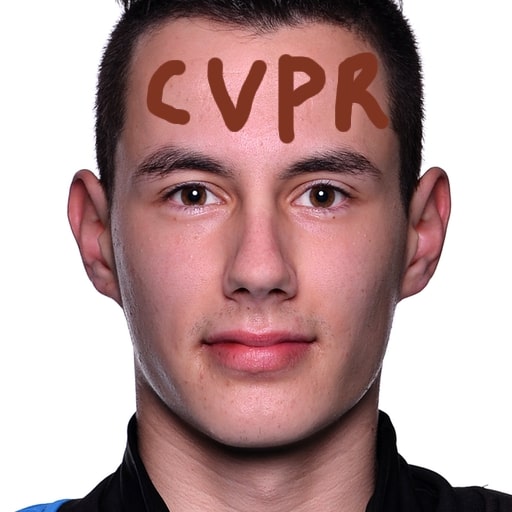}&
\includegraphics[width=0.10\linewidth,height=0.10\linewidth]{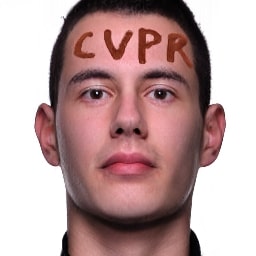}&
\includegraphics[width=0.10\linewidth,height=0.10\linewidth]{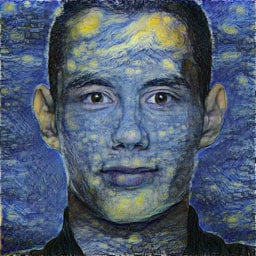}&
\includegraphics[trim={0cm 0cm 0.8cm 0.8cm},clip=true, width=0.10\linewidth,height=0.10\linewidth]{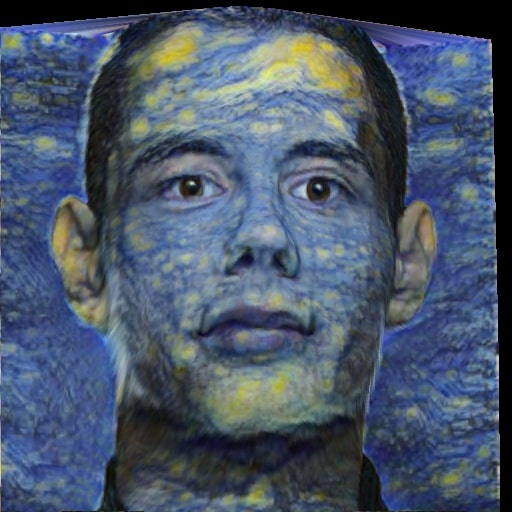}
\\

&
\includegraphics[width=0.10\linewidth,height=0.10\linewidth]{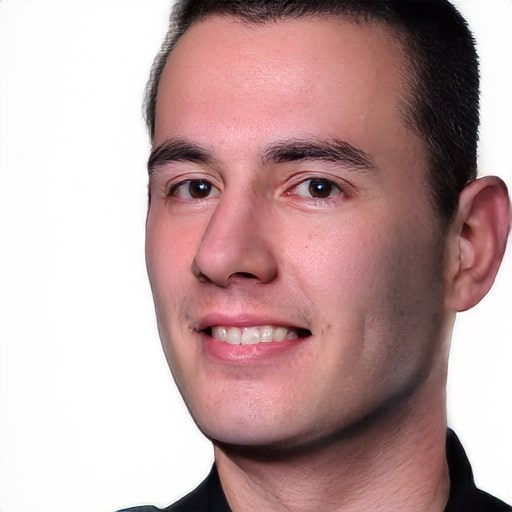}&
\includegraphics[width=0.10\linewidth,height=0.10\linewidth]{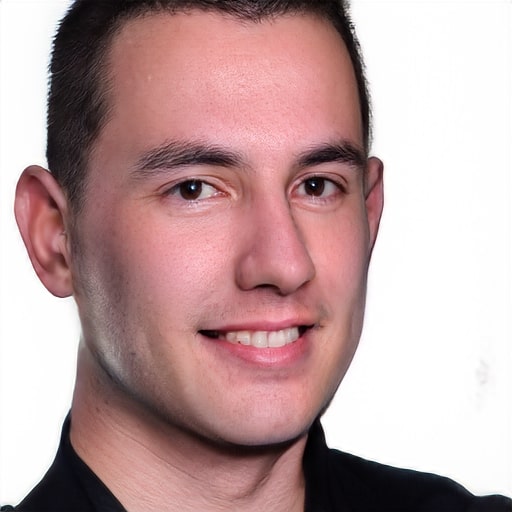}&
\includegraphics[width=0.10\linewidth,height=0.10\linewidth]{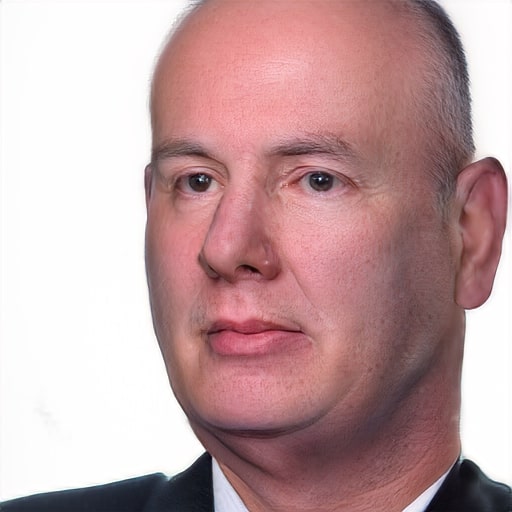}&
\includegraphics[width=0.10\linewidth,height=0.10\linewidth]{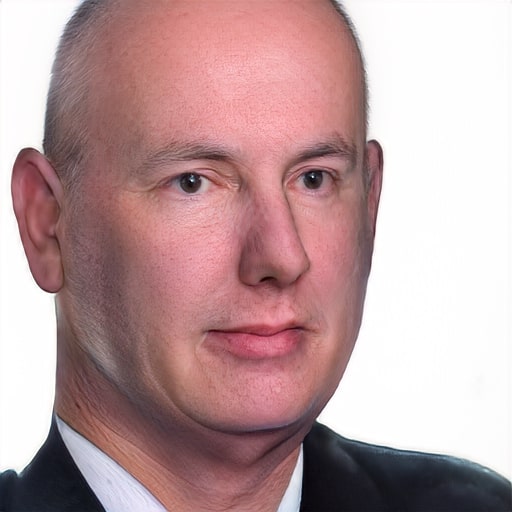}&
\includegraphics[width=0.10\linewidth,height=0.10\linewidth]{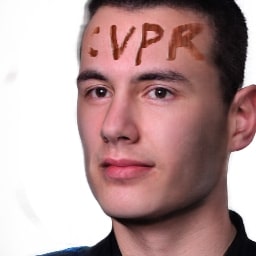}&
\includegraphics[width=0.10\linewidth,height=0.10\linewidth]{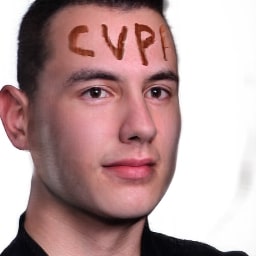}&
\includegraphics[trim={2.2cm 1.2cm 0cm 1cm},clip=true, width=0.10\linewidth,height=0.10\linewidth]{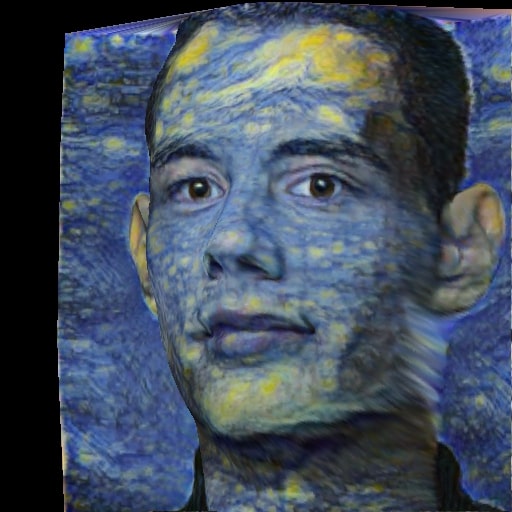}&
\includegraphics[trim={0cm 1.2cm 2.2cm 1cm},clip=true, width=0.10\linewidth,height=0.10\linewidth]{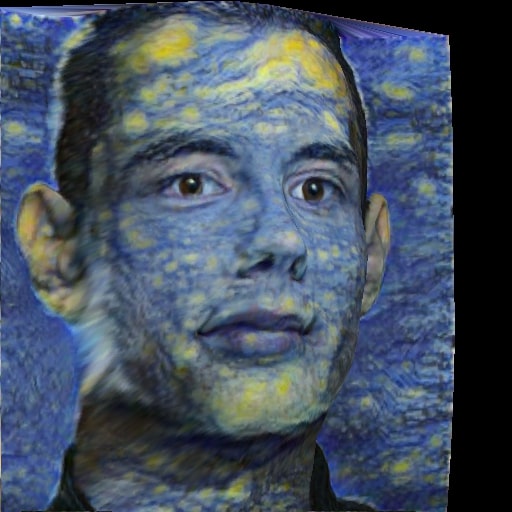}
\\

\multirow{2}[2]{0.20\linewidth}[15.0mm]
{\includegraphics[width=\linewidth]{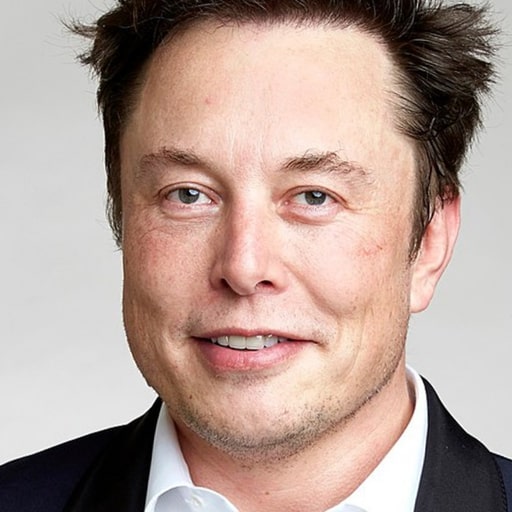}}&
\includegraphics[width=0.10\linewidth,height=0.10\linewidth]{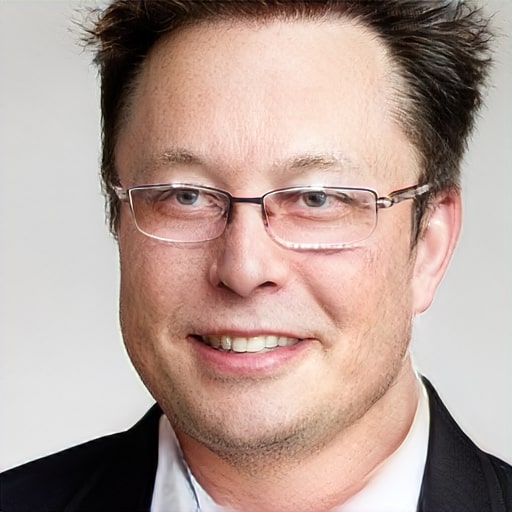}&
\includegraphics[width=0.10\linewidth,height=0.10\linewidth]{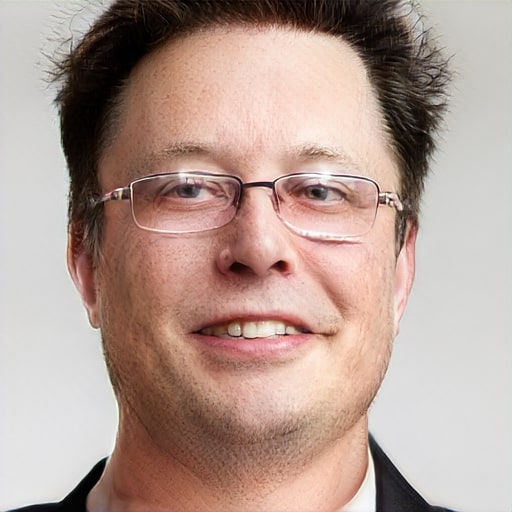}&
\includegraphics[width=0.10\linewidth,height=0.10\linewidth]{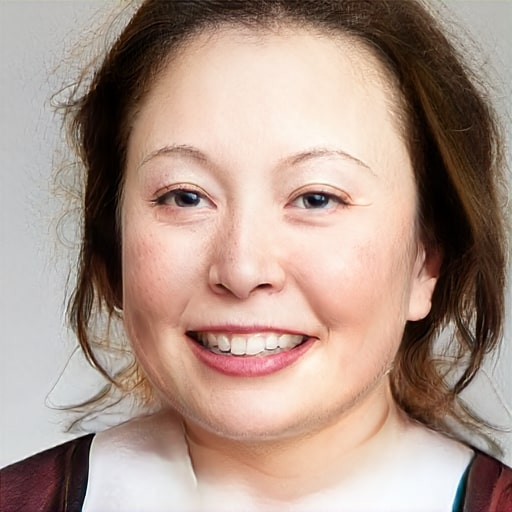}&
\includegraphics[width=0.10\linewidth,height=0.10\linewidth]{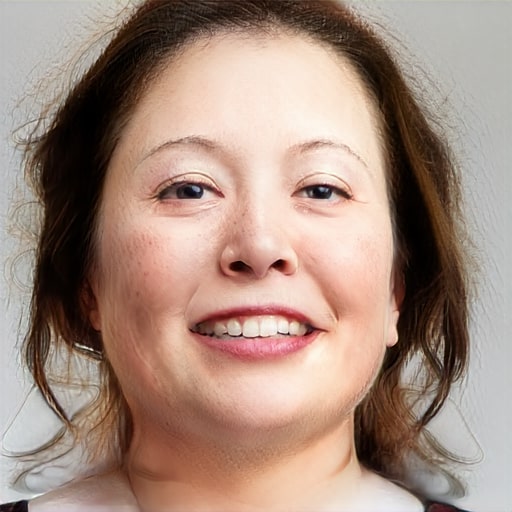}&
\includegraphics[width=0.10\linewidth,height=0.10\linewidth]{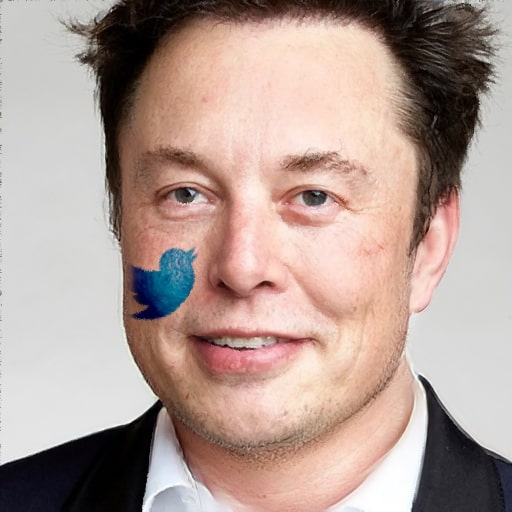}&
\includegraphics[width=0.10\linewidth,height=0.10\linewidth]{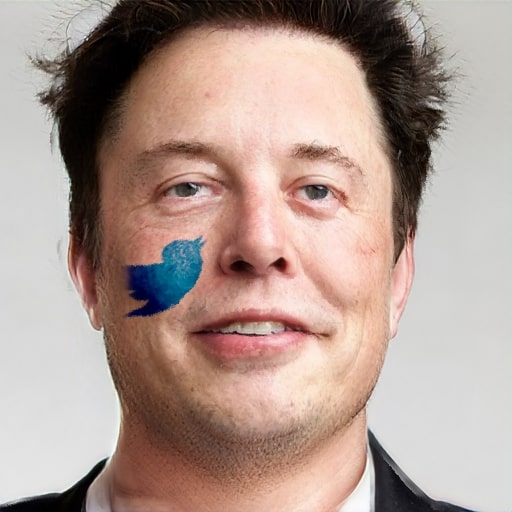}&
\includegraphics[width=0.10\linewidth,height=0.10\linewidth]{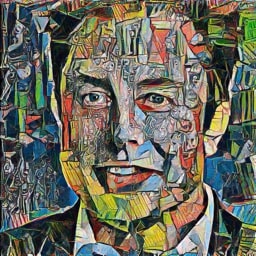}&
\includegraphics[trim={0cm 0cm 1.2cm 1.2cm},clip=true,width=0.10\linewidth,height=0.10\linewidth]{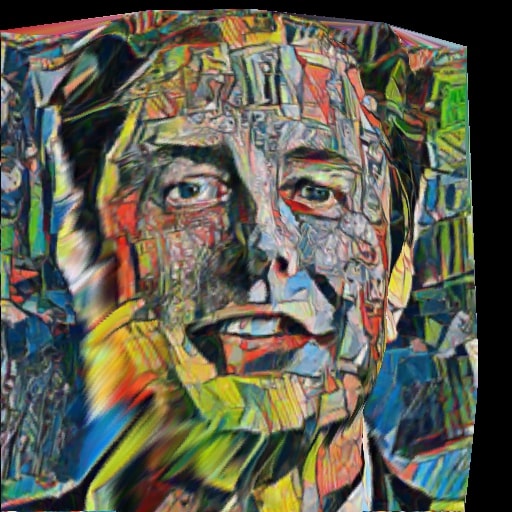}
\\

&
\includegraphics[width=0.10\linewidth,height=0.10\linewidth]{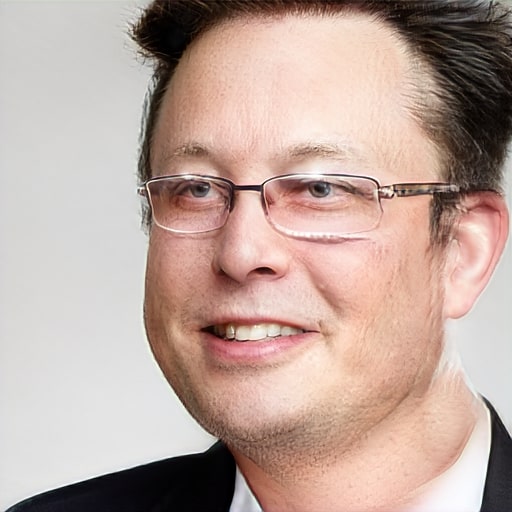}&
\includegraphics[width=0.10\linewidth,height=0.10\linewidth]{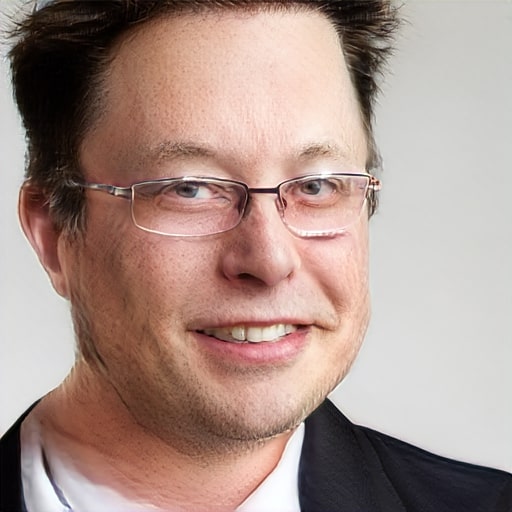}&
\includegraphics[width=0.10\linewidth,height=0.10\linewidth]{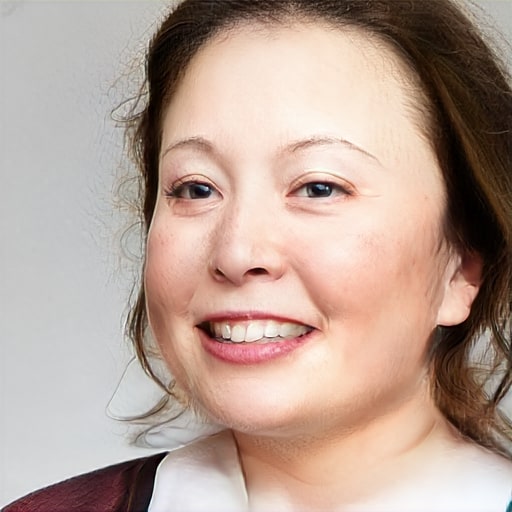}&
\includegraphics[width=0.10\linewidth,height=0.10\linewidth]{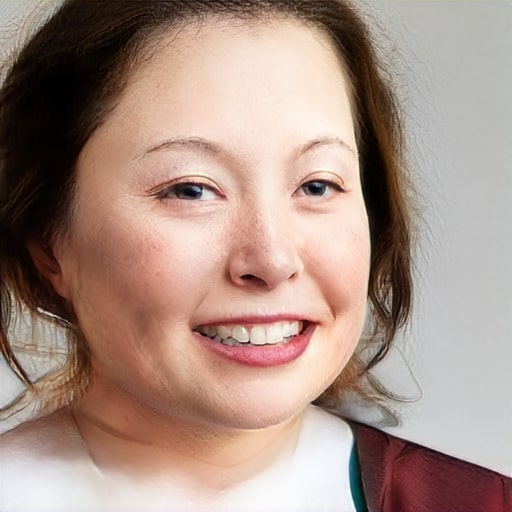}&
\includegraphics[width=0.10\linewidth,height=0.10\linewidth]{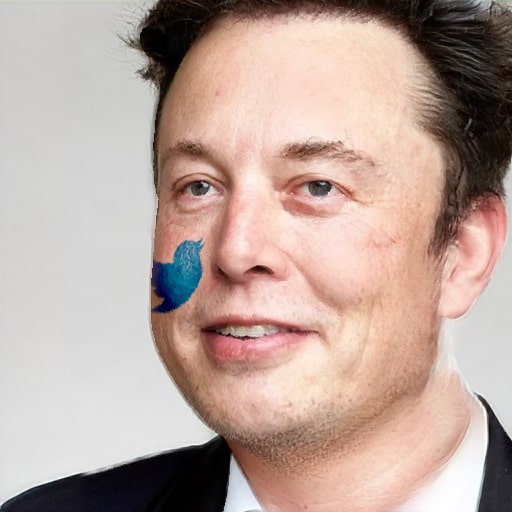}&
\includegraphics[width=0.10\linewidth,height=0.10\linewidth]{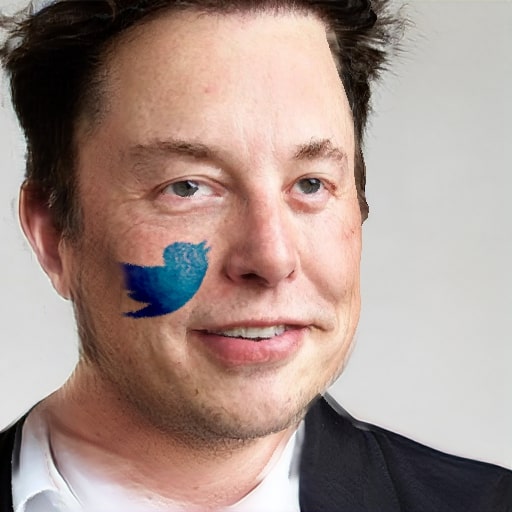}&
\includegraphics[trim={1.2cm 0cm 0cm 1.2cm},clip=true,width=0.10\linewidth,height=0.10\linewidth]{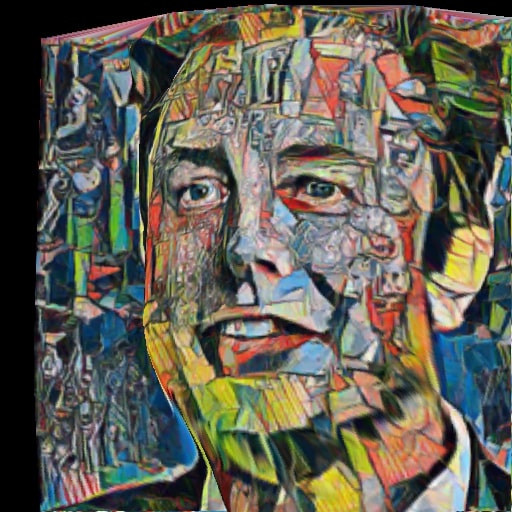}&
\includegraphics[trim={0cm 1.4cm 2.8cm 1.4cm},clip=true, width=0.10\linewidth,height=0.10\linewidth]{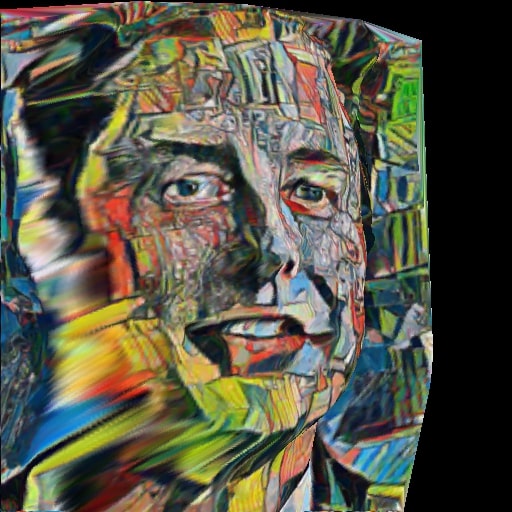}
\\

Input & \multicolumn{4}{c}{(a) Latent Attribute Editing}  &  \multicolumn{4}{c}{(b) 3D-aware Texture Modification}    \\

\end{tabular}
    \caption{3D-aware editing results on real-life photos. Our approach enables two types of editing: (a) latent attribute editing where we can modify specific attributes such as smile, age, glasses and gender. (b)3D-aware texture modification: by editing the textures on the input images such as adding logos or stylization. }
    \label{fig:editing}
    \vspace{-1mm}
\end{figure*}

\subsubsection{User study}
We conduct a user study to perceptually evaluate the reconstruction quality of our approach. We use 14 random images from our test dataset and perform the 3D inversion. For each vote, the user is provided with the input image, the video rendered with a sphere camera trajectory that looks at the center of the face of our approach, and the video of baselines rendered with the same trajectory. We ask the participant to compare which video is preferred in the following two aspects: keeping the best identity of the input image and inducing the least flicker. As in Table~\ref{tab:user_study}, from the collected 1120 votes from 40 participants, the proposed method outperforms other baselines by a large margin. 

\subsubsection{Ablation analysis}
We conduct the ablation analysis to show the effectiveness of our design. Specifically, we implement two ablated models: (1) Without occluded parts, we remove our occluded part reconstruction and apply bilinear interpolation instead of generative priors for invisible pixels; (2) Without original textures, we use the generated texture instead of the original textures for the visible parts. Figure~\ref{fig:ablation} show the qualitative comparison. We spot the following findings: (1) with only interpolation, the synthesized novel view contains obvious artifacts near the face boundary. The generated invisible part leads to obvious improvement in synthesizing the reasonable face shape; (2) with only the generated textures, the high-frequency details from the input images are lost.

\subsection{Applications}
Inversion of 3D GANs enables many applications, such as 3D-aware editing. We demonstrate two types of editing: latent-based attribute editing and texture-based image editing. For attribute editing, we follow the pipeline in ~\cite{shen2020interpreting} to calculate the moving direction in the latent space (Details attached in the \textit{supplement}). We show transferring gender, changing ages, smiling, and wearing glasses in Figure~\ref{fig:editing}(a) and render the corresponding 3D-consistent views. As our approach can handle out-of-distribution textures, we can perform texture editing on the input image and synthesize the novel view for the modified images. As in Figure~\ref{fig:editing}(b), we can paint the desired textures (e.g., CVPR on the forehead and logo on the face) or apply different styles on the input view. We can generate 3D-consistent views for the edited input images. 

\section{Conclusion}

This work studies high-fidelity 3D GAN inversion, which enables latent-based attribute editing and texture-based image editing. Extensive experiments demonstrate that our method can robustly synthesize the novel view of the input image with excellent detail preservation. The proposed pipeline is general and easy to apply as we can conduct the visibility analysis and the pseudo-multi-view generation for 3D-aware GANs. Still, our approach suffers from several limitations. One primary limitation is the difficulty in reconstructing the geometry of the out-of-distribution objects (e.g., trinkets and hands). As a result of the initially incorrect geometry, the following operations inevitably fail to synthesize reasonable results. Additionally, the estimated geometry for input with extreme poses may suffer from distortions. We attach examples of the failure cases in the \textit{supplement}.  Nevertheless, the proposed approach is promising to serve as a practical solution for 3D-aware reconstruction and editing with only a single input, and we expect future works to solve the remaining issues. 

\clearpage
{\small
\bibliographystyle{ieee_fullname}
\bibliography{egbib}
}

\clearpage


\appendix

\section{Implementation details} 
\subsection{Optimization settings} 
 We utilize a pretrained EG3D model trained on the FFHQ~\cite{karras2019style} dataset for optimization. The model utilizes the triplane with 256 resolution and generates 3D-aware photo-realistic images at 512 resolution. For the initial visibility estimation stage, we set the learning rate at 5e-3 and the training iteration at 1000. With the initial geometry, we first calculate the visibility for each vertex in the mesh. We utilize the z-buffer of the rasterization algorithm to find if other mesh faces occlude this vertex. Then we warp the texture of the input image to visible vertices. Finally, we rasterize the mesh to get novel view masks and images and complete the visible part reconstruction. Next, we utilize the generator to generate the invisible part. For the blending, we set the erosion radius to 1 pixel. To blur the boundary, we have tried two approaches: directly applying a Gaussian kernel with a radius of 10 or using the Poisson blending. The Poisson blending generally results in smoother images. We generate the pseudo-multi-views by blending visible and invisible parts with the smoothed boundary. With the synthesized pseudo-multi-views, we set the learning rate for the final optimization stage at 3e-4 and the training iteration at 3000. All experiments are done on a single NVIDIA 2080 GPU, and it takes five minutes to complete the estimation and optimization.


\subsection{Image attribute editing} 
For image attribute editing, we follow the pipeline proposed in ~\cite{shen2020interpreting} to calculate the latent direction. We first generate 500,000 images with the canonical view. We then predict the attributes of the synthesized images and rank them by scores. We use the 10,000 samples with the highest and 10,000 with the lowest scores. We then randomly use 70\% of them for training a linear SVM and 30\% for testing the accuracy of the trained classifier. We calculate the attribute direction with the trained SVM.

\section{Analysis}
\subsection{More quantitative evaluation metrics} 
\subsubsection{IBRNet}
We adopt the 3D consistency evaluation setting from ~\cite{gu2021stylenerf}, and we use five synthesized novel views to predict the input view using the pretrained  IBRNet~\cite{wang2021ibrnet} model. The five views are selected in the following way:  the canonical view and the four edge views on the sphere camera trajectory where the yaw range is [-0.35,0.35] rad and the pitch range is [-0.25,0.25] rad. In the main paper, we show the PSNR metric that indicates the difference between ground truth images and the reconstruction from IBRNet. In Table~\ref{tab:more_quantitative}, we also attach the SSIM and lpips metrics, and the proposed method outperforms the other baselines. 

\subsubsection{Pose accuracy}
We adopt the pose accuracy metric from~\cite{chan2022efficient}. With the face reconstruction model~\cite{deng2019accurate}, we randomly select one novel view for every CelebA-HQ test image and estimate the pose for the novel view outputs of all the methods. We compute the L2 loss between the estimated pose and the GT pose that is used to render the novel view. From table~\ref{tab:more_quantitative}, we find that the rendered novel view images of all three methods keep good pose consistency with the GT pose, the error is minor, and the margin between different methods is extremely small.

\subsubsection{Identity}
An intuitive criterion for judging the 3D-aware GAN inversion is whether it keeps the identity of the input person in novel views. We have evaluated the criterion with a user study in the main paper, asking people which video keeps the best identity of the input image. We also provide quantitative metrics. We randomly select one novel view for every CelebA-HQ test image and compute the mean Arcface~\cite{deng2019arcface} cosine similarity between rendered novel view images and input images. The comparison is shown in table~\ref{tab:more_quantitative}. Our ID loss exceeds other baselines by a large margin.

\subsection{Texture-geometry trade-off} 
\begin{figure}[t]
    \centering
    \includegraphics[width=1.0\linewidth]{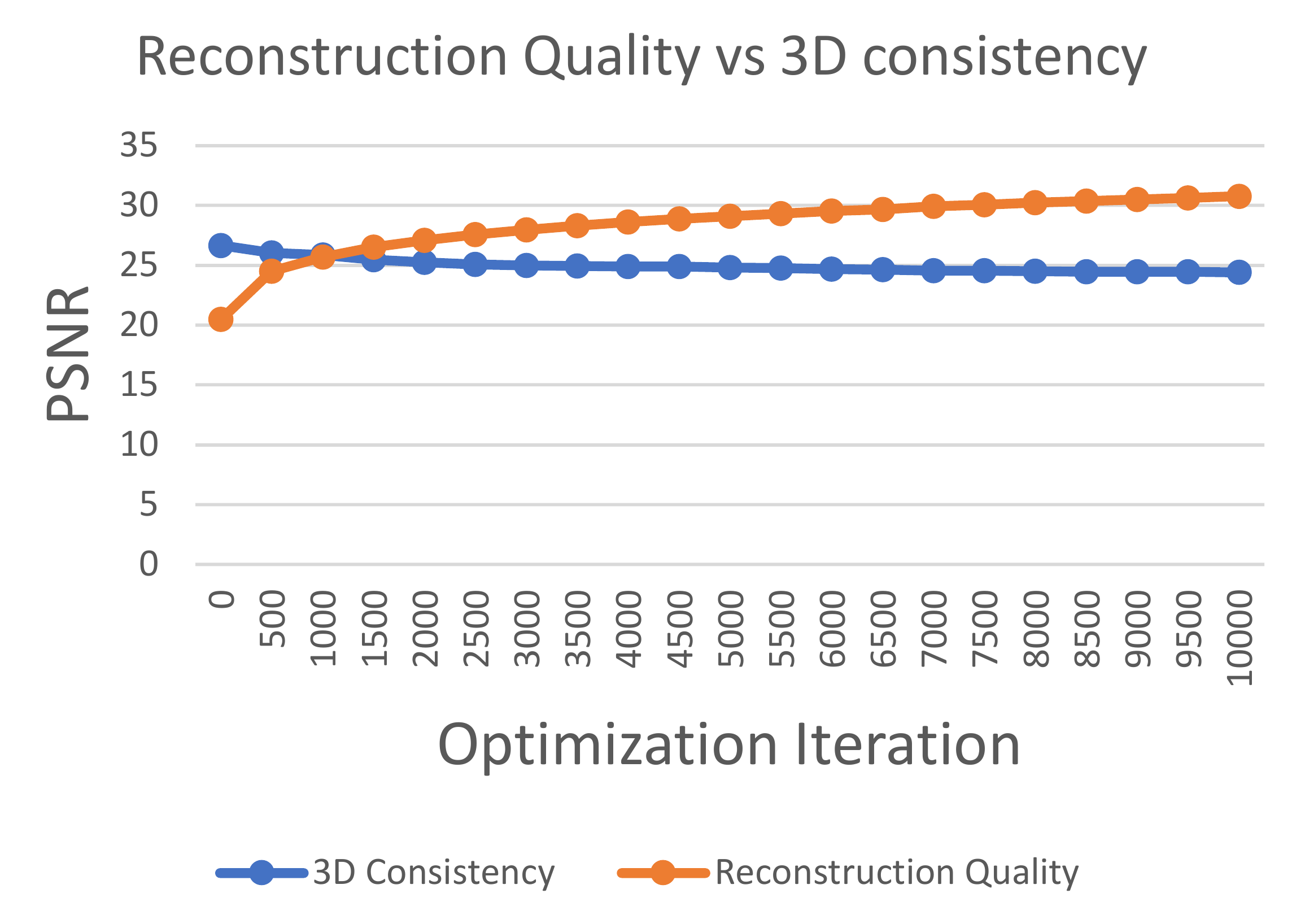}
    \caption{As the reconstruction quality increases with the optimization iteration, the 3D consistency on the contrary decreases.}
    \label{fig:tradeoff}
    \vspace{-1mm}
\end{figure}

As analyzed in Figure 2 of the main paper, the quality of the synthesized view severely decreases as the optimization iteration increases. We also provide quantitative metrics as shown in Figure~\ref{fig:tradeoff}. We calculate the 3D consistency metric using the IBRNet~\cite{wang2021ibrnet} for every 500 optimization steps. As the PSNR of the reconstruction of the source image increases, the 3D consistency metrics, on the contrary, decreases. The quantitative metrics also indicate a degradation in novel view quality as the optimization process continues, which matches the findings in our qualitative analysis. 
\section{Additional visual results} 
We provide more visual results, including the failure cases and the qualitative comparison with baselines. 

\subsection{Failure cases}
\begin{figure}[!t]
    \centering 
    \small
    \vspace{-1em}
    \begin{tabular}{@{}c@{\hspace{1mm}}c@{\hspace{1mm}}c@{}}
     \includegraphics[width=0.3\linewidth]{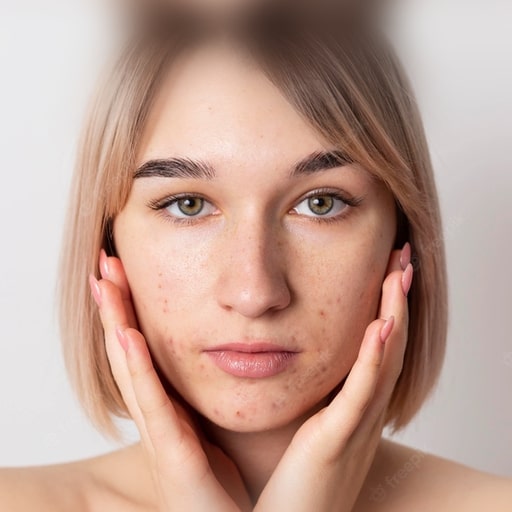}       &   
     \includegraphics[width=0.3\linewidth]{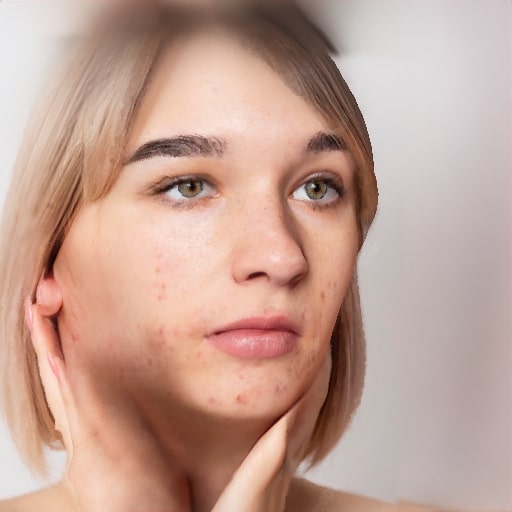}       &        
     \includegraphics[width=0.3\linewidth]{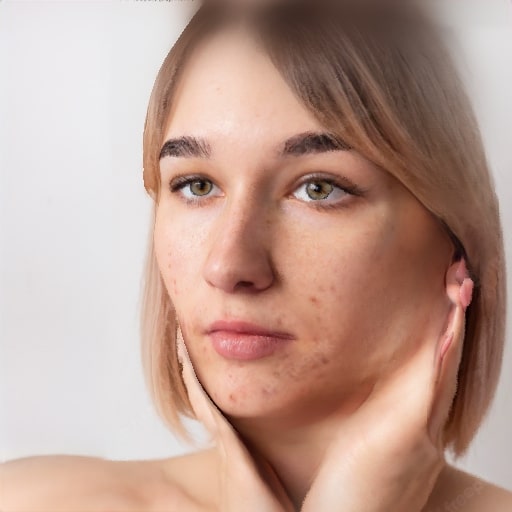}        \\ 

     \includegraphics[width=0.3\linewidth]{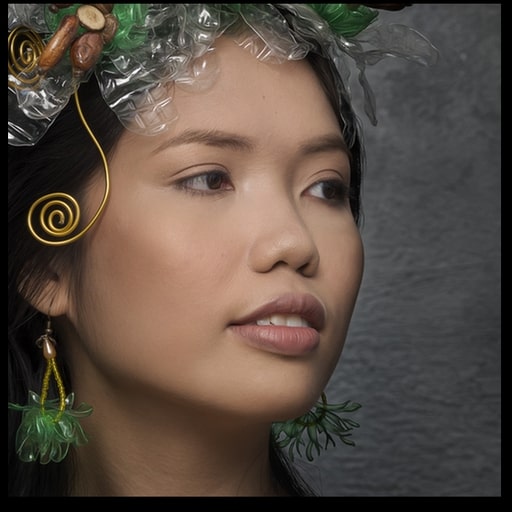}       &   
     \includegraphics[width=0.3\linewidth]{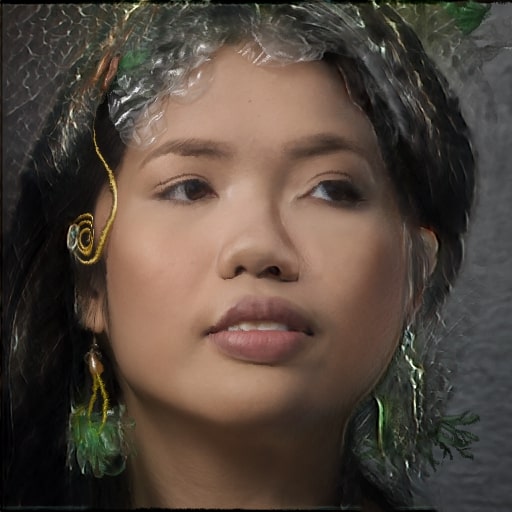}       &        
     \includegraphics[width=0.3\linewidth]{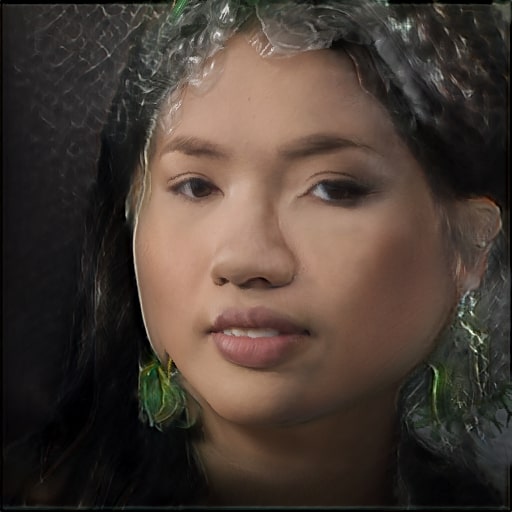} \\
    Input image & Novel view 1 & Novel view 2 \\
    \end{tabular}
   \caption{{Failure cases.}}
    \label{fig:failure case}
\end{figure}

In Figure~\ref{fig:failure case}, we demonstrate two failure cases for our approach. The first source image contains hands, and the estimated geometry is incorrect. We can see obvious blurry regions in the synthesized novel views. The second case contains the out-of-distribution pose with trinkets. The generated face suffers from slight distortion, and the trinkets' shape is incorrect.

\begin{table} [t]
\footnotesize
\centering 
\begin{tabular}{l@{\hspace{1mm}}c@{\hspace{6mm}}c@{\hspace{6mm}}c@{\hspace{6mm}}c@{\hspace{6mm}}c} 
\toprule 
Method &  PSNR$\uparrow$ &SSIM$\uparrow$  &Lpips$\downarrow$ &Pose $\downarrow$ &ID $\uparrow$ \\ 
\midrule 
PTI~\cite{roich2022pivotal}  &   21.20  &  0.697 & 0.457& 0.04178 &0.657\\
IDE-3D~\cite{sun2022ide}   &  20.69  &  0.676 & 0.462 & \textbf{0.04152} & 0.671\\
Ours &   \textbf{21.69}  &   \textbf{0.734} & \textbf{0.429} &0.04179& \textbf{0.744}\\
\bottomrule
\end{tabular}
\caption{More quantitative evaluation metric on 3D consistency.} 
\label{tab:more_quantitative} 
\end{table}

\begin{figure*}[t]
\centering
\small
\hspace*{-0.1cm}
\begin{tabular}{@{}c@{\hspace{0.4mm}}c@{\hspace{0.4mm}}c@{\hspace{0.4mm}}c@{\hspace{1.0mm} \unskip \vrule\hspace{1.0mm}}c@{}}

\multirow{2}[2]{0.16\linewidth}[13.2mm]
{
\includegraphics[width=\linewidth]{figures/figure2/stripe.jpg}}&
\includegraphics[width=0.16\linewidth,height=0.16\linewidth]
{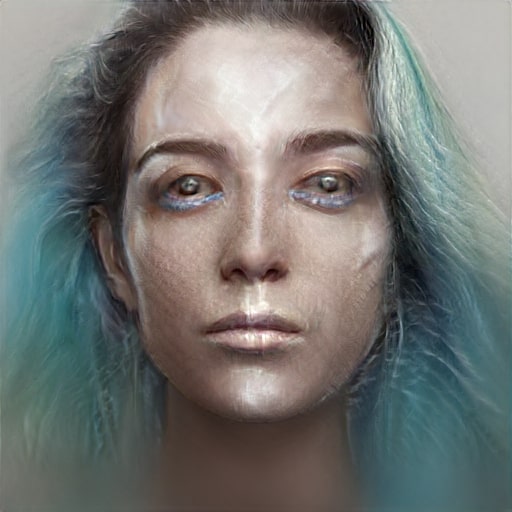}&
\includegraphics[width=0.16\linewidth,height=0.16\linewidth]
{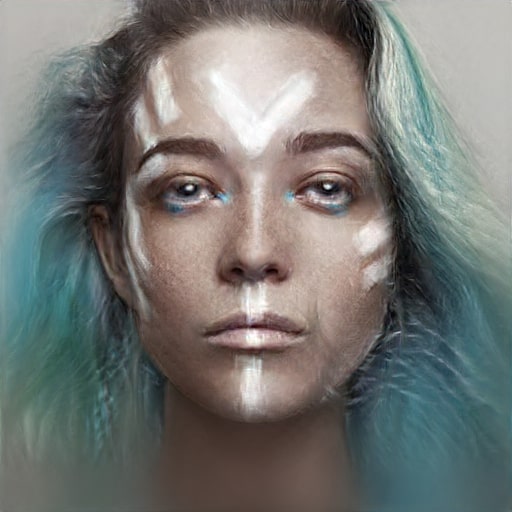}&
\includegraphics[width=0.16\linewidth,height=0.16\linewidth]{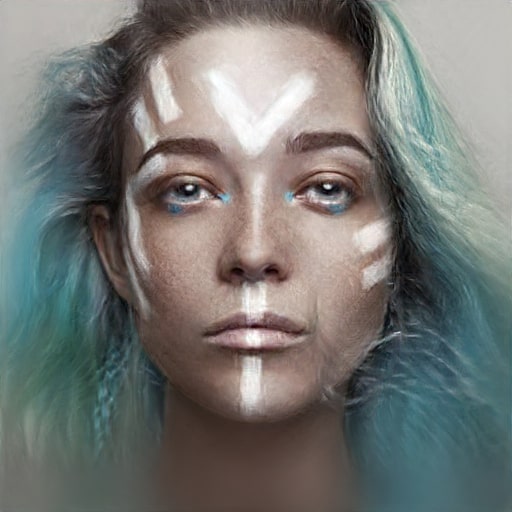}&
\includegraphics[width=0.16\linewidth,height=0.16\linewidth]{figures/figure2/blender_30_ours.jpg}
\put(-320,85){$\rightarrow{}$--------------------- Reconstruction Quality --------------------  $\rightarrow{}$} 
\put(-390,36){Input Image}

\put(-50,85){Ours}\\


&

\includegraphics[width=0.16\linewidth,height=0.16\linewidth]{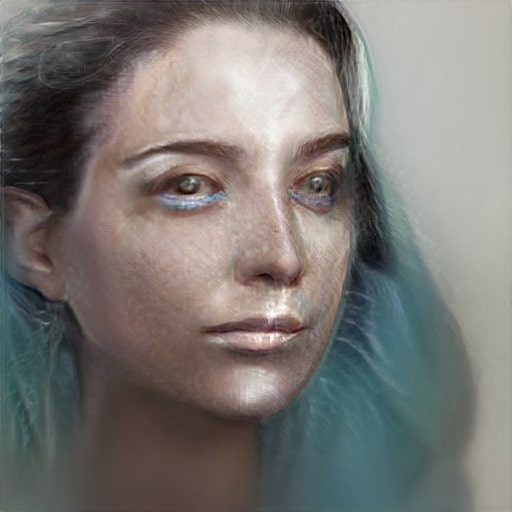}&
\includegraphics[width=0.16\linewidth,height=0.16\linewidth]
{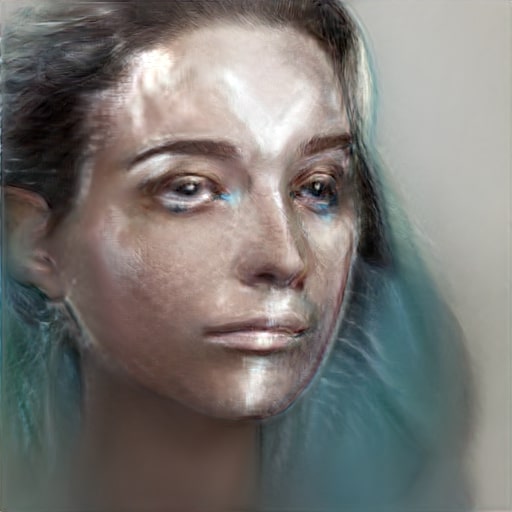}&
\includegraphics[width=0.16\linewidth,height=0.16\linewidth]{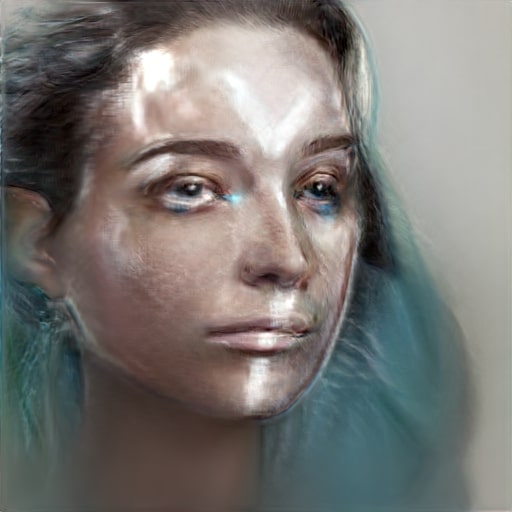}&
\includegraphics[width=0.16\linewidth,height=0.16\linewidth]
{figures/figure2/debug_hybrid.jpg} \\

\multirow{2}[2]{0.16\linewidth}[13.2mm]
{
\includegraphics[width=\linewidth]{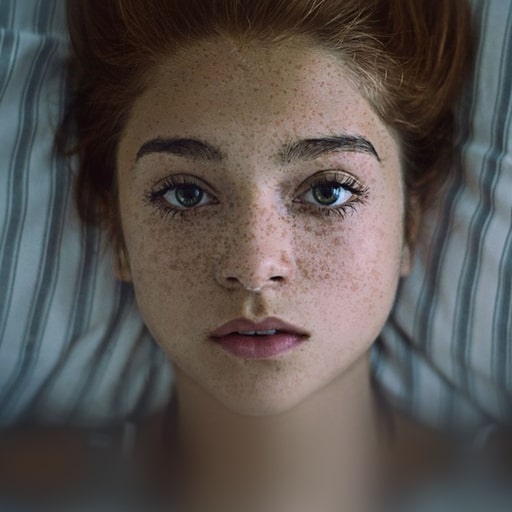}}&
\includegraphics[width=0.16\linewidth,height=0.16\linewidth]
{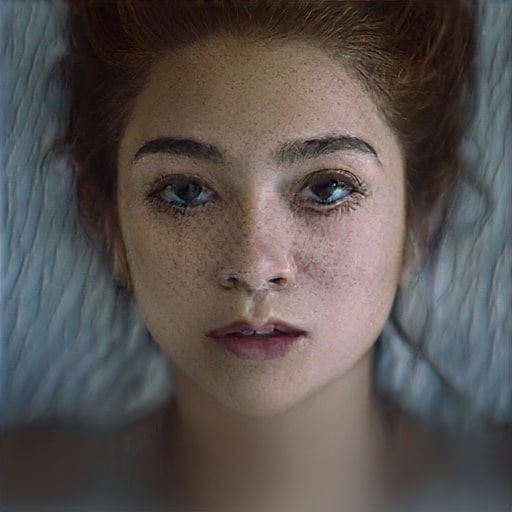}&
\includegraphics[width=0.16\linewidth,height=0.16\linewidth]
{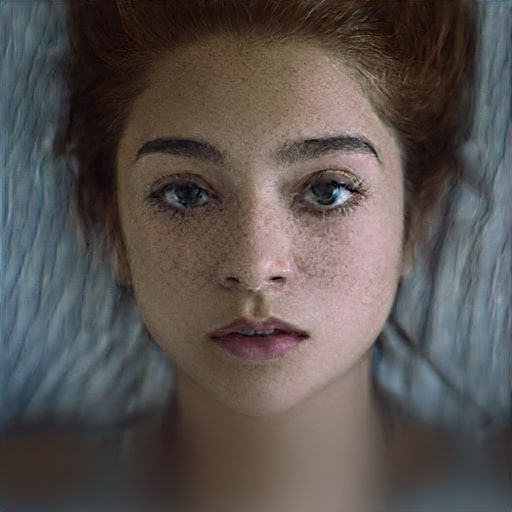}&
\includegraphics[width=0.16\linewidth,height=0.16\linewidth]{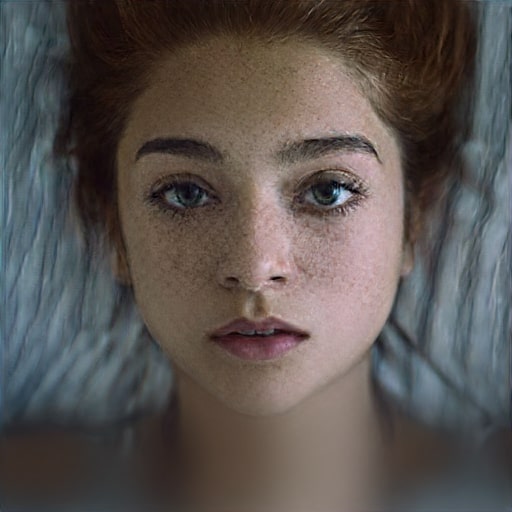}&
\includegraphics[width=0.16\linewidth,height=0.16\linewidth]{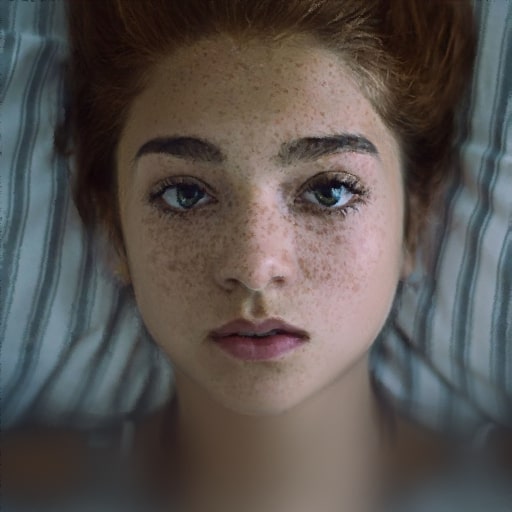}\\


&

\includegraphics[width=0.16\linewidth,height=0.16\linewidth]{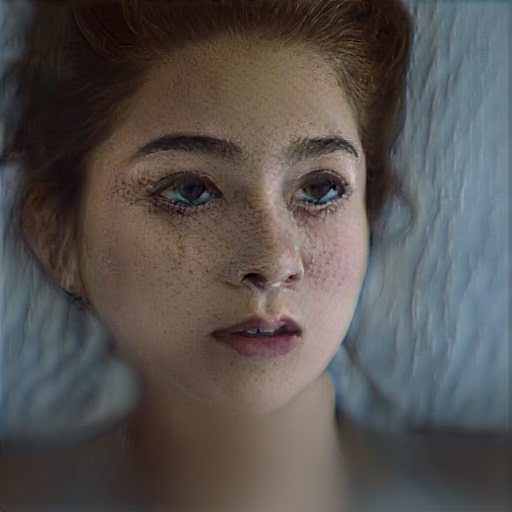}&
\includegraphics[width=0.16\linewidth,height=0.16\linewidth]
{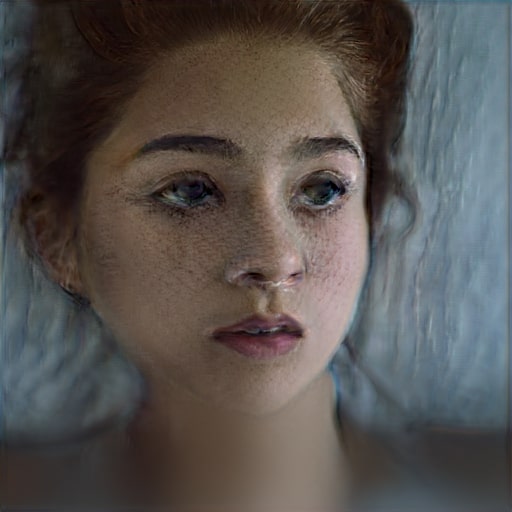}&
\includegraphics[width=0.16\linewidth,height=0.16\linewidth]
{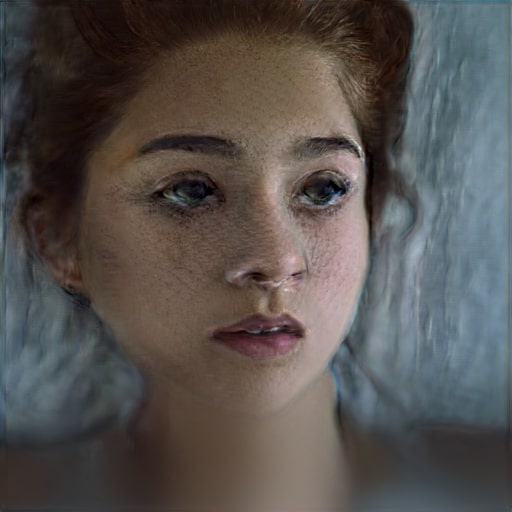}&
\includegraphics[width=0.16\linewidth,height=0.16\linewidth]
{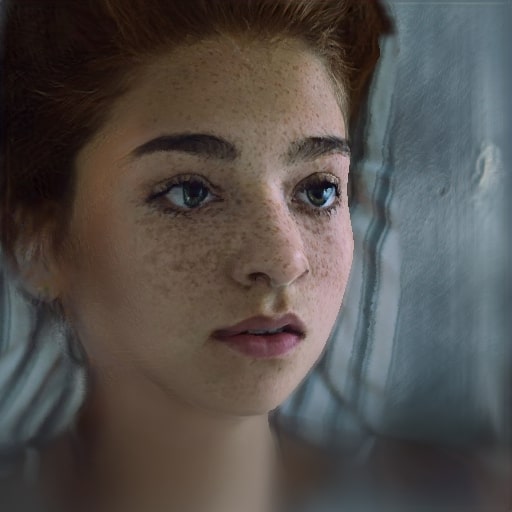} 
\put(-320,-10){$\rightarrow{}$-----------------------  Novel View Quality --------------------- $\rightarrow{}$} 
\put(-70,-10){Novel View(Ours)}\\

\end{tabular}
    \caption{Reconstruction quality vs. novel view quality during the optimization process using density regularization. Although the geometry of the synthesized novel view will not distort like in Figure 2 in the main paper, the generated textures contain visually-unpleasant noisy details compared to our results. Zoom for details}
    \label{fig:density regularization}
    \vspace{-1mm}
\end{figure*}

\subsection{Qualitative comparison} 
\begin{figure*}[t]
\centering
\small
\begin{tabular}{@{}c@{\hspace{1mm}}c@{\hspace{1mm}}c@{\hspace{1mm}}c@{\hspace{1mm}}c@{}}
\multirow{2}[2]{0.19\linewidth}[10.2mm]
{\includegraphics[width=\linewidth]{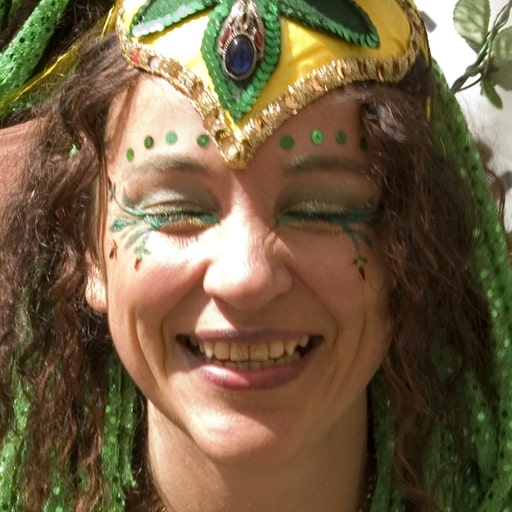}}&
\includegraphics[width=0.19\linewidth,height=0.19\linewidth]{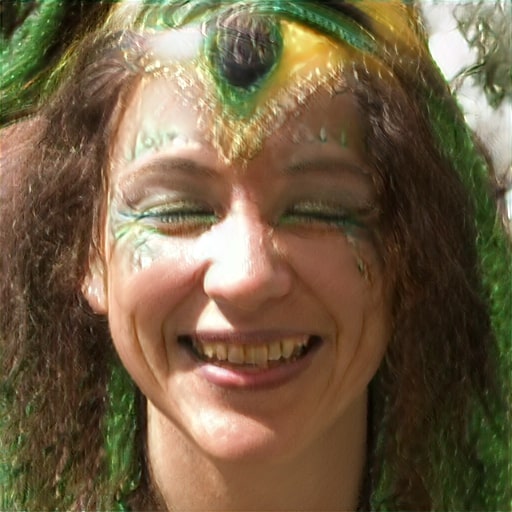}&
\includegraphics[width=0.19\linewidth,height=0.19\linewidth]{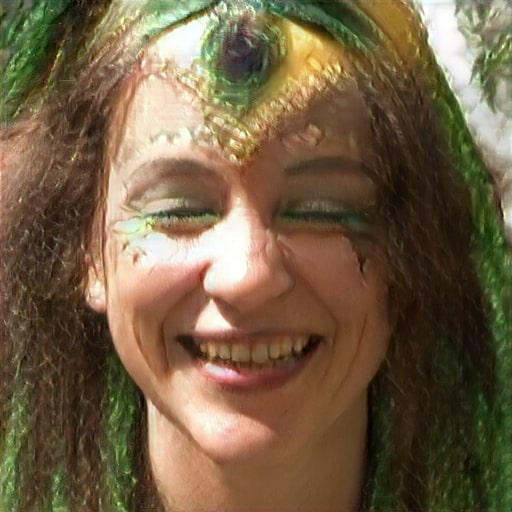}&
\includegraphics[width=0.19\linewidth,height=0.19\linewidth]{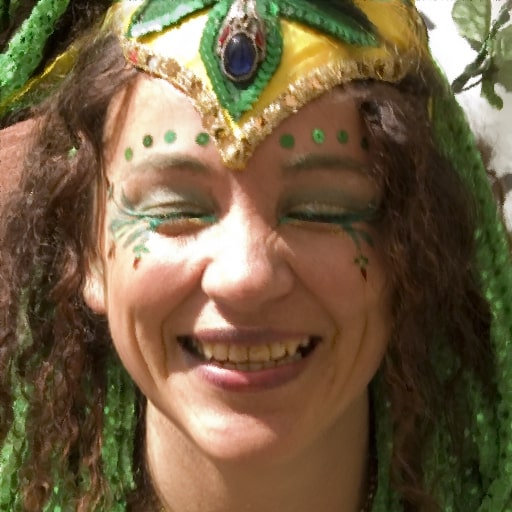}&
\rotatebox{90}{\small \hspace{3.5mm} Reconstruction}\\

&
\includegraphics[width=0.19\linewidth,height=0.19\linewidth]{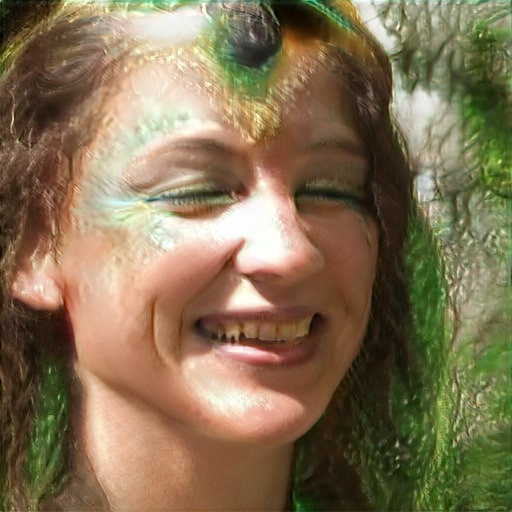}&
\includegraphics[width=0.19\linewidth,height=0.19\linewidth]{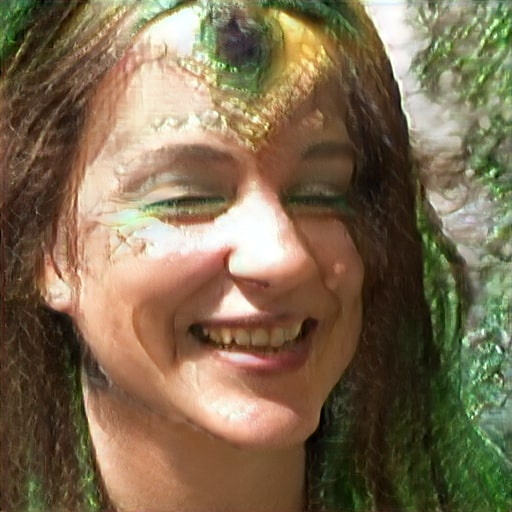}&
\includegraphics[width=0.19\linewidth,height=0.19\linewidth]{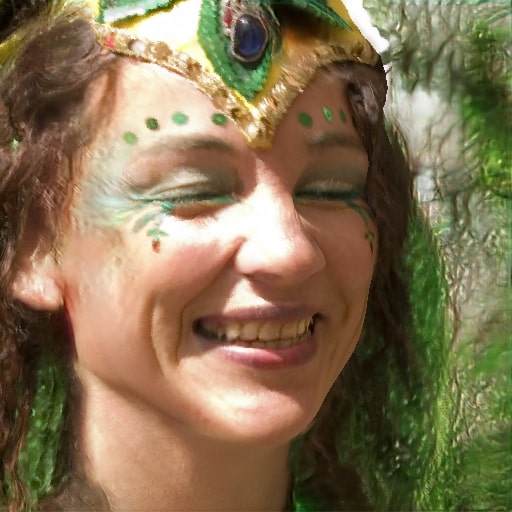}&
\rotatebox{90}{\small \hspace{5mm} Novel View}
\\

\multirow{2}[2]{0.19\linewidth}[10.2mm]
{\includegraphics[width=\linewidth]{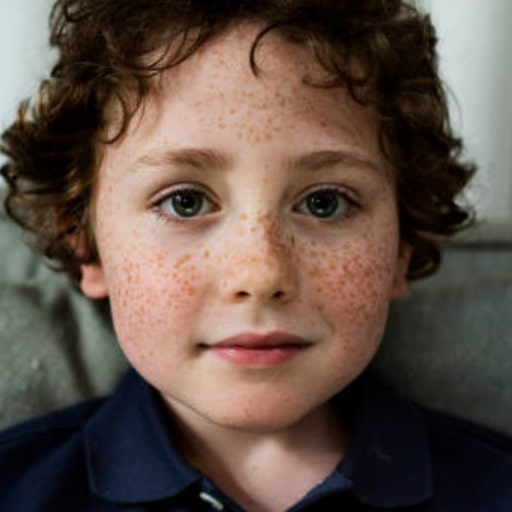}}&
\includegraphics[width=0.19\linewidth,height=0.19\linewidth]{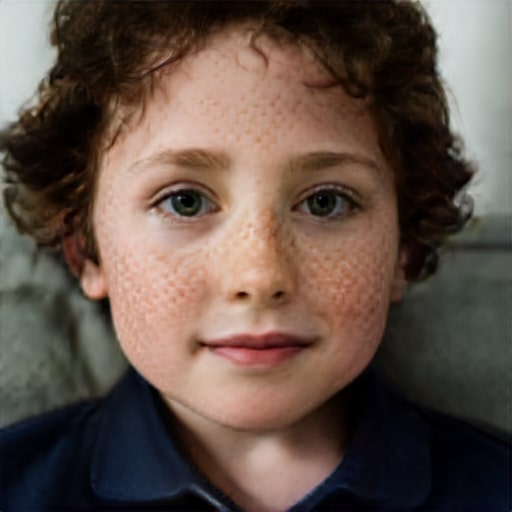}&
\includegraphics[width=0.19\linewidth,height=0.19\linewidth]{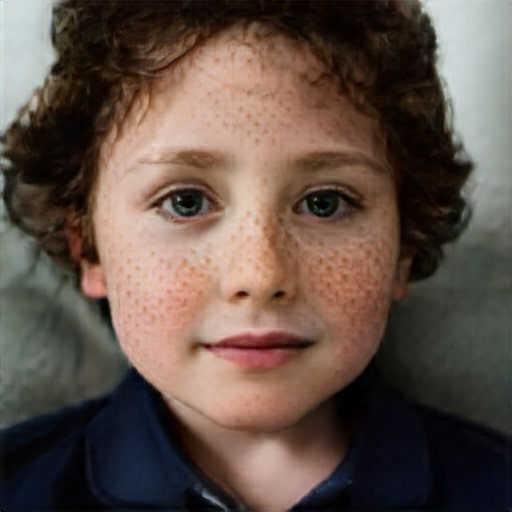}&
\includegraphics[width=0.19\linewidth,height=0.19\linewidth]{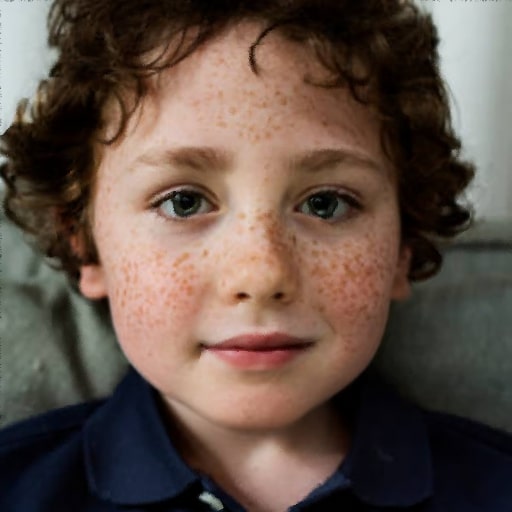}&
\rotatebox{90}{\small \hspace{3.5mm} Reconstruction}\\

&
\includegraphics[width=0.19\linewidth,height=0.19\linewidth]{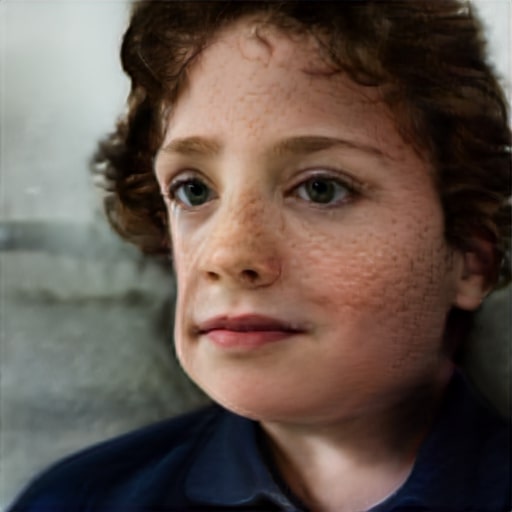}&
\includegraphics[width=0.19\linewidth,height=0.19\linewidth]{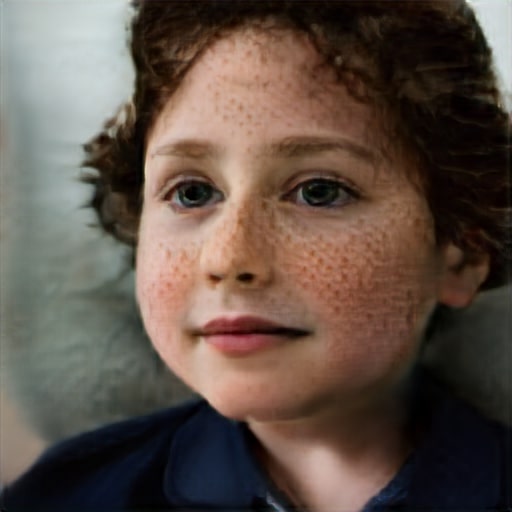}&
\includegraphics[width=0.19\linewidth,height=0.19\linewidth]{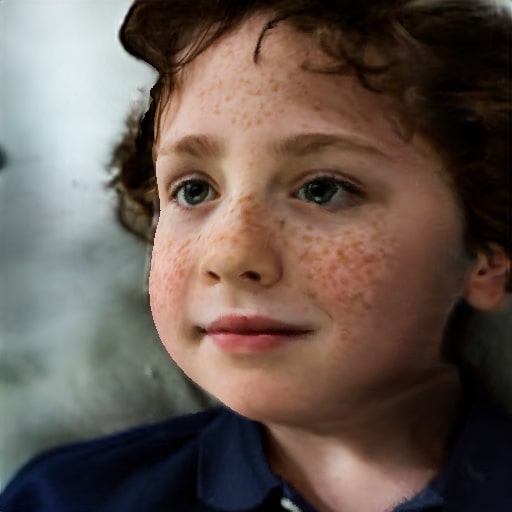}&
\rotatebox{90}{\small \hspace{5mm} Novel View}
\\

\multirow{2}[2]{0.19\linewidth}[10.2mm]
{\includegraphics[width=\linewidth]{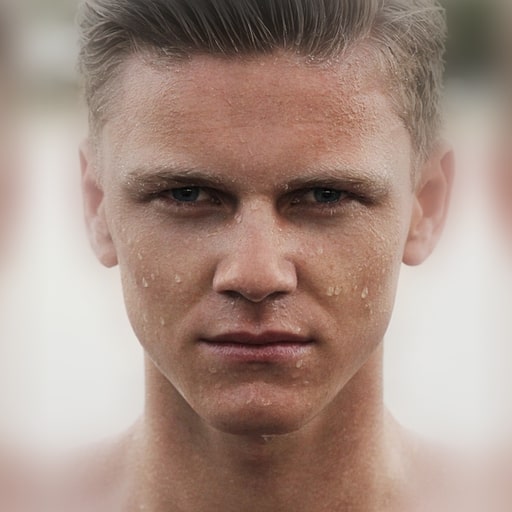}}&
\includegraphics[width=0.19\linewidth,height=0.19\linewidth]{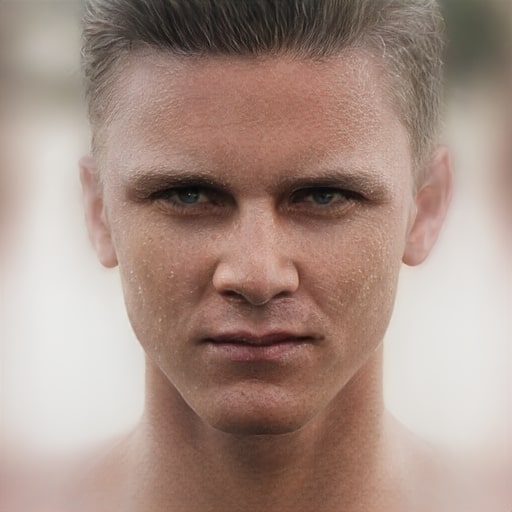}&
\includegraphics[width=0.19\linewidth,height=0.19\linewidth]{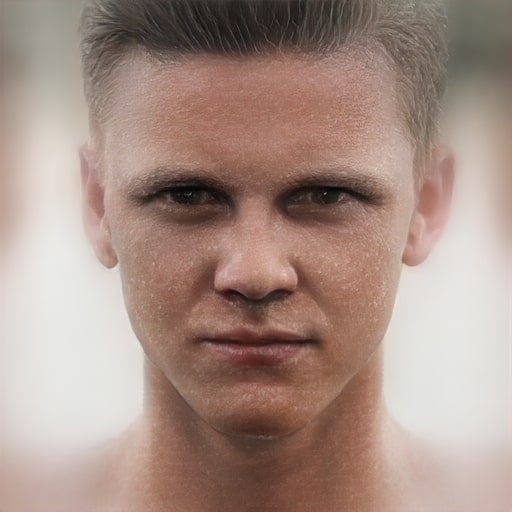}&
\includegraphics[width=0.19\linewidth,height=0.19\linewidth]{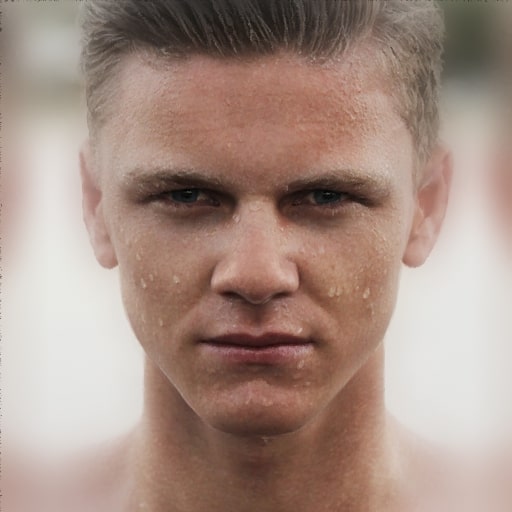}&
\rotatebox{90}{\small \hspace{3.5mm} Reconstruction}\\

&
\includegraphics[width=0.19\linewidth,height=0.19\linewidth]{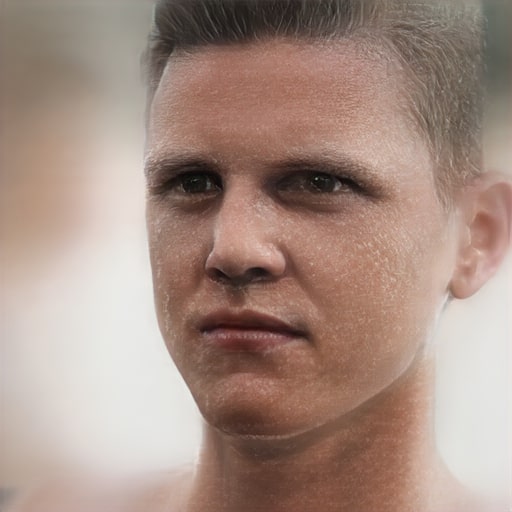}&
\includegraphics[width=0.19\linewidth,height=0.19\linewidth]{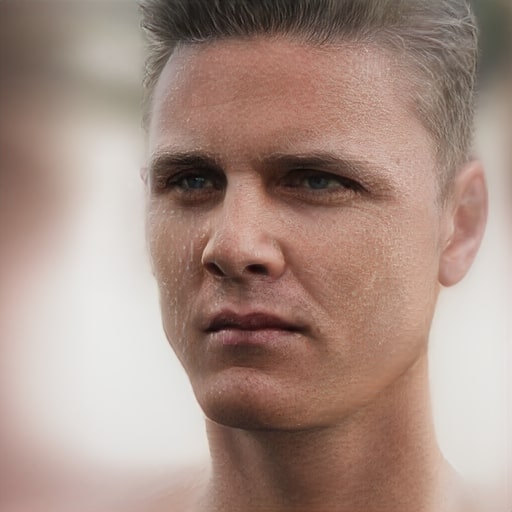}&
\includegraphics[width=0.19\linewidth,height=0.19\linewidth]{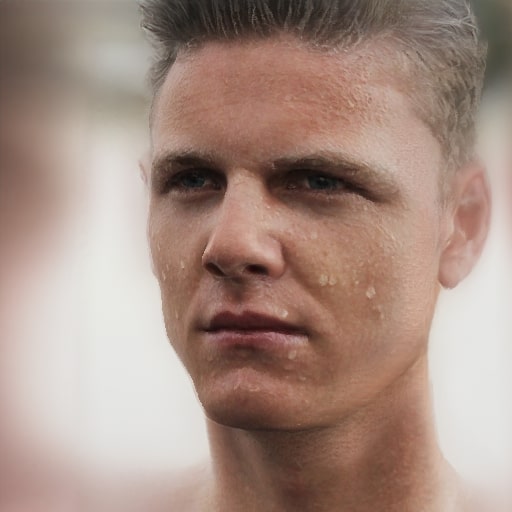}&
\rotatebox{90}{\small \hspace{5mm} Novel View}
\\

 Input image & PTI~\cite{roich2022pivotal} &IDE-3D~\cite{sun2022ide} &Ours\\ 
\end{tabular}
    \caption{{\textbf{More qualitative comparison with baselines.}}
    }
    \label{fig:more_qualitative_comparison_1}

\end{figure*}

\begin{figure*}[t]
\centering
\small
\begin{tabular}{@{}c@{\hspace{1mm}}c@{\hspace{1mm}}c@{\hspace{1mm}}c@{\hspace{1mm}}c@{}}
\multirow{2}[2]{0.19\linewidth}[10.2mm]
{\includegraphics[width=\linewidth]{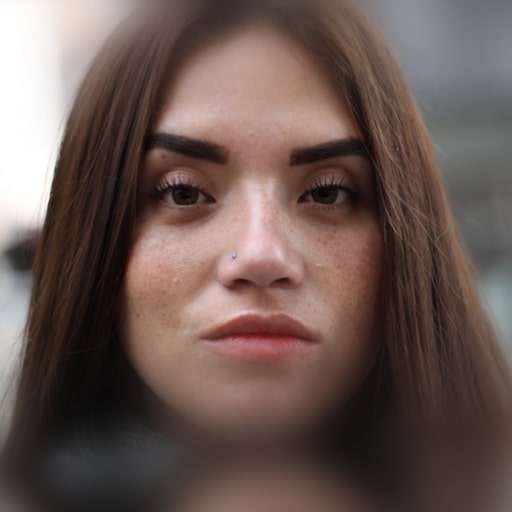}}&
\includegraphics[width=0.19\linewidth,height=0.19\linewidth]{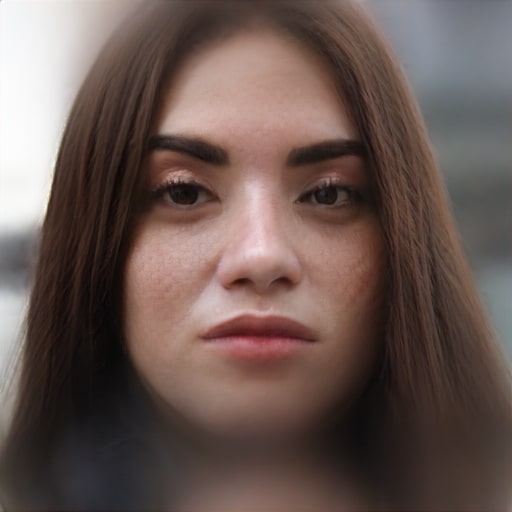}&
\includegraphics[width=0.19\linewidth,height=0.19\linewidth]{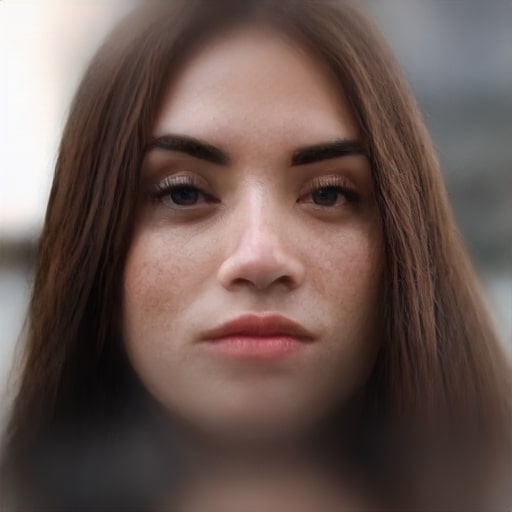}&
\includegraphics[width=0.19\linewidth,height=0.19\linewidth]{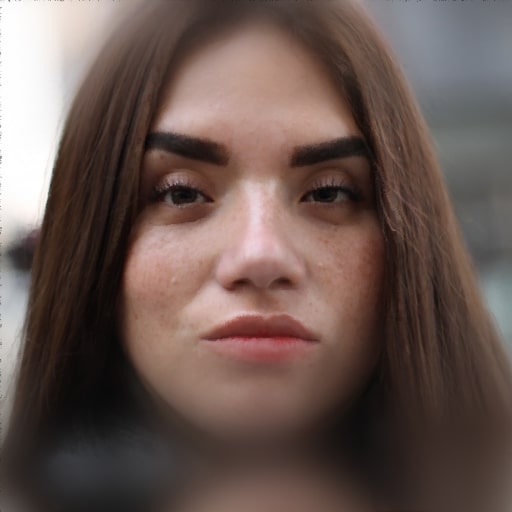}&
\rotatebox{90}{\small \hspace{3.5mm} Reconstruction}\\

&
\includegraphics[width=0.19\linewidth,height=0.19\linewidth]{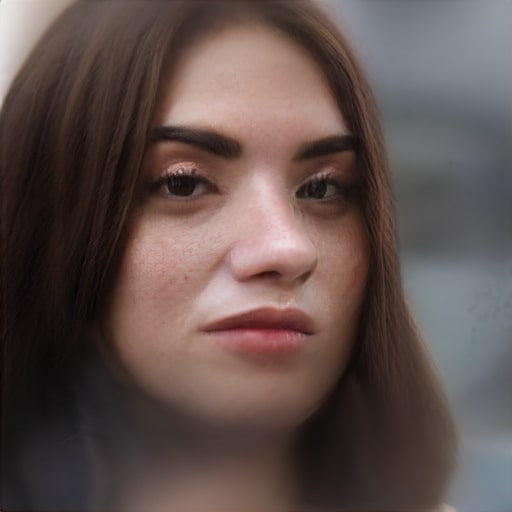}&
\includegraphics[width=0.19\linewidth,height=0.19\linewidth]{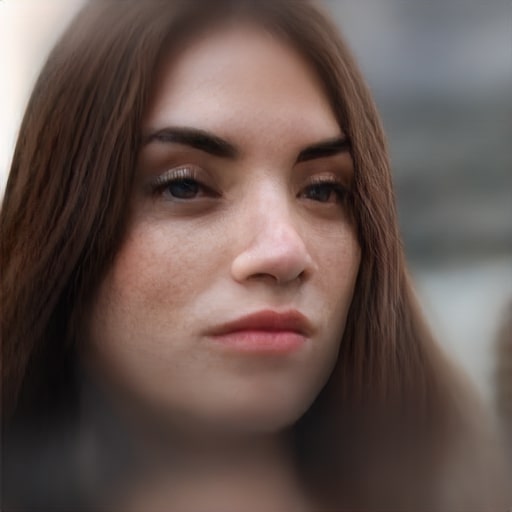}&
\includegraphics[width=0.19\linewidth,height=0.19\linewidth]{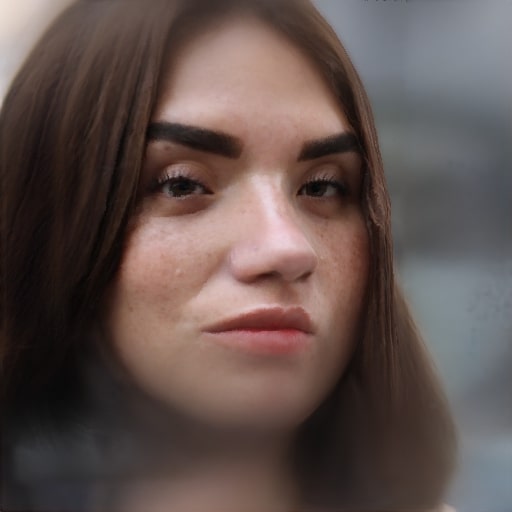}&
\rotatebox{90}{\small \hspace{5mm} Novel View}
\\

\multirow{2}[2]{0.19\linewidth}[10.2mm]
{\includegraphics[width=\linewidth]{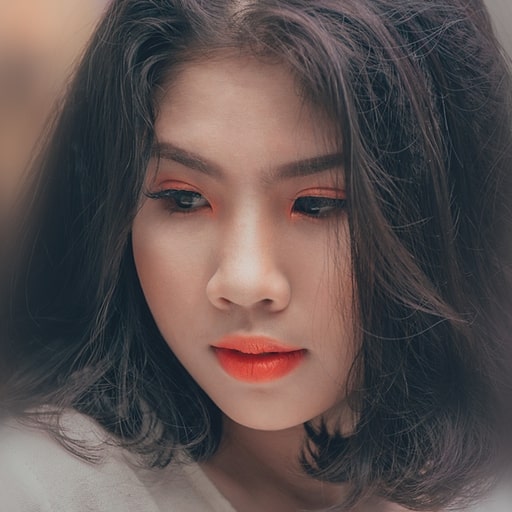}}&
\includegraphics[width=0.19\linewidth,height=0.19\linewidth]{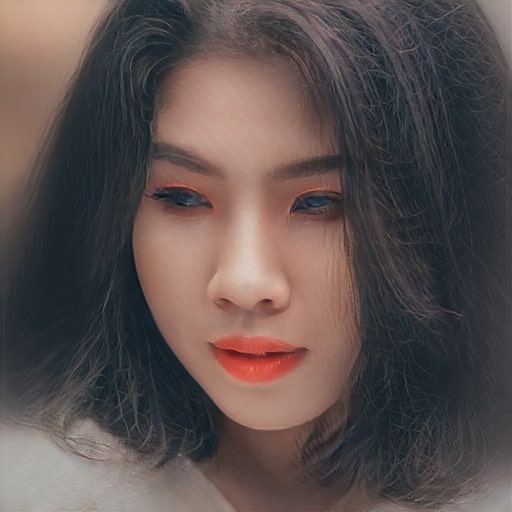}&
\includegraphics[width=0.19\linewidth,height=0.19\linewidth]{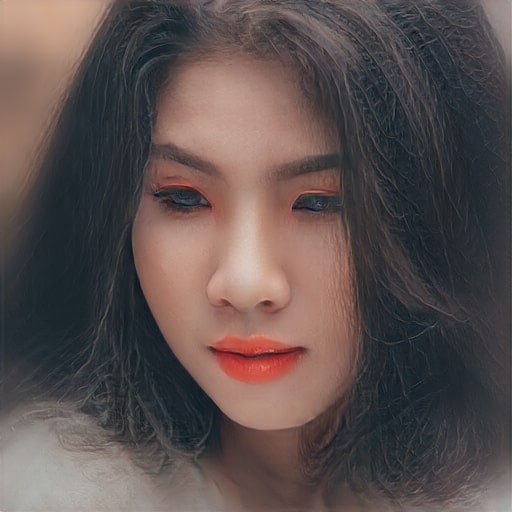}&
\includegraphics[width=0.19\linewidth,height=0.19\linewidth]{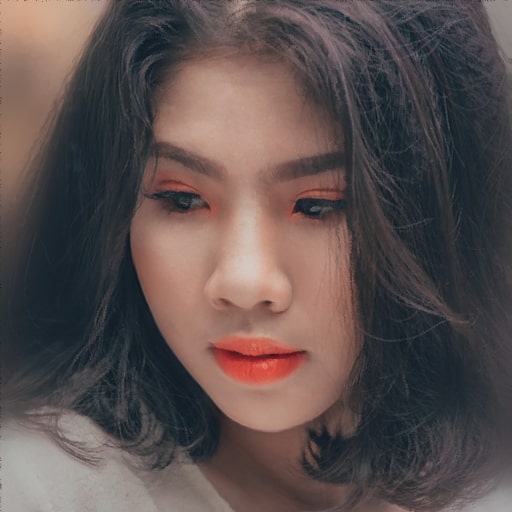}&
\rotatebox{90}{\small \hspace{3.5mm} Reconstruction}\\

&
\includegraphics[width=0.19\linewidth,height=0.19\linewidth]{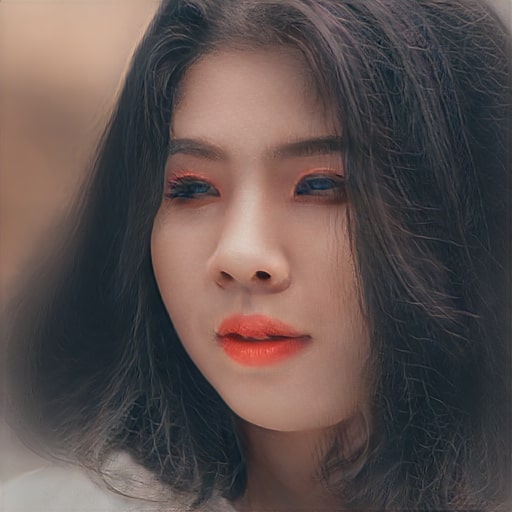}&
\includegraphics[width=0.19\linewidth,height=0.19\linewidth]{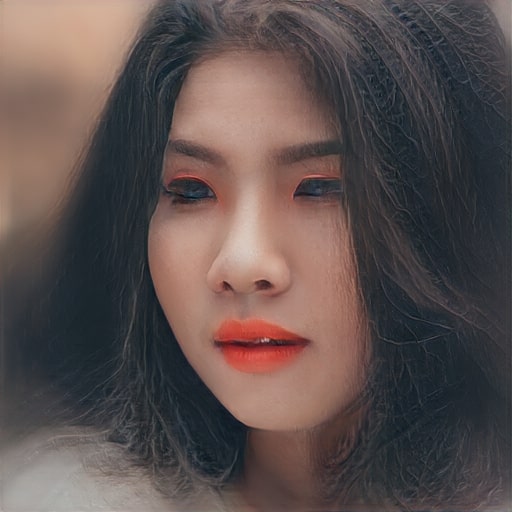}&
\includegraphics[width=0.19\linewidth,height=0.19\linewidth]{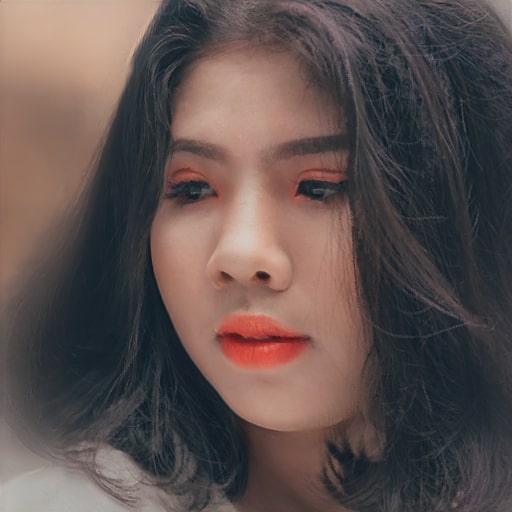}&
\rotatebox{90}{\small \hspace{5mm} Novel View}
\\

\multirow{2}[2]{0.19\linewidth}[10.2mm]
{\includegraphics[width=\linewidth]{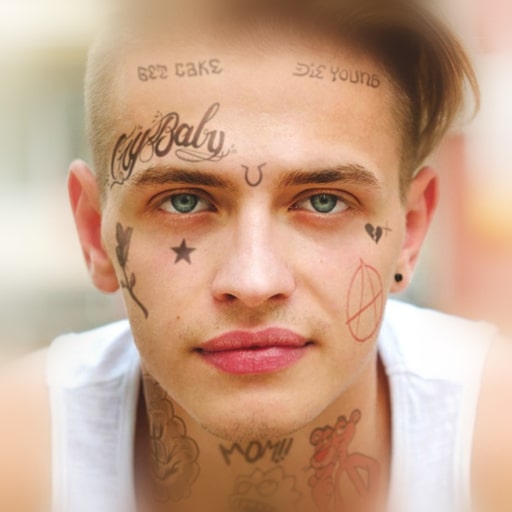}}&
\includegraphics[width=0.19\linewidth,height=0.19\linewidth]{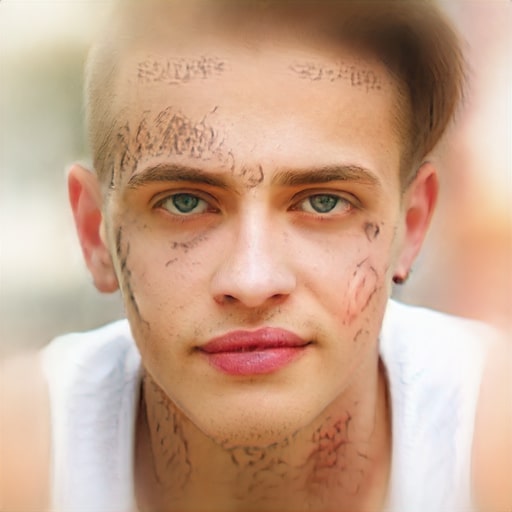}&
\includegraphics[width=0.19\linewidth,height=0.19\linewidth]{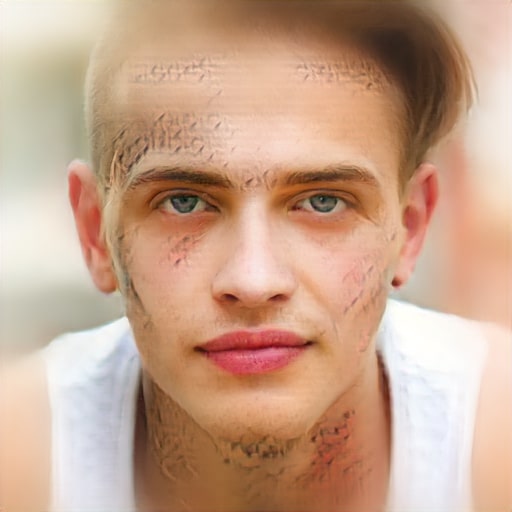}&
\includegraphics[width=0.19\linewidth,height=0.19\linewidth]{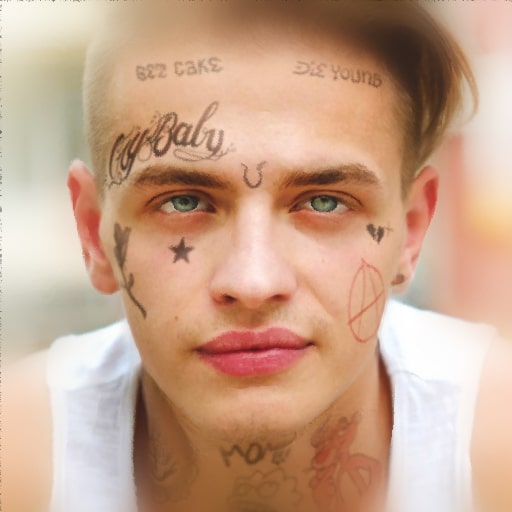}&
\rotatebox{90}{\small \hspace{3.5mm} Reconstruction}\\

&
\includegraphics[width=0.19\linewidth,height=0.19\linewidth]{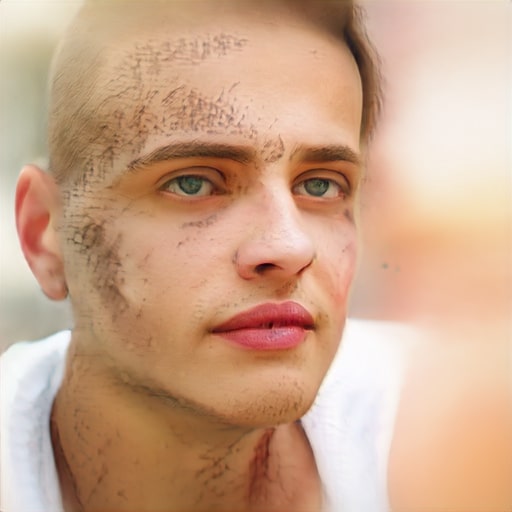}&
\includegraphics[width=0.19\linewidth,height=0.19\linewidth]{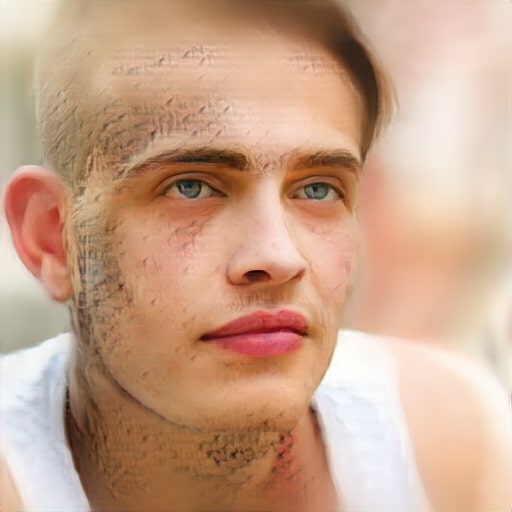}&
\includegraphics[width=0.19\linewidth,height=0.19\linewidth]{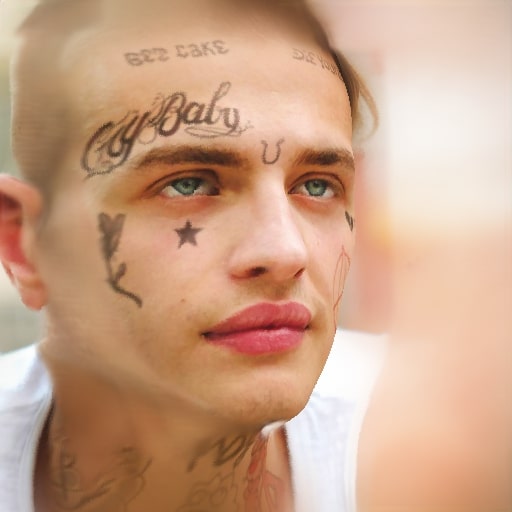}&
\rotatebox{90}{\small \hspace{5mm} Novel View}
\\

 Input image & PTI~\cite{roich2022pivotal} &IDE-3D~\cite{sun2022ide} &Ours\\ 
\end{tabular}
    \caption{{\textbf{More qualitative comparison with baselines.}}
    }
    \label{fig:more_qualitative_comparison_2}

\end{figure*}

\begin{figure*}[t]
\centering
\small
\begin{tabular}{@{}c@{\hspace{1mm}}c@{\hspace{1mm}}c@{\hspace{1mm}}c@{\hspace{1mm}}c@{}}
\multirow{2}[2]{0.19\linewidth}[10.2mm]
{\includegraphics[width=\linewidth]{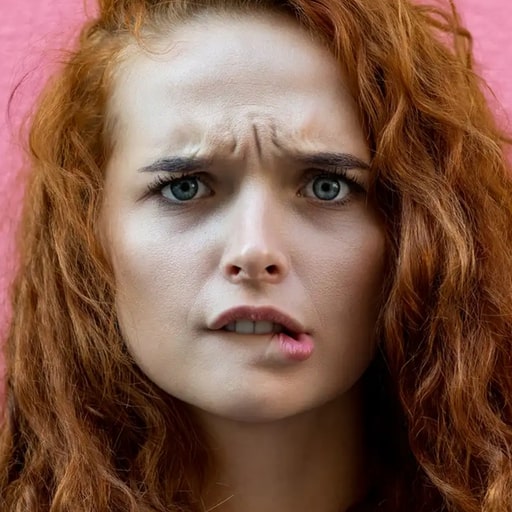}}&
\includegraphics[width=0.19\linewidth,height=0.19\linewidth]{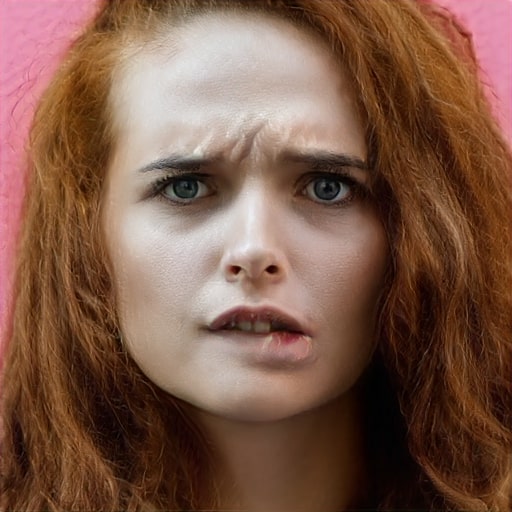}&
\includegraphics[width=0.19\linewidth,height=0.19\linewidth]{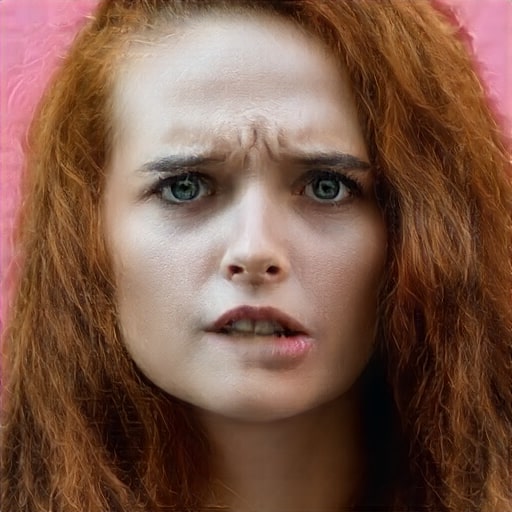}&
\includegraphics[width=0.19\linewidth,height=0.19\linewidth]{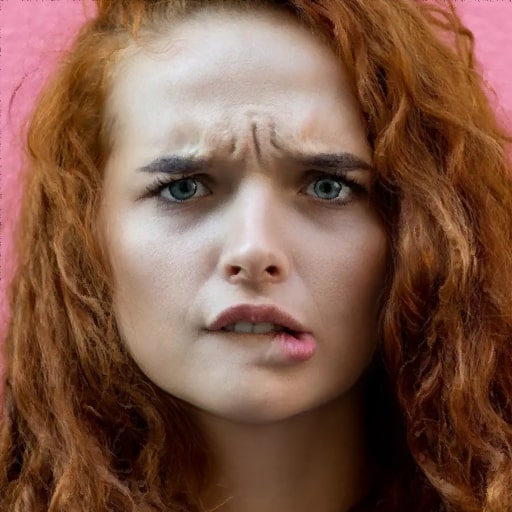}&
\rotatebox{90}{\small \hspace{3.5mm} Reconstruction}\\

&
\includegraphics[width=0.19\linewidth,height=0.19\linewidth]{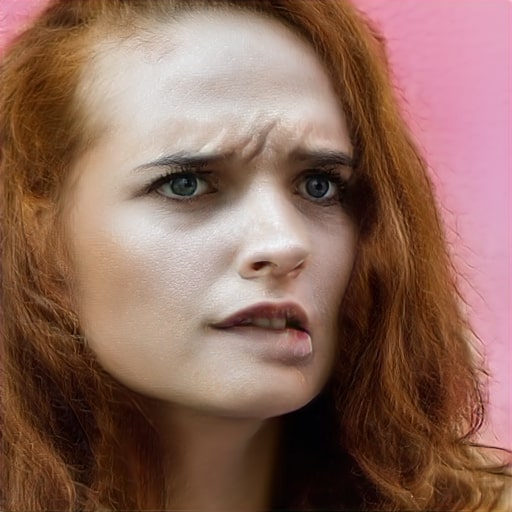}&
\includegraphics[width=0.19\linewidth,height=0.19\linewidth]{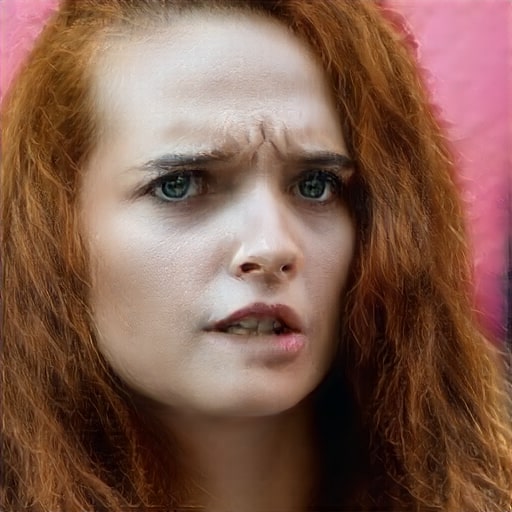}&
\includegraphics[width=0.19\linewidth,height=0.19\linewidth]{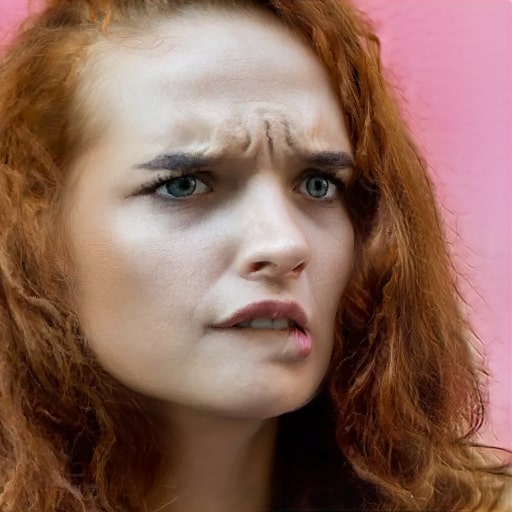}&
\rotatebox{90}{\small \hspace{5mm} Novel View}
\\

\multirow{2}[2]{0.19\linewidth}[10.2mm]
{\includegraphics[width=\linewidth]{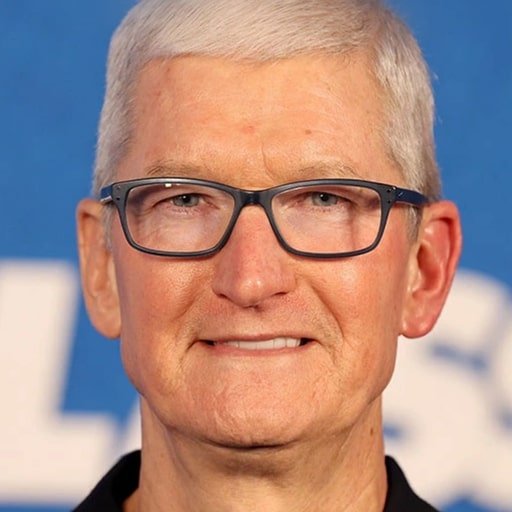}}&
\includegraphics[width=0.19\linewidth,height=0.19\linewidth]{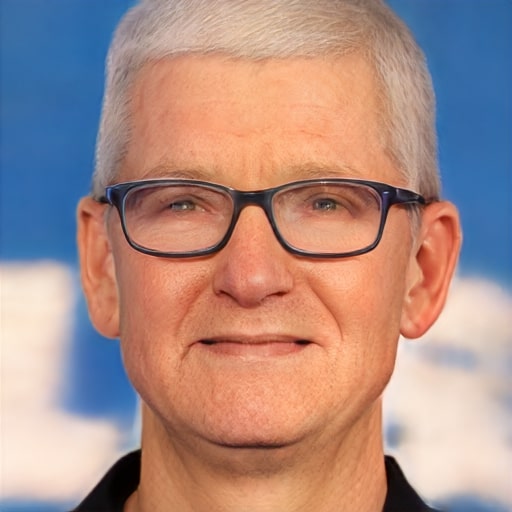}&
\includegraphics[width=0.19\linewidth,height=0.19\linewidth]{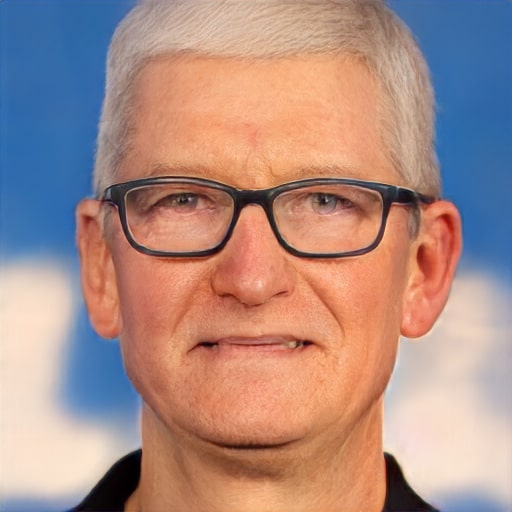}&
\includegraphics[width=0.19\linewidth,height=0.19\linewidth]{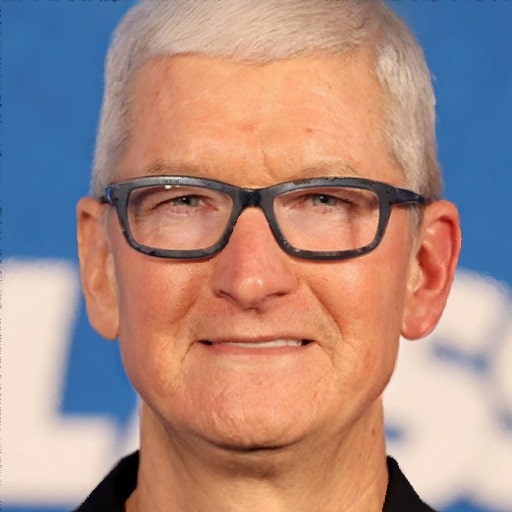}&
\rotatebox{90}{\small \hspace{3.5mm} Reconstruction}\\

&
\includegraphics[width=0.19\linewidth,height=0.19\linewidth]{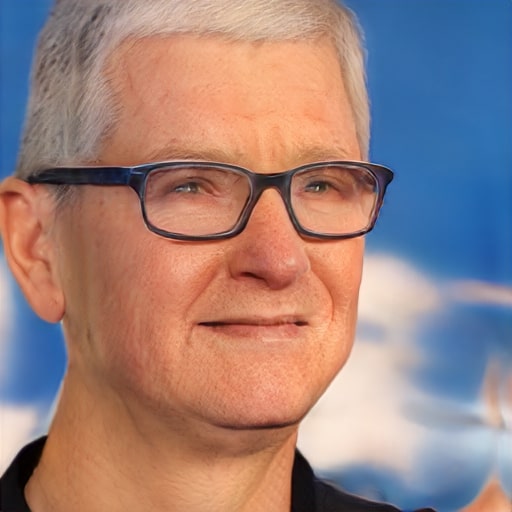}&
\includegraphics[width=0.19\linewidth,height=0.19\linewidth]{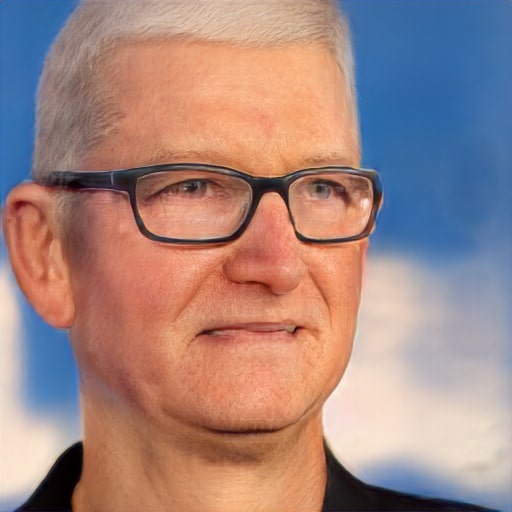}&
\includegraphics[width=0.19\linewidth,height=0.19\linewidth]{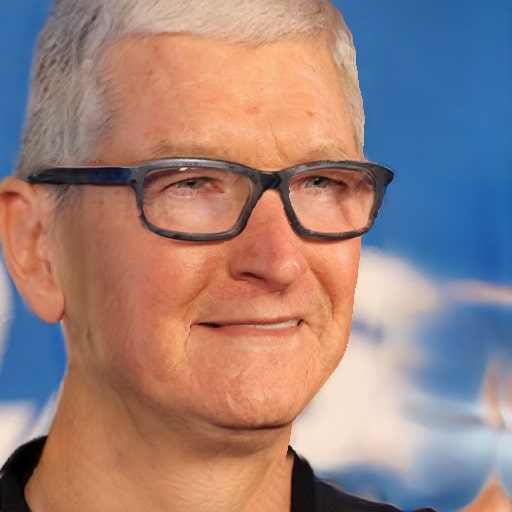}&
\rotatebox{90}{\small \hspace{5mm} Novel View}
\\

\multirow{2}[2]{0.19\linewidth}[10.2mm]
{\includegraphics[width=\linewidth]{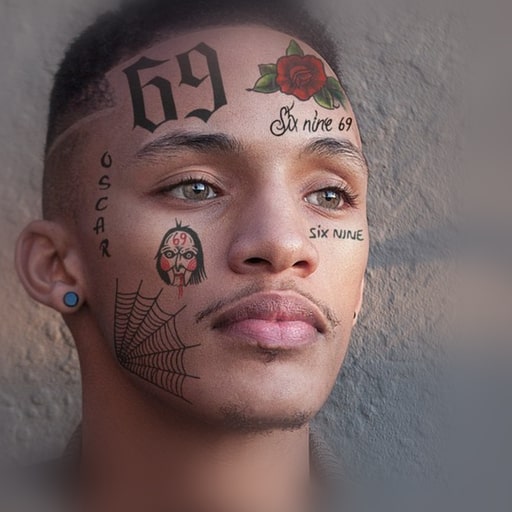}}&
\includegraphics[width=0.19\linewidth,height=0.19\linewidth]{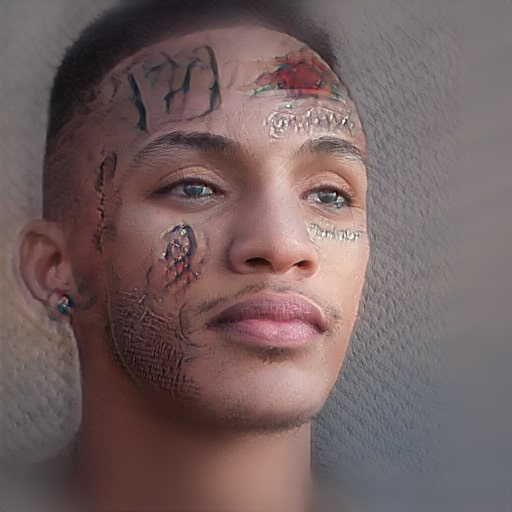}&
\includegraphics[width=0.19\linewidth,height=0.19\linewidth]{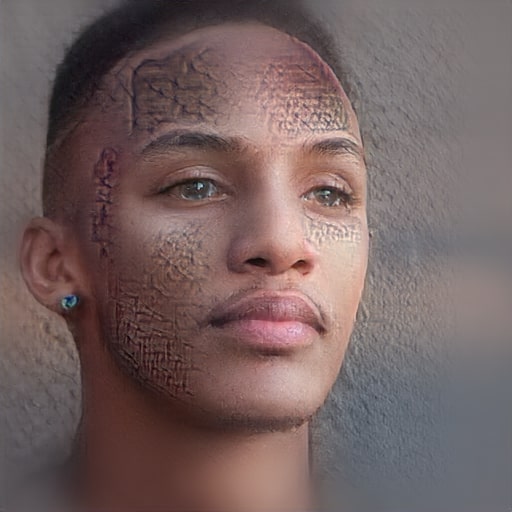}&
\includegraphics[width=0.19\linewidth,height=0.19\linewidth]{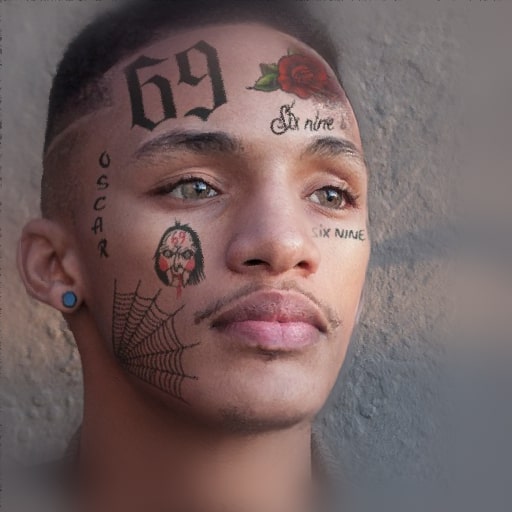}&
\rotatebox{90}{\small \hspace{3.5mm} Reconstruction}\\

&
\includegraphics[width=0.19\linewidth,height=0.19\linewidth]{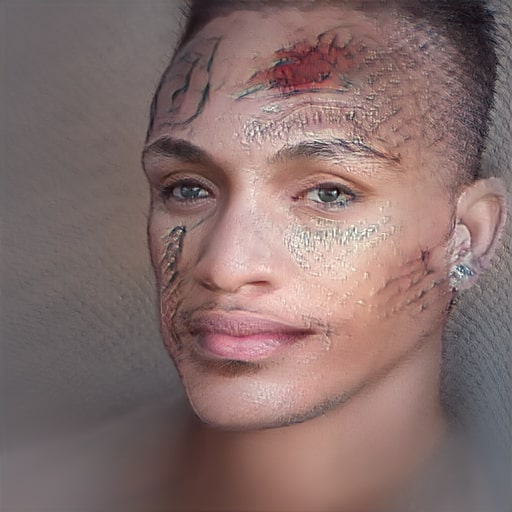}&
\includegraphics[width=0.19\linewidth,height=0.19\linewidth]{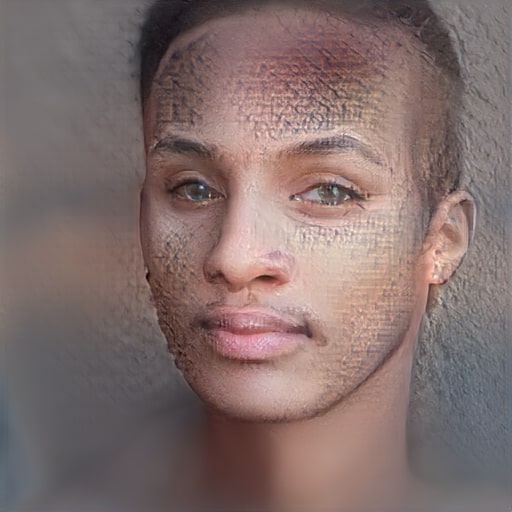}&
\includegraphics[width=0.19\linewidth,height=0.19\linewidth]{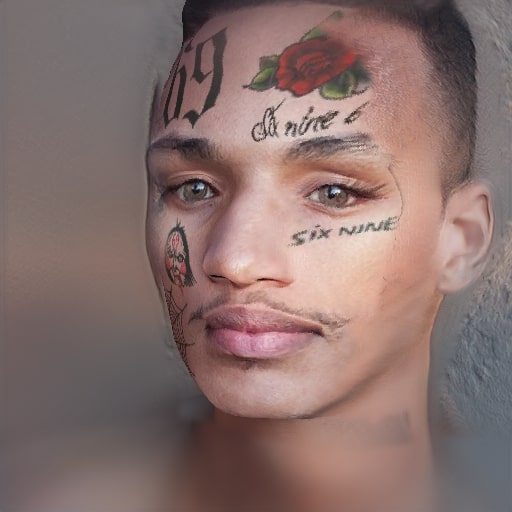}&
\rotatebox{90}{\small \hspace{5mm} Novel View}
\\

 Input image & PTI~\cite{roich2022pivotal} &IDE-3D~\cite{sun2022ide} &Ours\\ 
\end{tabular}
    \caption{{\textbf{More qualitative comparison with baselines.}}
    }
    \label{fig:more_qualitative_comparison_3}

\end{figure*}

More qualitative comparison are shown in Figure.~\ref{fig:more_qualitative_comparison_1},  Figure.~\ref{fig:more_qualitative_comparison_2} and Figure.~\ref{fig:more_qualitative_comparison_3}.

\section{Alternative choices} 
An alternative regularization strategy to improve the geometry is to add regularization on density while the reconstruction loss is still calculated on the single input. With an initially estimated geometry, we can regularize the density during the training. We add an additional loss which requires the density of the current output is similar to the correct geometry. However, although it helps to keep the geometry, the synthesized novel view contains blurry details. As in Figure~\ref{fig:density regularization}, compared to density regularization, our pseudo-multi-view generates clearer details and keeps higher fidelity. The possible reason is that the single input as supervision contains not enough information for constructing photo-realistic details. The pseudo-multi-view can better solve the ambiguity.

\clearpage


\end{document}